\documentclass[svgnames]{article}
\usepackage{times}

\usepackage[letterpaper,margin=1in]{geometry}

\usepackage[utf8]{inputenc} 
\usepackage[T1]{fontenc}    
\usepackage[numbers,sort&compress]{natbib} 
\usepackage{url}            
\usepackage{booktabs}       
\usepackage{amsfonts}       
\usepackage{nicefrac}       
\usepackage{microtype}      
\usepackage{xcolor}         

\usepackage{etoolbox}   

\newtoggle{restatingtheorem}
\togglefalse{restatingtheorem}

\usepackage{amsmath,amssymb,amsthm}
\usepackage{thmtools}
\usepackage{mathtools}
\usepackage{float}

\usepackage{subcaption}
\usepackage{enumitem}
\setlist{noitemsep,topsep=0pt,parsep=0pt,partopsep=0pt,leftmargin=*}

\usepackage{lipsum}
\usepackage{pgf}
\usepackage{tikz}
\usepackage{pgfplots}
\pgfplotsset{compat=1.18}

\usepackage{booktabs}      
\usepackage{tabularx}
\usepackage{wrapfig}
\usepackage{multirow}
\usepackage{adjustbox}

\colorlet{MyBlue}{DodgerBlue!75!Black}
\colorlet{MyGreen}{DarkGreen!85!Black}
\colorlet{MyGray}{White!75!Black}

\usepackage[pagebackref]{hyperref}       

\renewcommand*{\backrefalt}[4]{
\ifcase #1 %
  Not cited.%
\or
  {\footnotesize Cited on page~#2. }%
\else
  {\footnotesize Cited on pages~#2. }%
\fi}

\hypersetup{
final,
colorlinks=true,
linktocpage=true,
pdfstartview=FitH,
breaklinks=true,
pdfpagemode=UseNone,
pageanchor=true,
pdfpagemode=UseOutlines,
plainpages=false,
bookmarksnumbered,
bookmarksopen=false,
bookmarksopenlevel=1,
hypertexnames=true,
pdfhighlight=/O,
urlcolor=Maroon,linkcolor=MyBlue!60!black,citecolor=DarkGreen!70!black,	
pdftitle={},
pdfauthor={},
pdfsubject={},
pdfkeywords={},
pdfcreator={pdfLaTeX},
pdfproducer={LaTeX with hyperref}
}

\usepackage[nameinlink,capitalise]{cleveref}
\crefname{assumption}{Assumption}{Assumptions}

\usepackage{autonum}

\usepackage{algorithm}
\usepackage{algorithmic}

\usepackage{subcaption}
\usepackage{graphicx}

\usepackage[symbols,nogroupskip,sort=none]{glossaries-extra}
\usepackage{nomencl}

\usepackage{titletoc}

\newtheorem{definition}{Definition}
\newtheorem{proposition}{Proposition}
\newtheorem{theorem}{Theorem}
\newtheorem{lemma}{Lemma}
\newtheorem{example}{Example}
\newtheorem{corollary}{Corollary}

\newtheorem{assumption}{Assumption}
\newtheorem{remark}{Remark}

\newtheorem{nonexample}{Non-example}

\usepackage{color-edits}
\addauthor{gb}{blue}
\addauthor{jl}{orange}
\addauthor{fa}{green}
\addauthor{ag}{red}

\usepackage{color-edits}
\usepackage{diagbox}

\colorlet{MyGreen}{DarkGreen!80!Black}
\addauthor{gb}{MyGreen}
\addauthor{jl}{orange}
\addauthor{ab}{purple}
\addauthor{ak}{green}
\addauthor{fa}{green} 
\addauthor{ash}{blue}

\newcommand{\dist}{\operatorname{dist}}
\newcommand{\iid}{\overset{\smash{\text{\tiny iid}}}{\sim}}
\newcommand{\Exp}[1]{\textbf{(E$_{#1}$)}}

\newcommand{\Var}[1]{\operatorname{Var}}

\newcommand{\PrimalUpdate}{\texttt{PrimalUpdate}}

\title{Stochastic Penalty-Barrier Method for\\Constrained Machine Learning}

\author{%
\parbox[t]{0.49\textwidth}{\centering
{\small\linespread{0.96}\selectfont
\textbf{Adam Bos\'ak}, \textbf{Andrii Kliachkin}\\
Artificial Intelligence Center\\
Czech Technical University in Prague\\
}
}\hspace{0.1em}
\parbox[t]{0.49\textwidth}{\centering
{\small\selectfont
\textbf{Gilles Bareilles}\\
Centre de Math\'ematiques Appliqu\'ees\\
\'Ecole Polytechnique\\
}
}\\[2.5em]
\parbox[t]{\textwidth}{\centering
{\small\selectfont
\textbf{Allen Gehret}, \textbf{Allahkaram Shafiei}, \textbf{Jana Lep\v{s}ov\'a}, \textbf{Jakub Mare\v{c}ek}\\
Artificial Intelligence Center\\
Czech Technical University in Prague\\
}
}%
}
\date{}

\begin{document}

\maketitle

\begin{abstract}

Constrained Machine Learning (CML) enables fairness-aware training, physics-informed neural networks, and integration of symbolic  domain knowledge into statistical models. In this work, we introduce the Stochastic Penalty-Barrier Method (SPBM) for CML problems. SPBM extends classical penalty and barrier methods by incorporating an exponential averaging of the dual variables, a stabilized penalty schedule, and the Moreau envelope to handle non-smoothness. We analyze the bias that mini-batching introduces in the barrier function and show that the feasible set of the resulting transformed problem is contained within the original one. We compare SPBM with CML baselines across multiple fairness and physics informed neural networks experiments. We find that SPBM is competitive with state-of-the-art methods. We also observe, on our fairness-based computational benchmark, that the per-epoch runtime of CML methods is largely independent of the number of constraints, and within $1.3\times$ of the per-epoch runtime of regularized Adam, for a number of constraints ranging from $90$ to $9900$.

\end{abstract}

\section{Introduction}

In recent years, artificial intelligence (AI) has become an indispensable part of many fields such as healthcare, credit scoring, criminal justice or the car industry.
In such high-risk settings, AI predictions should be as accurate as possible, but also meet critical requirements,
whether normative, such as fairness or safety, or structural, such as consistency with physical laws.
There, constrained machine learning (CML) becomes increasingly important \citep{ramirezPositionAdoptConstraints2025} enabling the enforcement of fairness constraints \citep{liKernelDependenceRegularizers2022,buyl2024fairret}, safety constraints \citep{pmlr-v70-achiam17a}, or ensuring consistency with physical laws \citep{lu2021physics,son2023enhanced, song2024admm}.

From a mathematical point of view, general CML is cast as an optimization problem with a stochastic non-convex non-smooth objective subject to stochastic non-convex non-smooth constraints:
\begin{equation}
\label{eq:stoctrsto}
    \min_{x\in\mathbb{R}^n} f(x) \quad \text{s.t.} \quad g(x) \leq 0,
\end{equation}
where $f(x) = \mathbb{E}_{\xi\sim\Xi}[f(x, \xi)]$ and $g(x) = \mathbb{E}_{\xi\sim\Xi}[g(x, \xi)]$. On the one hand, there is a substantial and recent body of work on CML with simplifying assumptions:  \citet{curtis2026single,fang2026trust} focus on deterministic constraints; \citet{curtis2025stochastic} consider deterministic linear constraints; \citet{necoara2021minibatch,necoara2022stochastic,singh2024mini,singh2024stochastic, boob2023stochastic} address stochastic constrained convex composite problems; finally, \citet{HuaLin2023} assume stochastic non-convex and weakly convex constraints in their Stochastic Switching Subgradient (SSw). Interestingly, SSL-ALM, a method of \citet{pmlr-v267-huang25au}, was shown by \citet{kliachkin2026benchmarking} to be the state of the art for general CML, despite being proposed for stochastic weakly-convex objective with linear constraints. On the other hand, \citet{kliachkin2026benchmarking} identify that the above-mentioned settings do not capture realistic CML problems. Therefore;
\begin{quote}
\emph{It remains unclear how to implement CML efficiently, and little can be said about the theoretical guarantees of methods for these problems.}
\end{quote}
In this paper, we aim to partially bridge this gap. Firstly, we propose a new method, called Stochastic Penalty-Barrier Method (SPBM), and give a theoretical analysis of the transformed problem SPBM solves and of a mild convergence result. Secondly, we analyse the computational demand of CML methods to answer whether it is feasible to train deep neural networks with constraints. Lastly, we show why the general CML setting is difficult to solve. Specifically, we exhibit an empirical risk minimization (ERM) problem built from a two-parameter single-ReLU network that neither SPBM nor state-of-the-art methods can solve. This raises the question: \emph{What are natural assumptions that would include a general CML problem while remaining strong enough to prove convergence results for \textbf{some} practical algorithm?}
In this paper, we assume the mini-batch estimates $\bar f, \bar g$ to be integrable in~$\xi$, and locally Lipschitz for every realisation. In addition, we assume a tame setting, in which both the objective and the constraints are definable in a common o-minimal structure, enabling the mild convergence result of SPBM. In Appendix~\ref{sec:preliminaries} we give background on o-minimality and argue that tameness is a natural assumption in deep learning. However, whether the o-minimal assumption is an answer to our question remains unresolved. The contributions of this paper are as follows:
\begin{itemize}
    \item \emph{New method:} We introduce the Stochastic Penalty-Barrier Method (SPBM) for CML problems, which extends deterministic penalty-barrier methods \citep{ben1997penalty} by exponential averaging of the dual variables, a stabilized penalty schedule, and the Moreau envelope to handle non-smoothness. Interestingly, for a specific choice of parameters, SPBM reduces to a SSL-ALM-like primal-dual update (see Appendix~\ref{sec:equivalence_to_alm}). We compare SPBM with CML baselines across multiple fairness and physics-informed neural networks experiments. Our experiments show that SPBM matches or outperforms SSL-ALM.
    \item \emph{Analysis:} We analyze the bias that mini-batching introduces in the barrier function and show that the feasible set of the resulting transformed problem is contained within the original one. We also prove almost sure convergence of expected values of the Lagrangian in the o-minimal setting, although \emph{we provide no complementarity or feasibility} guarantees.
  \item \emph{Why is CML difficult:} we provide a simple CML instance on which SPBM, SSL-ALM, SSw, and methods that assume simplified weakly-convex settings fail to achieve feasibility, and can remain arbitrarily far from the feasible set; complementing known hardness results for unconstrained nonsmooth non-convex optimization \citep{zhang2020complexity, arjevani2023lower}.
  \item \emph{Open-source implementation:} we release our implementation, in the form of a drop-in addition to PyTorch unconstrained neural network training loops requiring only a few additional lines of code.
\end{itemize}
\paragraph{Related work.} Several works offer different routes to CML. \citet{Cotter2019} tackle indicator constraints via a two-player game whose solution, although approximately feasible, is a mixture of up to $m+1$ models, which is impractical when $m$ is large. \citet{chamon2022constrained} reduce the original stochastic constrained problem to a deterministic unconstrained one, and bound the resulting duality gap. The bound requires rich parametrization of the network, which may fail for smaller networks. \citet{hounieResilientConstrainedLearning2023} build upon this theory and provide an adaptive relaxation method that loosens constraints that are harder to satisfy. \citet{elenterNearOptimalSolutionsConstrained2024} bound the infeasibility of Lagrangian minimizers associated with optimal dual variables by reparametrizing the non-convex CML problem over a function space. Their final result, however, applies to exact minimizers of the Lagrangian rather than to the iterates produced in practice, and relies on the Fréchet derivative of the (unparametrized) constraints having full row rank at the optimum -- a LICQ condition that they acknowledge is not straightforward to satisfy. Libraries for CML include:  Cooper, the PyTorch framework for constrained deep learning \citep{gallego-posadaCooperLibraryConstrained2025}, implementing a single-loop ALM without smoothing; and humancompatible.train \citep{kliachkin2025humancompatible} with smoothed ALM. 

Recent works in optimization by
\citet{necoara2021minibatch,necoara2022stochastic,singh2024mini,singh2024stochastic} and \citet{boob2023stochastic} address functionally constrained convex composite problems. This allows guarantees, but does not extend to the non-convex setting of problem \eqref{eq:stoctrsto}. There also exist SQP-based methods; their assumptions include stochastic objectives subject to deterministic constraints \citep{fang2026trust,CurRob2024}, stochastic non-convex objectives subject to deterministic constraints \citep{BerCur2023, Curtis2024, FaNaMaKo2024, NaAnKo2022}, stochastic non-convex objective subject to deterministic non-convex constraints \citep{NaAnKo2023}, and finally stochastic objectives subject to stochastic constraints \citep{Ozto2023}. These SQP methods, however, require the computation and storage of a Hessian, which  makes them impractical in our case where
$f$ is a neural network. An approach similar to SQP was proposed by \cite{ma2020quadratically} for weakly-convex objectives and constraints, which solves a~sequence of strongly convex subproblems formed with a~quadratic regularization term; and by \cite{FacKun2023} for stochastic objective and constraints, which computes the update direction by solving a number of quadratic subproblems in each iteration.  Methods incorporating an augmented Lagrangian include \citep{ShiWaWa2022} with stochastic non-convex objective under deterministic non-convex inequality and/or equality constraints and \citep{Bolla2023} with stochastic objectives subject to deterministic linear constraints. Finally, some interior point methods have been proposed to solve non-convex constrained optimization \citep{curtis2025stochastic} (bound constraints) and \citep{curtis2026single} (nonlinear inequality constraints).
However, they assume a strictly feasible initial point or deterministic constraints, making them unsuitable for our nonconvex stochastic constrained setting. \citet{yang2025single,liu2026spider} present single-loop algorithms designed for non-smooth weakly-convex problems. The penalty-barrier multiplier method (PBM) was generalized to solve deterministic non-convex non-linear problems \citep{kovcvara2003pennon, kovcvara2003pennonb, kocvara2012pennon}. Recently, \citet{de2025penalty}
defined a deterministic non-convex variant of problem~\eqref{eq:stoctrsto} by combining a barrier term with an $L_1$ penalty into a new objective function.
Our method is also closely related to the proximal augmented Lagrangian method, which has been studied since the 1970s~\citep{rockafellar1976augmented, rockafellar1974augmented, rockafellar1973multiplier, rockafellar1973dual, powell1969method, hestenes1969multiplier, kong2019complexity, kong2023iteration, pu2024smoothed, lin2022complexity, zhang2020proximal, zhang2022iteration}.

\section{Review of Penalty/Barrier Multiplier Method (PBM)}
\label{sec:recspbm}

The PBM method \citep{ben1997penalty} was proposed for convex programming problems:
\begin{equation}\label{eq:convexprob}
f^* = \inf\{ f(x) : g_i(x) \leq 0, \; i = 1, \dots, m, \;  x \in \mathbb{R}^n\},
\end{equation}
where $f, g_1, \dots, g_m$ are closed proper convex functions.
Ben-Tal and Zibulevsky showed that PBM is closely related to the proximal point algorithm in the sense of the sequence of generated dual variables.
One of the main ideas behind the method is the transformation of the constraints onto a uniform space using a strictly increasing twice differentiable strictly convex function $\varphi$ with some additional properties; see \cite[Section 3]{ben1997penalty} for details.
These properties imply that, given a positive number $p$, the function $t \mapsto \varphi(t/p)$ is convex increasing and
\begin{equation}\label{eq:varphi_transformation}
    p\varphi(t/p) \leq 0 \quad \text{ if and only if } \quad t \leq 0.
\end{equation}
Using \cref{eq:varphi_transformation}, the original problem \eqref{eq:convexprob} can be rewritten as
\begin{equation}\label{eq:convexprobtransformed}
f^* = \inf\{ f(x) : p_i\varphi(g_i(x) / p_i) \leq 0, \; i = 1, \ldots, m, \; x \in \mathbb{R}^n, p \in \mathbb{R}_{++}^m \},
\end{equation}
where $p_i$ is referred to as the \textit{penalty parameter} associated with the $i$-th constraint, and $\varphi$ is referred to as the \textit{penalty-barrier function}.
The function $\mathcal{L} : \mathbb{R}^n \times \mathbb{R}_{+}^m \times \mathbb{R}_{++}^m \rightarrow \mathbb{R}$, referred to as the \textit{Lagrangian} for problem \eqref{eq:convexprobtransformed}, has the form
\( \mathcal{L}_p(x, \lambda) = f(x) +  \sum_{i=1}^m\lambda_i p_i\varphi(g_i(x) / p_i) \)
and it is used in the following iterative scheme for solving \eqref{eq:convexprobtransformed}:
\begin{align}
      &x^{k+1} = \arg \min_x \mathcal{L}_{p^k}(x, \lambda^k)\label{eq:scheme1}\\
      &\lambda^{k+1} = \lambda^k \odot \varphi'\left(g(x^{k+1}) \oslash p^k\right) \label{eq:scheme2}\\
      &p^{k+1} = \pi^k(\lambda^{k+1})\label{eq:scheme3}
\end{align}
where $\odot$ and $\oslash$ denote the element-wise product and division, and $\varphi'$ and $\pi^k$ are applied coordinate-wise.
The initial \textit{dual variable} $\lambda^0$ is required to be positive and $\pi^k : \mathbb{R}_{++}\rightarrow \mathbb{R}_{++}$ is referred to as the \textit{penalty updating function}.
We recall two glued penalty/barrier functions:
the \textit{Quadratic-Logarithmic penalty/barrier function} $\varphi_{\text{QL}}$ and the \textit{Quadratic-Reciprocal penalty/barrier function} $\varphi_{\text{QR}}$:
\begin{equation}\label{eq:penalty_barrier_functions}
\begin{aligned}
\varphi_{\text{QL}}(t) =
\begin{cases}
t + \frac{1}{2}t^2 & \text{if } t \ge -\frac{1}{2}, \\
-\frac{1}{4}\log(-2t) - \frac{3}{8} & \text{if } t < -\frac{1}{2},
\end{cases}
\quad
\varphi_{\text{QR}}(t) =
\begin{cases}
t + \frac{1}{2}t^2 & \text{if } t \ge -\frac{1}{3}, \\
\frac{32}{27}(\frac{1}{1-t}) - \frac{7}{6} & \text{if } t < -\frac{1}{3}.
\end{cases}
\end{aligned}
\end{equation}
The penalty updating functions in \citet{ben1997penalty} are constructed so that in the limit $p \to 0$.
The overall idea of the algorithm is as follows. As in Lagrangian methods, larger dual variables are enforced for violated constraints and smaller ones for satisfied constraints which is achieved by~\eqref{eq:scheme2}. Since $\varphi'$ is strictly increasing with $\varphi'(0)=1$ and $p^k > 0$, we have 
\begin{equation}\label{eq:derivative_phi}
    \varphi'\!\left(g_i(x^{k+1}) / p^k\right)
    \begin{cases}
        > 1 & \text{if } g_i(x^{k+1}) > 0, \\
        = 1 & \text{if } g_i(x^{k+1}) = 0, \\
        \in (0, 1) & \text{if } g_i(x^{k+1}) < 0.
    \end{cases}
\end{equation}
The penalty parameters also govern the speed of change in the dual variables (therefore also the speed of convergence towards the feasible region): the smaller the value of the $i$th coordinate of $p^k$, the larger the update of the $i$th dual variable. This dual update direction is larger either towards $0$ (for feasible iterate $g_i(x^{k+1}) < 0$) or towards unbounded values (for infeasible iterate $g_i(x^{k+1}) > 0$). 





\section{Stochastic Penalty/Barrier Method (SPBM)}
\label{sec:spbm}
Our method, Stochastic Penalty-Barrier Method (SPBM), extends the
penalty and barrier framework of \citet{ben1997penalty} to the
stochastic setting. While the general structure follows their iterative
update rules, we propose three key
modifications. Firstly, to deal with the non-smoothness of the resulting objective, we perform an inexact proximal point update to recover a tractable minimization step. In theory, we assume our oracle to be a Clarke subgradient (denoted as $\partial$ operator); however, for the maximum compatibility and impact, our implementation uses PyTorch (which might not output an element of Clarke \citep{kakade2018provably}). Secondly, the application of the barrier function in the stochastic setting transforms the original problem \eqref{eq:stoctrsto} into a stricter version (proven in \cref{sec:theoretical_main_body}), which motivates different penalty-updating functions compared to the original deterministic one. Lastly, we utilize exponential averaging. In the following paragraphs, we address the main challenges in detail. The full algorithm is presented
in \cref{alg:spbm}.

\paragraph{Mini-batching notation.} We use mini-batching with batch size $B\in\mathbb{N}$ to sample both the objective and the constraint distributions. A mini-batch is a~tuple $\{\xi_b\}_{b=1}^B$ of
i.i.d. random variables $\xi_1,\ldots,\xi_B\iid\Xi$. Stochastic estimates are then computed from the mini-batch as follows:
\begin{equation}\label{eq:minibatching_notation}
    \overline{f}(x) = \frac{1}{B}\sum_{b=1}^Bf(x, \xi_b), \quad \overline{g}(x) = \frac{1}{B}\sum_{b=1}^B g(x, \xi_b).
\end{equation}
Similarly, we denote a stochastic estimate of the Lagrangian $\mathcal{L}_{p}(x, \lambda)$ and its subgradient:
\begin{align}\label{eq:lagrangian_estimate}
    \overline{\mathcal{L}}_{p}(x, \lambda) &= \overline{f}(x)
+ \sum_{i=1}^m \lambda_i p_i \, \varphi (\overline{g}_i(x)/p_i ),
\qquad 
\overline{\nabla} \mathcal{L}_{p}(x, \lambda) \in \partial_x \overline{\mathcal{L}}_{p}(x, \lambda).
\end{align}


\paragraph{Challenge 1: Non-smoothness.}
We treat SPBM as a subgradient method. The intuition is to optimize over the approximate solution of the Moreau envelope of the Lagrangian, with given $p^{k}$, $s^k$, and $\lambda^{k}$:
\begin{equation}\label{eq:moreau_env_prob}
\min_{x}
\left(\overline{\mathcal{L}}_{p^{k}}(x, \lambda^{k}) + \frac{\mu}{2} \| x - s^k\|^2\right).
\end{equation}
In theory, if $\overline{\mathcal{L}}_{p^{k}}(\cdot, \lambda^{k})$ is $\rho$-weakly convex and $\mu > \rho$, its Moreau envelope is a smooth approximation that shares the same minimizers~\citep{beck2017first}; we use this as motivation, noting that weak convexity is not established in our setting.
In practice, we aim to find an approximate minimizer (with no proven guarantees) of the Moreau envelope by performing a single step of a gradient-based primal update with respect to $x$, e.g.\ Adam or SGD ($\PrimalUpdate$ at line 5).
This is the same as performing an inexact proximal-point step on the function $\frac{1}{\mu} \overline{\mathcal{L}}_{p^{k}}(\cdot, \lambda^k)$ \citep{beck2017first}.
We then utilize exponential averaging on the argument $s$ of the proximal operator (line 8, inspired by \citet{zhang2020proximal}, \citet{pmlr-v267-huang25au}). 
\begin{algorithm}[tb]
  \caption{Stochastic Penalty-Barrier Method (SPBM). \PrimalUpdate{} can be one Adam / SGD step; $\mathrm{state}$ carries its internal metadata across iterations as well as learning rate scheduling.}
  \label{alg:spbm}
  \begin{algorithmic}[1]
    \STATE {\bfseries Input:} $x^0\in\mathbb{R}^n$; $\lambda^0\in\mathbb{R}^m_{++}$; $p^0=\mathbf{1} (\text{or } p^0 = \rho \,\lambda^0  \text{ if }\hat\pi = \hat\pi_\rho^{\operatorname{ALM}})$; $s^0=x^0$; $\kappa \in [0, 1) \bigl(\text{alternatively }\rho \in \mathbb{R}_{++}, \text{ if }\hat\pi = \hat\pi_\rho^{\operatorname{ALM}}\bigr)$; Primal optimizer state $\mathrm{state}^0$; batch size $B$;
    $\hat\pi\in\{\hat\pi^{\operatorname{ALM}}_{\rho},
    \hat\pi^{\operatorname{ADAPT}}_{\kappa}\}$, $\gamma \in [0, 1)$, $\mu \geq 0$, and $\delta  \in [0, 1)$; $\Lambda \in \mathbb{R}_{++}$.
    \FOR{$k=0$ {\bfseries to} $K$}
      \STATE Sample $\xi_1,\dots,\xi_B \iid \Xi$
      \STATE $\bar y^{k} = -\overline{\nabla}\mathcal{L}_{p^{k}}(x^k,\lambda^{k}) - \mu(x^k-s^k)$ \hfill\COMMENT{Moreau-augmented gradient, \eqref{eq:moreau_env_prob}}
      \STATE $x^{k+1}, \mathrm{state}^{k+1} = \PrimalUpdate(x^k, \bar y^k, \mathrm{state}^k)$ \hfill\COMMENT{Primal variable step, \eqref{eq:scheme1}}
      \STATE $\lambda^{k+1} = \operatorname{Proj}_{[0, \Lambda]^m}\left[\gamma\lambda^k + (1-\gamma)\,\lambda^k \odot \varphi'(\overline{g}(x^k)\oslash p^k)\right]$ \hfill\COMMENT{Dual variable step, \eqref{eq:scheme2}}
      \STATE $p^{k+1} = \hat\pi(\overline{g}(x^k), p^k, \lambda^{k+1})$ \hfill\COMMENT{Penalty parameter step, \eqref{eq:scheme3};\eqref{penalty_updating_formulas}/\eqref{penalty_updating_formulas_alm}}
      \STATE $s^{k+1} = \delta s^k + (1-\delta)x^{k+1}$ \hfill\COMMENT{Prox-center update}
    \ENDFOR
  \end{algorithmic}
\end{algorithm}

\paragraph{Challenge 2: Penalty updating functions.}


The original penalty updating function in the deterministic iterative scheme \cref{eq:scheme3}
implies that eventually $p\rightarrow 0$. In the stochastic setting, however, such procedure might tighten the original feasible set at a cost of a worse objective value; on the other hand, it introduces a controllable bias for constraints that are difficult to satisfy (details in \cref{sec:theoretical_main_body}). Therefore, we propose an adaptive penalty-updating formula $\hat{\pi}: \mathbb{R}^m \times \mathbb{R}^m \times \mathbb{R}^m \to \mathbb{R}^m$, which adaptively decreases penalties only for unsatisfied constraints, while maintaining larger penalties for satisfied constraints. Otherwise, the motivation behind penalty parameters described in PBM \cref{sec:recspbm} consists. The adaptive formula reads as follows: 
\begin{equation}\label{penalty_updating_formulas}
    \hat{\pi}^{\operatorname{ADAPT}}_{\kappa}(\bar{g}, p, \lambda)
    = \operatorname{Proj}_{[\ell_c, u_c]^m} \left[ \left(\kappa + \frac{ (1-\kappa) }{\varphi'(\bar{g}) + \varepsilon}\right) p \right], \quad  \varepsilon = 10^{-8},
\end{equation}
where all operations (sum, product, division) are performed coordinate-wise, and $\operatorname{Proj}_{[\ell_c, u_c]^m}$ denotes the coordinate-wise clipping onto a box $[\ell_c, u_c]^m$ for $0 < \ell_c \leq u_c$. 
As a second choice of the penalty-updating formula, we suggest
\begin{equation}\label{penalty_updating_formulas_alm}
\hat \pi^{\operatorname{ALM}}_{\rho}(\bar{g}, p, \lambda) = \operatorname{Proj}_{[\ell_c, u_c]^m} \left[ \lambda\cdot \rho \right ], \quad \text{where } \rho > 0.
\end{equation}
Interestingly, as shown in Appendix~\ref{sec:equivalence_to_alm}, for \cref{penalty_updating_formulas_alm}, SPBM reduces to SSL-ALM-like primal-dual update. Note \emph{(i)} by choosing $\cref{penalty_updating_formulas_alm}$, the hyperparameter $\kappa$ becomes redundant, same applies to $\rho$ in case of \cref{penalty_updating_formulas}; \emph{(ii)} clipping safeguards are proposed for purposes of a controllable bias (\cref{sec:theoretical_main_body}); \emph{(iii)} hyperparameters $\kappa, \rho$ can be thought of as step sizes of the penalty parameters. 

\subsection{Theoretical Analysis of SPBM}\label{sec:theoretical_main_body}

In the deterministic setting, it is immediate that the original formulation
\eqref{eq:convexprob} and its transformed counterpart \eqref{eq:convexprobtransformed}
define the same problem: the reformulation \eqref{eq:convexprobtransformed} is exact. In the
stochastic setting, however, SPBM evaluates the
penalty/barrier function on mini-batch estimates $\bar{g}_i$ rather than on the exact
expectations, and therefore implicitly targets the problem
\begin{equation}\label{eq:stoctrsto_transformed}
    \min_{x\in\mathbb{R}^n} \; \mathbb{E}_\xi[f(x, \xi)]
    \quad \text{s.t.} \quad
    \mathbb{E}_{\xi_{1:B}}\bigl[ p_i \, \varphi\bigl( \bar{g}_i(x) / p_i \bigr) \bigr] \leq 0,
    \quad i = 1, \dots, m,
\end{equation}
instead of \cref{eq:stoctrsto}. Here $\bar{g}_i$ is the mini-batch estimator defined
in \cref{eq:minibatching_notation} with $\xi_1, \dots, \xi_B \overset{\text{iid}}{\sim} \Xi$,
and $\mathbb{E}_{\xi_{1:B}}$ denotes the expectation over the mini-batch; the
objective is unaffected, since $\bar{f}$ is an unbiased estimator of
$\mathbb{E}_{\xi}[f(x,\xi)]$ and enters SPBM linearly. The constraints, in contrast,
are distorted: since $\varphi$ is convex and non-linear by definition, Jensen's
inequality~\citep{jensen1906fonctions} gives, for any fixed $p_i > 0$,
\begin{equation}\label{eq:jensen_constraint}
   p_i \, \varphi \bigl( \mathbb{E}_{\xi_{1:B}}[\bar{g}_i(x)] / p_i \bigr)
   \;\leq\;
   \mathbb{E}_{\xi_{1:B}} \bigl[ p_i \, \varphi\bigl( \bar{g}_i(x) / p_i \bigr) \bigr].
\end{equation}
The difference between the right- and left-hand sides of
\cref{eq:jensen_constraint} is known as the Jensen
gap~\citep{strukov1977mathematical, gao2017bounds, abramovich2016some}. Its
non-negativity means that the constraints in \cref{eq:stoctrsto_transformed} are at
least as strict as those of the original problem~\eqref{eq:stoctrsto}: any point
feasible for the transformed problem is feasible for the original one. This is formalized in \cref{theorem:feasible_subset}:
\begin{restatable}[Transformed-feasibility gives original-feasibility]{proposition}{theoremFeasibleTransformed}\label{theorem:feasible_subset}
Consider a penalty/barrier function $\varphi \in \{\varphi_{\mathrm{QL}}, \varphi_{\mathrm{QR}}\}$ \eqref{eq:penalty_barrier_functions}, a positive integer $B\in\mathbb{N}$, and random variables $\xi_1, \dots, \xi_B\overset{\text{iid}}{\sim} \Xi$.
Consider the feasible set of the original problem~\eqref{eq:stoctrsto} and of the transformed problem \eqref{eq:stoctrsto_transformed}, defined as
$$
   \mathcal{S} := \bigcap_{i = 1}^m \bigl\{ x \in \mathbb{R}^n : \mathbb{E}_{\xi}\bigl[g_i(x, \xi)\bigr] \leq 0\bigr\}, 
   \quad
\mathcal{S}^B_{\operatorname{transform}} := \bigcap_{i = 1}^m \bigl\{ x \in \mathbb{R}^n :
\mathbb{E}_{\xi_{1:B}} \bigl[ p_i \varphi \bigl( \bar{g}_i(x) / p_i \bigr) \bigr] \leq 0\bigr\}.
$$
Then, $\mathcal{S}^B_{\operatorname{transform}} \subseteq \mathcal{S}$.\unskip\iftoggle{restatingtheorem}{}{\hfill\textit{(For proof see Appendix~\ref{proof:feasible_subset}.)}}
\end{restatable}

The next proposition bounds the corresponding Jensen gap (see also Appendix~\ref{proof:feasible_subset} for proof).
\begin{restatable}[Jensen gap bounds]{proposition}{theoremJensenGap}\label{theorem:jensengap}
Let $\varphi \in \{\varphi_{\mathrm{QL}}, \varphi_{\mathrm{QR}}\}$ be a penalty/barrier
function as defined in \cref{eq:penalty_barrier_functions}, let $B \in \mathbb{N}$ be a positive integer, let
$\xi_1, \dots, \xi_B \overset{\text{iid}}{\sim} \Xi$, and assume
$\operatorname{Var}(g_i(x,\xi)) < \infty$. Then, for every $x \in \mathbb{R}^n$,
$i \in \{1, \dots, m\}$, and $p_i > 0$,
$$ \mathbb{E}_{\xi_{1:B}} \bigl[ p_i \varphi\bigl( \bar{g}_i(x) / p_i \bigr) \bigr]
   - p_i \varphi \bigl( \mathbb{E}[ g_i(x, \xi)] / p_i \bigr)
   \;\leq\; \frac{\operatorname{Var}\bigl( g_i(x, \xi) \bigr)}{2 B p_i}. $$
\end{restatable}

Two consequences follow. Firstly, as $B \to \infty$,
the gap vanishes and the transformed constraints recover the original ones; the bias
is thus a finite-batch effect, decaying at rate $1/B$. Secondly, as $p_i \to 0$, the
gap grows without bound. This is in direct tension with the penalty
schedule \eqref{eq:scheme3} of PBM, which drives
$p \to 0$, and justifies our modified update \eqref{penalty_updating_formulas},
whose clipping to $[\ell_c, u_c]$ keeps the bias bounded by
$\operatorname{Var}(g_i(x,\xi))/(2 B \ell_c)$ uniformly over iterations. The bound also
guides the choice of the lower clipping bound $\ell_c$: when SPBM struggles to reach
feasibility, decreasing $\ell_c$ tightens the transformed constraints (introduces a controllable bias) and enlarges the
dual updates of violated constraints, driving the
iterates toward $\mathcal{S}^B_{\operatorname{transform}}$ more aggressively -
and, by \cref{theorem:feasible_subset}, any point of
$\mathcal{S}^B_{\operatorname{transform}}$ is feasible for the original problem.
There is a price for this: a smaller $\ell_c$ shrinks the surrogate feasible set
and may exclude solutions of \cref{eq:stoctrsto} with lower objective value.

\subsection{Convergence Analysis}

Consider the two-parameter stochastic fitting problem of Appendix~\ref{sec:counter_example}: a single ReLU unit under a fairness constraint. We show it is an instance of \cref{eq:stoctrsto} (\cref{prop:counterexample_is_stosto}) but not of the weakly convex setting (\cref{prop:weakly_convex_setting}); so related work assuming weak convexity excludes even a single-ReLU ERM task. We then prove that both state-of-the-art CML methods and SPBM make no progress from an initialisation that can be arbitrarily far from the feasible region (proof can be found in Appendix~\ref{sec:counter_example}).
\begin{proposition}[No progress on the instance of Appendix~\ref{sec:counter_example}]\label{cor:no_methods_converge}
    Assume the problem described in Appendix~\ref{sec:counter_example}, with $x^k = w^k + b^k$ and $\theta^k = (w^k, b^k)$. Then SPBM, SSw, and SSL-ALM, and any method with an update scheme of type $\theta^{k+1} = \theta^k + h( \theta^k )$ for $h( \theta^k ) = 0$ whenever $\partial  \bar f(\theta^k) = \partial \bar g( \theta^k ) = \{ 0 \}$, fail to reach the feasible region for preactivation $x^0 < 0$ for any realisation. Moreover, it holds for any realisation and any $k\geq 0$
    \[
    \operatorname{dist}(x^k, \{ x: g(x) \leq 0 \} ) = \operatorname{dist}(x^0, \{ x: g(x) \leq 0 \}).
    \]
\end{proposition}


We therefore aim at a weaker, but non-empty, guarantee. For this, we require additional assumptions which are described in detail in \cref{alg:spbm_schedules} and informally here.
Let the dataset have size $M$, and the batch size be $B$, and define the function $\mathcal{L}^{\mathrm{emp}}_{p}(\cdot,\lambda) : \mathbb{R}^n \to \mathbb{R}$, and a conservative field of $-\mathcal{L}^{\mathrm{emp}}_{p}(\cdot,\lambda)$ denoted as $\mathcal{D}_{p}(\cdot, \lambda) : \mathbb{R}^n \rightrightarrows \mathbb{R}^n$ such that
\begin{equation}
\label{eq:auglagfield}
\mathcal{L}^{\mathrm{emp}}_{p}(\cdot,\lambda)
= \mathbb{E}_{\xi_{1:B}}\big[\bar{\mathcal{L}}_{p}(\cdot,\lambda)\big], \quad \mathcal{D}_{p}(\cdot, \lambda) = -\frac{1}{M^B} \sum_{\mathbf{j} \in \{ 1, \dots, M\}^B} \partial_x \overline{\mathcal{L}}_{p}(\cdot, \lambda; \mathbf{j}),
\end{equation}
where $\overline{\mathcal{L}}_{p}(\cdot, \lambda; \mathbf{j})$ is the estimate dependent on batch draw $\mathbf{j}$. We refer the 
reader to Appendix~\ref{conservative_fields_subsection} for background on conservative fields. Informally, the mild convergence result reads as follows: 
\begin{theorem}[Primal criticality -- informal statement]\label{main_theorem_approx}
Assume definable and locally Lipschitz $\bar g, \bar f$, almost sure boundedness of the primal variable.
Run SPBM with $\gamma^k, \kappa^k \to 1$ with summable deviations, square-summable but not summable primal schedule, dual clipping onto $[0, \Lambda]^m$, $\bar y^k$ being an element of Clarke, SGD dynamics (\cref{alg:spbm_schedules}). Then almost
surely there exist random $\hat \lambda \in [0, \Lambda]^m, \hat p \in [\ell_c, u_c]^m$ such that
    \emph{(1)} $\lambda^k\to\hat\lambda$, $p^k\to\hat p$;
    \emph{(2)} $\|x^k-s^k\|\to0$;
    \emph{(3)} the function values $\{\mathcal{L}^{\mathrm{emp}}_{\hat p}(x^k, \hat \lambda) \}_{k \geq 0}$ converge;
    \emph{(4)} every limit point $\hat x$ of $\{ x^k \}_{k \geq 0}$ satisfies $0 \in \mathcal{D}_{\hat p}(\hat x, \hat \lambda)$.
\end{theorem}
\begin{proof}[Proof sketch (full proof in Appendix~\ref{sec:convergence_analysis})]
SPBM is a primal-dual algorithm.
As such, it performs descent on the primal variables, and ascent on the dual variables of the 
penalised Lagrangian $\mathcal{L}^{\mathrm{emp}}_{p^k}(\cdot,\lambda^k)$.
This is the main difficulty in obtaining results for SPBM, and for primal-dual methods more generally.
We consider the general setting where the objective function and constraints are neither differentiable nor (weakly)-convex.
We build on the results obtained by \citet{DacDruKakLee2018,bolte2021conservative} for SGD in such a general setting.
Their result would provide primal variable convergence were the dual variables and penalty parameters fixed.

We fill this gap by assuming 
summability of $(1-\gamma^k)$ and $(1-\kappa^k)$.
This gives the dual and penalty paths finite length, so they admit limit points $\hat\lambda$ and $\hat p$ almost surely.
A central observation is that fixing the dual and penalty variables reduces SPBM to SGD, which allows applying the above mentioned convergence guarantees.
Therefore, SPBM with dual and penalty variables fixed to their limiting values converges to critical points of  $\mathcal{L}^{\mathrm{emp}}_{\hat p}(\cdot,\hat\lambda)$.
The rest of the proof consists in bounding the distance between the dual and penalty iterates of the actual SPBM scheme and the SPBM scheme with fixed dual and penalty variables.
Along the way, we justify that the conservative field associated with the penalty barrier, defined in \cref{eq:auglagfield}, is definable in the same structure as the objective and constraint functions, which allows applying the standard SGD guarantees.
\end{proof}

    The function $\mathcal{L}^{\mathrm{emp}}_{p}(\cdot,\lambda)$ is the Lagrangian of the
    transformed problem \eqref{eq:stoctrsto_transformed}, so item \emph{(4)} says that $\hat x$ is
    stationary for \eqref{eq:stoctrsto_transformed} rather than for the original problem. This is informative in two ways: The feasible set of \eqref{eq:stoctrsto_transformed} lies inside
    $\mathcal{S}$ by \cref{theorem:feasible_subset}, so a point that is feasible for the
    transformed problem is feasible for the original one, and by \cref{theorem:jensengap}
    the gap is at most $\operatorname{Var}(g_i(x,\xi))/(2B\ell_c)$. 
    By the Clarke sum rule $- \partial \mathcal{L}_{\hat p}(\hat x, \hat \lambda) \subseteq \mathcal{D}_{\hat p}(\hat x, \hat \lambda)$, hence \emph{(4)} is implied by, but does not imply Clarke criticality. As already discussed, the theorem \emph{does not} provide any primal feasibility or complementarity guarantees.
  Full analysis is provided in Appendix~\ref{sec:convergence_analysis}. \cref{alg:spbm_schedules} is used in the analysis, whereas \cref{alg:spbm} is used for benchmarking.

\section{Numerical Experiments}
\label{sec:numexp}
\paragraph{Overview.}
We create a benchmark comparing SPBM to three other optimization methods: Adam~\citep{kingma2014adam} (baseline with no constraints\footnote{For PINNs, we use Adam to optimize the network as suggested in the original paper \citep{raissi2017physics} with a fixed regularization term (a hyperparameter) of the initial and boundary constraints. }), SSw~\citep{HuaLin2023}, and SSL-ALM \citep{pmlr-v267-huang25au}.
Our benchmark contains 5 classification problems based on the ACSIncome dataset (\texttt{folktables}; binary classification), the Dutch Census dataset (binary classification), and the CIFAR-10 and CIFAR-100 datasets (multiclass classification), and three experiments involving Physics-Informed Neural Networks (PINNs). We benchmark the performance of each method on the training data of the datasets. We present the constraints of each problem below.

\paragraph{Fairness and Regularization as constrained optimization.}\label{sec:paragraph_fairness} In classification experiments, we implement two types of constraints. The first type is deterministic and it enforces weights of a neural network to lie in an $\varepsilon$ ball:
\begin{equation}\label{eq:weight_regularization}
\quad \|W_i\|_F \le \varepsilon, \quad \text{ for each layer }i,
\end{equation}
where $W_i$ is a weight matrix of the fully connected neural network and $\|\cdot\|_F$ is the Frobenius norm. The other constraint is stochastic and its goal is to enforce a similar prediction accuracy across different groups -- either demographic groups (\emph{e.g.} sex), or classes in the CIFAR datasets. 
%
We model this by constraining differences between all pairs of groups:
\begin{align}
&-\varepsilon \leq \mathbb{E}_{\xi \sim \text{group }R_i} [ f(x, \xi) ]  - \mathbb{E}_{\xi \sim  \text{group }R_j} [ f(x, \xi) ]
\le \varepsilon, \quad \text{for every } i\neq j,\label{eq:eqop}
\end{align}
 
 which results, for a set of all demographic groups $\mathcal{R}$, in $m=|\mathcal{R}|\cdot(|\mathcal{R}|-1)$ constraints. 


\paragraph{Solving PDEs as constrained optimization.}\label{PINNs_para} PINNs are neural networks trained to satisfy 
physical laws governed by nonlinear partial differential equations (PDEs). An example of such a~physical law can be the Helmholtz equation:
\begin{align}\label{helmholtz_loss}
    &\Delta u + u = q(z_1,z_2), &&\text{ for every } (z_1, z_2) \in \mathcal{Z}, \\\label{helmholtz_constraint}
    &u(z_1,z_2) = 0, &&\text{ for every } (z_1,z_2) \in \operatorname{Bd} \mathcal{Z}, 
\end{align}
where $\mathcal{Z} = [-1,1]^2 \subseteq \mathbb{R}^2$ and $\operatorname{Bd}\mathcal{Z}$ is its
boundary, and the function $q$ is fixed (see Appendix~\ref{sec:PINNs_appendix}). Since boundary conditions 
are known to cause instability in PINN training \citep{wang2022and, 
son2023enhanced, wang2021understanding}, it is natural to enforce 
them as constraints, yielding the constrained optimization problem:
\begin{align}\label{eq:pinns_opt_problem_loss}
\min_{x} \quad & \mathbb{E}_{\xi \sim \Xi_{\mathcal{Z}}}\!\left[|N u_{\text{nn}}(x,\xi) - h(\xi)|^2\right]\\
\label{eq:pinns_opt_problem_constr}
\text{s.t.} \quad
& \mathbb{E}_{\xi \sim \Xi_{\Gamma(i)}}\!\left[|N_i u_{\text{nn}}(x,\xi) - h_i(\xi)|^2\right] \le \varepsilon,
\quad \text{for each } i \in \{1,\dots,m\},
\end{align}
where $\varepsilon>0$ is small, $N$ and $N_i$ are differential operators, $u_{\text{nn}}$ is a
neural network representing the solution with weights $x$, and $h, h_i$ are prescribed.
Each $\Gamma(i) \subseteq \operatorname{Bd}\mathcal{Z}$ is the set on which the $i$th
condition is imposed; $\Xi_{\mathcal{Z}}$ and $\Xi_{\Gamma(i)}$ denote the corresponding
distributions of collocation points.

\paragraph{Experiments.} We denote our experiments by \Exp{1}--\Exp{8} and summarize them in \Cref{tab:experiments} together with the demographic attributes used in the fairness experiments. \Exp{6}, \Exp{7}, and \Exp{8} are based on the Helmholtz equation, viscous Burgers equation and Klein-Gordon equation, respectively; see Appendix~\ref{appendix:experimental_setup} for details on the experimental setup such as the datasets, data processing, implementation details of methods, network architectures, choice of hyperparameters, number of seeds, and hardware.

\begin{table}[t]
\centering
\caption{Summary of experiments; MSE stands for Mean Squared Error.}
\label{tab:experiments}
\resizebox{0.99\textwidth}{!}{
\begin{tabular}{l l l l l l l}
\toprule
Exp. & Dataset & Objective function $f$ & Groups $\mathcal{R}$ & Constraint $g$ & \# Constraints ($m$) & Bounds $\varepsilon$\\
\midrule
\Exp{1} & ACSIncome & binary cross-entropy & -- & Bound constraint \cref{eq:weight_regularization}  & 6 &  $2$  \\
\Exp{2} & ACSIncome & binary cross-entropy & 6 (sex $\times$ marital) & \cref{eq:eqop} & 30  &  $10^{-1}$  \\
\Exp{3} & Dutch Census & binary cross-entropy & 18 (sex $\times$ age) & \cref{eq:eqop} & 306 & $10^{-1}$  \\
\Exp{4} & CIFAR-10 & multiclass cross-entropy & 10 (classes) & \cref{eq:eqop} & 90 &  $10^{-1}$  \\
\Exp{5} & CIFAR-100 & multiclass cross-entropy & 100 (classes) & \cref{eq:eqop} & 9900 &  $10^{-1}$  \\
\Exp{6} & Sampled $[-1,1]^2$ & MSE \cref{eq:pinns_opt_problem_loss} with \cref{helmholtz_loss} & -- & \cref{eq:pinns_opt_problem_constr} with \cref{helmholtz_constraint}& 1 & $10^{-4}$ \\
\Exp{7} & Sampled $[0,1]\times [-1,1]$ & MSE \cref{eq:pinns_opt_problem_loss} with \cref{burgers_equation} & -- & \cref{eq:pinns_opt_problem_constr} with \cref{burgers_constr1}, \cref{burgers_constr2} & 2 & $10^{-4}$\\
\Exp{8} & Sampled $[0,1]^2$ & MSE \cref{eq:pinns_opt_problem_loss} with \cref{Klein_Gordon_equation}& -- & \cref{eq:pinns_opt_problem_constr} with \cref{Klein_Gordon_constr1}, \cref{Klein_Gordon_constr2}, \cref{Klein_Gordon_constr3} & 3 & $10^{-4}$\\
\bottomrule
\end{tabular}
}
\end{table}
\subsection{Methodology}\label{subsec:methodology}
\paragraph{Moré--Wild profiles.} Inspired by \citet{more2009benchmarking}, we measure performance across hyperparameter configurations rather than across problems. Let $\mathcal{E}$ be
the set of experiments, $\mathcal{M}$ the set of methods, and
$\mathcal{H}(\mathrm{mt})$ the hyperparameter grid of method $\mathrm{mt}$. Let
$\mathcal{Q}(\mathrm{ex},\mathrm{mt},\mathrm{hp}) \in \mathbb{R}$
denote a metric; in our case training loss or constraint violation. Write $ \mathcal{Q}^\star(\mathrm{ex}) := \min_{\mathrm{mt} \in \mathcal{M},\, \mathrm{hp} \in \mathcal{H}(\mathrm{mt})}\mathcal{Q}(\mathrm{ex},\mathrm{mt},\mathrm{hp})$, and denote the metric at initialization $\mathcal{Q}^0(\mathrm{ex})$, which is the same for all methods. For each triple we define the score:
\[
  \mathrm{score}(\mathrm{ex},\mathrm{mt},\mathrm{hp}) :=
    \max \!\left(
      \dfrac{\mathcal{Q}(\mathrm{ex},\mathrm{mt},\mathrm{hp})
             - \mathcal{Q}^\star(\mathrm{ex})}
            {\mathcal{Q}^0(\mathrm{ex}) - \mathcal{Q}^\star(\mathrm{ex})},
      \; 0 \right).
\]
For each method  $\mathrm{mt} \in \mathcal{M}$ we plot the empirical CDF:
\[
  \mathrm{cdf}_{\mathrm{mt}}(\tau) :=
    \frac{1}{|\mathcal{E}|\cdot|\mathcal{H}(\mathrm{mt})|}
    \Big|\big\{(\mathrm{ex},\mathrm{hp})
      \in \mathcal{E}\times\mathcal{H}(\mathrm{mt})
      \;:\; \mathrm{score}(\mathrm{ex},\mathrm{mt},\mathrm{hp}) \le \tau \big\}\Big| ,
    \qquad \tau \in \mathbb{R}_{+}.
\]
In other words, a configuration passes the $\tau$-test at $\tau = 0$ only if it matches the best value reached by any method on that experiment;  all non-divergent configurations pass by $\tau = 1$. Therefore, a CDF that rises earlier and higher indicates a method that is more robust to the hyper-parameter choice. 
\begin{figure}[t]
    \caption{Moré-Wild profiles over the hyperparameter grid. For each method,
    the curve gives the fraction of hyperparameters whose relative score is at most $\tau$. A score of $0$ means the run matched the best value reached by any method, and a score of $1$ means it made no progress from initialisation. Earlier and higher curve indicates a method that is more robust to the hyper-parameter choice. See \cref{subsec:methodology} for the definition. Loss metric is plotted only for feasible runs; dashed line indicates feasibility within $1\%$ tolerance, solid line indicates feasibility within $10\%$ tolerance. Feasibility is defined as $[ \max_{i \in \{1,\dots,m\}} \bar g_i - \varepsilon]_+$, where $\bar g_i$ here is averaged over all seeds.}
    \centering
    \begin{subfigure}{0.49\linewidth}
        \centering
        \includegraphics[width=\linewidth]{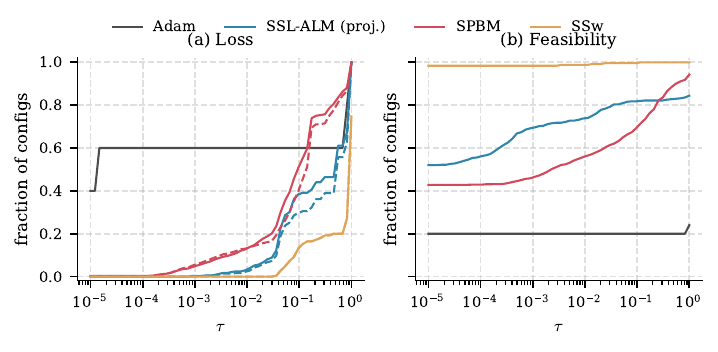}
        \caption{\Exp{2,3,4,5}: Moré--Wild profile.}
        \label{fig:fair_moorewild}
    \end{subfigure}
    \hfill
    \begin{subfigure}{0.49\linewidth}
        \centering
        \includegraphics[width=\linewidth]{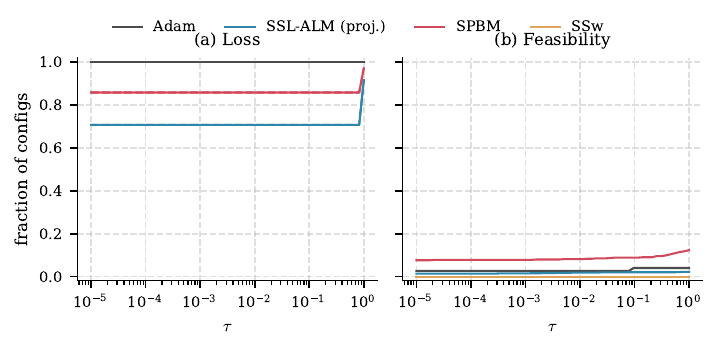}
        \caption{\Exp{6,7,8}: Moré--Wild profile.}
        \label{fig:pinns_moorewild}
    \end{subfigure}
    \label{fig:more_wild_plots}
\end{figure}

\paragraph{Best configurations.} We report a table of the best run per method over the hyperparameter grid. Selection is based on the tail-mean over the last $N$ epochs  ($N=5$ for fairness; $N=50$ for PINN experiments) in the following way: we take a configuration feasible within a $10\%$ tolerance with the lowest training loss; if no feasible configuration is available, we choose the least infeasible one. The full hyperparameter grids and the final hyperparameter choices are listed in Appendix \ref{appendix:experimental_setup}. We also provide a computational benchmark based on \Exp{5} with a varying number of constraints. Finally, we provide convergence plots in
Appendix~\ref{subsec:convergence_plots} with an ablation study of SPBM in
Appendix~\ref{subsec:ablation_study}.
\newpage
\subsection{Results}\label{sec:analysis_experiments}
Two observations emerge from \cref{tab:best_results}. First, SPBM clearly outperforms SSL-ALM on \Exp{4,5}, the two multiclass experiments, which include the largest task ($m=9900$); on \Exp{8} it attains a similar loss with better constraint satisfaction; and it matches SSL-ALM on the remaining experiments. Second, SPBM also improves on the unconstrained Adam baseline in \Exp{4}. We conjecture that the constraints prevent the optimizer from concentrating on the easier classes, which is supported by the accuracy curves in \cref{fig:e4_convergence} (Appendix~\ref{subsec:convergence_plots}).

\cref{fig:more_wild_plots} gives more detail on robustness. A larger fraction of SPBM's configurations attains a low loss than SSL-ALM's on both fairness and PINNs tasks; SSL-ALM compensates with more feasible configurations on the fairness tasks, while SPBM leads on both counts for PINNs. SPBM has more hyperparameters than SSL-ALM, so its grid is larger: 360 vs. 100 configurations on the fairness problems and 384 vs. 360 on the PINN problems.
Lastly, we observe from \cref{tab:cifar100_runtime} that the per-epoch runtime of CML methods is within $1.3\times$ of the per-epoch runtime of regularized Adam. 
\begin{table}[t]
\centering
\caption{Final results on our benchmark for SPBM, SSL-ALM, SSw, and Adam
(unconstrained baseline). For each method and experiment we report the
best-performing hyperparameter configuration from our grid, selected as the
feasible run with the lowest training loss; see \cref{subsec:methodology} for the
selection rule. Values are means $\pm$ standard deviation over $5$ seeds. \textbf{Bold} marks the best value and \underline{underline} the second best within each experiment. We omit marking the unconstrained Adam. Feasibility is defined as $[ \max_{i \in \{1,\dots,m\}} \bar g_i - \varepsilon]_+$, where $\bar g_i$ here is averaged over all seeds.}
\label{tab:best_results}
\small
\setlength{\tabcolsep}{4pt}
\newcommand{\sd}[1]{\scriptsize{$\pm$ #1}}
\begin{tabular}{@{}c@{\hspace{0.3em}}c@{}}
\begin{tabular}{l l r r}
\toprule
Exp. & Method & Best loss & Feasibility \\
\midrule
\multirow{4}{*}{\Exp{1}}
 & Adam & 0.186 \sd{0.004} & 52.39 \sd{1.1} \\
 & SPBM & \textbf{0.373} \sd{0.001} & \textbf{0.000} \sd{0.003} \\
 & SSL-ALM & \textbf{0.373} \sd{0.000} & \textbf{0.000} \sd{0.000} \\
 & SSw & 0.383 \sd{0.007} & \textbf{0.000} \sd{0.035} \\
\midrule
\multirow{4}{*}{\Exp{2}}
 & Adam & 0.305 \sd{0.002} & 0.210 \sd{0.002} \\
 & SPBM & \textbf{0.350} \sd{0.003} & \underline{0.002} \sd{0.007} \\
 & SSL-ALM & \underline{0.351} \sd{0.003} & \textbf{0.000} \sd{0.006} \\
 & SSw & 0.623 \sd{0.013} & \textbf{0.000} \sd{0.006} \\
\midrule
\multirow{4}{*}{\Exp{3}}
 & Adam & 0.394 \sd{0.000} & 0.497 \sd{0.001} \\
 & SPBM & \textbf{0.470} \sd{0.002} & 0.006 \sd{0.005} \\
 & SSL-ALM & \underline{0.478} \sd{0.004} & \underline{0.003} \sd{0.012} \\
 & SSw & 0.669 \sd{0.011} & \textbf{0.000} \sd{0.009} \\
\midrule
\multirow{4}{*}{\Exp{4}}
 & Adam & 0.161 \sd{0.036} & 0.167 \sd{0.054} \\
 & SPBM & \textbf{0.016} \sd{0.004} & \textbf{0.000} \sd{0.013} \\
 & SSL-ALM & \underline{1.380} \sd{0.025} & \textbf{0.000} \sd{0.006} \\
 & SSw & 2.203 \sd{0.014} & \textbf{0.000} \sd{0.020} \\
\bottomrule
\end{tabular}
&
\begin{tabular}{l l r r}
\toprule
Exp. & Method & Best loss & Feasibility\\
\midrule
\multirow{4}{*}{\Exp{5}}
 & Adam & 0.000 \sd{0.000} & 0.000 \sd{0.002} \\
 & SPBM & \textbf{0.001} \sd{0.000} & \textbf{0.000} \sd{0.002} \\
 & SSL-ALM & \underline{4.577} \sd{0.004} & \textbf{0.000} \sd{0.009} \\
 & SSw & 4.587 \sd{0.002} & \textbf{0.000} \sd{0.008} \\
\midrule
\multirow{4}{*}{\Exp{6}}
 & Adam (reg.) & 0.153 \sd{0.032} & 0.0085 \sd{0.0064} \\
 & SPBM & \textbf{4010.463} \sd{3274.490} & \underline{0.0001} \sd{0.0002} \\
 & SSL-ALM & \underline{5347.374} \sd{2673.665} & \textbf{0.0000} \sd{0.0002} \\
 & SSw & 6460.98 \sd{446.31} & 0.0025 \sd{0.0047} \\
\midrule
\multirow{4}{*}{\Exp{7}}
 & Adam (reg.)  & 0.000 \sd{0.000} & 0.4901 \sd{0.0000} \\
 & SPBM & \textbf{0.000} \sd{0.000} & \textbf{0.0000} \sd{0.0000} \\
 & SSL-ALM & \textbf{0.000} \sd{0.000} & \textbf{0.0000} \sd{0.0001} \\
 & SSw & \textbf{0.000} \sd{0.000} & \underline{0.4945} \sd{0.0037} \\
\midrule
\multirow{4}{*}{\Exp{8}}
 & Adam (reg.)  & 14.4 \sd{24.6} & 79.907 \sd{158.29} \\
 & SPBM & \underline{0.252} \sd{0.109} & \textbf{0.0002} \sd{0.0004} \\
 & SSL-ALM & \textbf{0.243} \sd{0.141} & \underline{0.0014} \sd{0.0013} \\
 & SSw & 10196.38 \sd{389.67} & 0.3459 \sd{0.0309} \\
\bottomrule
\end{tabular}
\end{tabular}
\end{table}

\begin{table}[H]
\centering
\small
\setlength{\tabcolsep}{3.0pt}
\begin{minipage}[c]{0.56\linewidth}\vspace{0pt}
\centering
\begin{tabular}{@{}lccc@{}}
\toprule
 Method & $m{=}100$ & $m{=}5000$ & $m{=}9900$ \\
\midrule
Adam        & 1.74 $\pm$ 0.10 & 1.71 $\pm$ 0.00 & 1.71 $\pm$ 0.00 \\
Adam + Reg. & 1.79 $\pm$ 0.09 & 1.82 $\pm$ 0.03 & 1.84 $\pm$ 0.02 \\
SPBM        & 2.20 $\pm$ 0.01 & 2.26 $\pm$ 0.01 & 2.26 $\pm$ 0.01 \\
SSL-ALM     & 1.98 $\pm$ 0.01 & 1.99 $\pm$ 0.01 & 1.98 $\pm$ 0.01 \\
SSw         & 1.98 $\pm$ 0.09 & 1.96 $\pm$ 0.03 & 1.96 $\pm$ 0.02 \\
\bottomrule
\end{tabular}
\end{minipage}
\hfill
\begin{minipage}[c]{0.40\linewidth}
\caption{Computational benchmark on \Exp{5}: per-epoch runtime (s), mean $\pm$
std over epochs and seeds. Runtime is essentially constant in the number of
constraints $m$.}
\label{tab:cifar100_runtime}
\end{minipage}
\end{table}

\textbf{Conclusion.} We have aimed at partially bridging the gap between theory and practice of constrained machine learning. We proposed and analysed SPBM, which matches or outperforms the state of the art across fairness and PINN benchmarks. We also exhibited an ERM instance on which SPBM and state-of-the-art methods fail: evidence that current assumptions are too strong for realistic CML. 

\textbf{Limitations.} SPBM has more hyperparameters than SSL-ALM, though a comparable or higher fraction of configurations performs well. Our convergence result is partial: it gives criticality of the transformed Lagrangian but no feasibility or complementarity. What assumptions are weak enough to cover general CML yet strong enough for a KKT guarantee remains open.

\bibliography{biblio}

@article{elenterNearOptimalSolutionsConstrained2024,
  title = {Near-{{Optimal Solutions}} of {{Constrained Learning Problems}}},
  author = {Elenter, Juan and Chamon, Luiz and Ribeiro, Alejandro},
  year = 2024,
  month = may,
  journal = {International Conference on Learning Representations},
  volume = {2024},
  pages = {17426--17450},
  urldate = {2026-09-14},
  langid = {english},
}

@inproceedings{hounieResilientConstrainedLearning2023,
  title = {Resilient {{Constrained Learning}}},
  booktitle = {Advances in {{Neural Information Processing Systems}}},
  author = {Hounie, Ignacio and Ribeiro, Alejandro and Chamon, Luiz F. O.},
  year = 2023,
  volume = {36},
  pages = {71767--71798},
  publisher = {Curran Associates, Inc.},
  doi = {10.52202/075280-3142},
  urldate = {2026-09-14},
}

@inproceedings{
kliachkin2026benchmarking,
title={Benchmarking Stochastic Approximation Algorithms for Fairness-Constrained Training of Deep Neural Networks},
author={Andrii Kliachkin and Jana Lep{\v{s}}ov{\'a} and Gilles Bareilles and Jakub Marecek},
booktitle={The Fourteenth International Conference on Learning Representations},
year={2026},
url={https://openreview.net/forum?id=JxmjzC6syB}
}

@article{nekvindaSimpleProofRademacher1988,
  title = {A Simple Proof of the {{Rademacher}} Theorem},
  author = {Nekvinda, Ale{\v s} and Zaj{\'i}{\v c}ek, Lud{\v e}k},
  year = {1988},
  journal = {{\v C}asopis pro p{\v e}stov{\'a}n{\'i} matematiky},
  volume = {113},
  number = {4},
  pages = {337--341},
  publisher = {Mathematical Institute of the Czechoslovak Academy of Sciences},
  issn = {0528-2195},
  urldate = {2025-09-08},
  langid = {english},
}

@article{van1996geometric,
  title={Geometric categories and o-minimal structures},
  author={Van den Dries, Lou and Miller, Chris},
  year={1996}
}

@article{more2009benchmarking,
  title={Benchmarking derivative-free optimization algorithms},
  author={Mor{\'e}, Jorge J and Wild, Stefan M},
  journal={SIAM Journal on Optimization},
  volume={20},
  number={1},
  pages={172--191},
  year={2009},
  publisher={SIAM}
}

@article{bianchi2022convergence,
  title={Convergence of constant step stochastic gradient descent for non-smooth non-convex functions},
  author={Bianchi, Pascal and Hachem, Walid and Schechtman, Sholom},
  journal={Set-Valued and Variational Analysis},
  volume={30},
  number={3},
  pages={1117--1147},
  year={2022},
  publisher={Springer}
}

@article{bolte2023subgradient,
  title={Subgradient sampling for nonsmooth nonconvex minimization},
  author={Bolte, J{\'e}r{\^o}me and Le, Tam and Pauwels, Edouard},
  journal={SIAM Journal on Optimization},
  volume={33},
  number={4},
  pages={2542--2569},
  year={2023},
  publisher={SIAM}
}

@article{bolte2007clarke,
  title={Clarke subgradients of stratifiable functions},
  author={Bolte, J{\'e}r{\^o}me and Daniilidis, Aris and Lewis, Adrian and Shiota, Masahiro},
  journal={SIAM Journal on Optimization},
  volume={18},
  number={2},
  pages={556--572},
  year={2007},
  publisher={SIAM}
}

@article{bolte2021conservative,
  title={Conservative set valued fields, automatic differentiation, stochastic gradient methods and deep learning},
  author={Bolte, J{\'e}r{\^o}me and Pauwels, Edouard},
  journal={Mathematical Programming},
  volume={188},
  number={1},
  pages={19--51},
  year={2021},
  publisher={Springer}
}

@book{clarke1990optimization,
  title={Optimization and nonsmooth analysis},
  author={Clarke, Frank H},
  year={1990},
  publisher={SIAM}
}

@article{hestenes1969multiplier,
  title={Multiplier and gradient methods},
  author={Hestenes, Magnus R},
  journal={Journal of optimization theory and applications},
  volume={4},
  number={5},
  pages={303--320},
  year={1969},
  publisher={Springer}
}

@article{raissi2017physics,
  title={Physics informed deep learning (part i): Data-driven solutions of nonlinear partial differential equations},
  author={Raissi, Maziar and Perdikaris, Paris and Karniadakis, George Em},
  journal={arXiv preprint arXiv:1711.10561},
  year={2017}
}

@article{wang2022and,
  title={When and why PINNs fail to train: A neural tangent kernel perspective},
  author={Wang, Sifan and Yu, Xinling and Perdikaris, Paris},
  journal={Journal of Computational Physics},
  volume={449},
  pages={110768},
  year={2022},
  publisher={Elsevier}
}

@misc{dembo2019lecture,
  title={Lecture notes for probability theory: STAT310/MATH230},
  author={Dembo, A},
  year={2019},
  url={https://web.stanford.edu/class/stats310a/lnotes.pdf}
}

@article{kakade2018provably,
  title={Provably correct automatic sub-differentiation for qualified programs},
  author={Kakade, Sham M and Lee, Jason D},
  journal={Advances in neural information processing systems},
  volume={31},
  year={2018}
}

@article{arjevani2023lower,
  title={Lower bounds for non-convex stochastic optimization: Y. Arjevani et al.},
  author={Arjevani, Yossi and Carmon, Yair and Duchi, John C and Foster, Dylan J and Srebro, Nathan and Woodworth, Blake},
  journal={Mathematical Programming},
  volume={199},
  number={1},
  pages={165--214},
  year={2023},
  publisher={Springer}
}

@book{clason2026introduction,
  title={Introduction to nonsmooth analysis and optimization},
  author={Clason, Christian and Valkonen, Tuomo},
  year={2026},
  publisher={SIAM}
}

@article{liu2026spider,
  title={A SPIDER-Type Stochastic Subgradient Method for Expectation-Constrained Nonconvex Nonsmooth Optimization},
  author={Liu, Wei and Xu, Yangyang},
  journal={SIAM Journal on Optimization},
  volume={36},
  number={2},
  pages={1125--1153},
  year={2026},
  publisher={SIAM}
}

@InProceedings{pmlr-v70-achiam17a,
  title = 	 {Constrained Policy Optimization},
  author =       {Joshua Achiam and David Held and Aviv Tamar and Pieter Abbeel},
  booktitle = 	 {Proceedings of the 34th International Conference on Machine Learning},
  pages = 	 {22--31},
  year = 	 {2017},
  editor = 	 {Precup, Doina and Teh, Yee Whye},
  volume = 	 {70},
  series = 	 {Proceedings of Machine Learning Research},
  month = 	 {06--11 Aug},
  publisher =    {PMLR},
  url = 	 {https://proceedings.mlr.press/v70/achiam17a.html}
}

@article{strukov1977mathematical,
  title={Mathematical expectation of continuous functions of random variables. Smoothness and variance},
  author={Strukov, LI and Timan, Aleksandr Filippovich},
  journal={Siberian Mathematical Journal},
  volume={18},
  number={3},
  pages={469--474},
  year={1977},
  publisher={Springer}
}

@article{yang2025single,
  title={Single-loop algorithms for stochastic non-convex optimization with weakly-convex constraints},
  author={Yang, Ming and Li, Gang and Hu, Quanqi and Lin, Qihang and Yang, Tianbao},
  journal={arXiv preprint arXiv:2504.15243},
  year={2025}
}

@article{abramovich2016some,
  title={Some new estimates of the ‘Jensen gap’},
  author={Abramovich, Shoshana and Persson, Lars-Erik},
  journal={Journal of Inequalities and Applications},
  volume={2016},
  number={1},
  pages={39},
  year={2016},
  publisher={Springer}
}

@article{gao2017bounds,
  title={Bounds on the Jensen gap, and implications for mean-concentrated distributions},
  author={Gao, Xiang and Sitharam, Meera and Roitberg, Adrian E},
  journal={arXiv preprint arXiv:1712.05267},
  year={2017}
}

@article{loshchilov2016sgdr,
  title={Sgdr: Stochastic gradient descent with warm restarts},
  author={Loshchilov, Ilya and Hutter, Frank},
  journal={arXiv preprint arXiv:1608.03983},
  year={2016}
}

@article{song2024admm,
  title={The ADMM-PINNs algorithmic framework for nonsmooth PDE-constrained optimization: a deep learning approach},
  author={Song, Yongcun and Yuan, Xiaoming and Yue, Hangrui},
  journal={SIAM Journal on Scientific Computing},
  volume={46},
  number={6},
  pages={C659--C687},
  year={2024},
  publisher={SIAM}
}

@article{son2023enhanced,
  title={Enhanced physics-informed neural networks with augmented Lagrangian relaxation method (AL-PINNs)},
  author={Son, Hwijae and Cho, Sung Woong and Hwang, Hyung Ju},
  journal={Neurocomputing},
  volume={548},
  pages={126424},
  year={2023},
  publisher={Elsevier}
}

@article{wang2021understanding,
  title={Understanding and mitigating gradient flow pathologies in physics-informed neural networks},
  author={Wang, Sifan and Teng, Yujun and Perdikaris, Paris},
  journal={SIAM Journal on Scientific Computing},
  volume={43},
  number={5},
  pages={A3055--A3081},
  year={2021},
  publisher={SIAM}
}

@inproceedings{zhang2020complexity,
  title={Complexity of finding stationary points of nonconvex nonsmooth functions},
  author={Zhang, Jingzhao and Lin, Hongzhou and Jegelka, Stefanie and Sra, Suvrit and Jadbabaie, Ali},
  booktitle={International Conference on Machine Learning},
  pages={11173--11182},
  year={2020},
  organization={PMLR}
}

@article{powell1969method,
  title={A method for nonlinear constraints in minimization problems},
  author={Powell, Michael JD},
  journal={Optimization},
  pages={283--298},
  year={1969},
  publisher={Academic Press}
}

@article{rockafellar1973dual,
  title={A dual approach to solving nonlinear programming problems by unconstrained optimization},
  author={Rockafellar, R Tyrrell},
  journal={Mathematical programming},
  volume={5},
  number={1},
  pages={354--373},
  year={1973},
  publisher={Springer}
}

@article{rockafellar1973multiplier,
  title={The multiplier method of Hestenes and Powell applied to convex programming},
  author={Rockafellar, R Tyrrell},
  journal={Journal of Optimization Theory and applications},
  volume={12},
  number={6},
  pages={555--562},
  year={1973},
  publisher={Springer}
}

@article{rockafellar1974augmented,
  title={Augmented Lagrange multiplier functions and duality in nonconvex programming},
  author={Rockafellar, R Tyrrell},
  journal={SIAM Journal on Control},
  volume={12},
  number={2},
  pages={268--285},
  year={1974},
  publisher={SIAM}
}

@article{rockafellar1976augmented,
  title={Augmented Lagrangians and applications of the proximal point algorithm in convex programming},
  author={Rockafellar, R Tyrrell},
  journal={Mathematics of operations research},
  volume={1},
  number={2},
  pages={97--116},
  year={1976},
  publisher={INFORMS}
}

@article{zhang2022iteration,
  title={On the iteration complexity of smoothed proximal ALM for nonconvex optimization problem with convex constraints},
  author={Zhang, Jiawei and Pu, Wenqiang and Luo, Zhi-Quan},
  journal={arXiv preprint arXiv:2207.06304},
  year={2022}
}

@article{zhang2020proximal,
  title={A proximal alternating direction method of multiplier for linearly constrained nonconvex minimization},
  author={Zhang, Jiawei and Luo, Zhi-Quan},
  journal={SIAM Journal on Optimization},
  volume={30},
  number={3},
  pages={2272--2302},
  year={2020},
  publisher={SIAM}
}

@article{lin2022complexity,
  title={Complexity of an inexact proximal-point penalty method for constrained smooth non-convex optimization},
  author={Lin, Qihang and Ma, Runchao and Xu, Yangyang},
  journal={Computational optimization and applications},
  volume={82},
  number={1},
  pages={175--224},
  year={2022},
  publisher={Springer}
}

@article{pu2024smoothed,
  title={Smoothed proximal Lagrangian method for nonlinear constrained programs},
  author={Pu, Wenqiang and Sun, Kaizhao and Zhang, Jiawei},
  journal={arXiv preprint arXiv:2408.15047},
  year={2024}
}

@article{kong2023iteration,
  title={Iteration complexity of a proximal augmented Lagrangian method for solving nonconvex composite optimization problems with nonlinear convex constraints},
  author={Kong, Weiwei and Melo, Jefferson G and Monteiro, Renato DC},
  journal={Mathematics of Operations Research},
  volume={48},
  number={2},
  pages={1066--1094},
  year={2023},
  publisher={INFORMS}
}

@article{kong2019complexity,
  title={Complexity of a quadratic penalty accelerated inexact proximal point method for solving linearly constrained nonconvex composite programs},
  author={Kong, Weiwei and Melo, Jefferson G and Monteiro, Renato DC},
  journal={SIAM Journal on Optimization},
  volume={29},
  number={4},
  pages={2566--2593},
  year={2019},
  publisher={SIAM}
}

@article{ben1997penalty,
  title={Penalty/barrier multiplier methods for convex programming problems},
  author={Ben-Tal, Aharon and Zibulevsky, Michael},
  journal={SIAM Journal on Optimization},
  volume={7},
  number={2},
  pages={347--366},
  year={1997},
  publisher={SIAM}
}

@article{kliachkin2025humancompatible,
  title={humancompatible.train: Implementing Optimization Algorithms for Stochastically-Constrained Stochastic Optimization Problems},
  author={Kliachkin, Andrii and Lep{\v{s}}ov{\'a}, Jana and Bareilles, Gilles and Mare{\v{c}}ek, Jakub}, 
  journal={NeurIPS Workshop on Constrained Optimization; arXiv preprint arXiv:2509.21254},
  year={2025}
}

@article{singh2024mini,
  title={Mini-batch stochastic subgradient for functional constrained optimization},
  author={Singh, Nitesh Kumar and Necoara, Ion and Kungurtsev, Vyacheslav},
  journal={Optimization},
  volume={73},
  number={7},
  pages={2159--2185},
  year={2024},
  publisher={Taylor \& Francis}
}

@article{necoara2021minibatch,
  title={Minibatch stochastic subgradient-based projection algorithms for feasibility problems with convex inequalities},
  author={Necoara, Ion and Nedi{\'c}, Angelia},
  journal={Computational Optimization and Applications},
  volume={80},
  number={1},
  pages={121--152},
  year={2021},
  publisher={Springer}
}

@article{singh2024stochastic,
  title={Stochastic halfspace approximation method for convex optimization with nonsmooth functional constraints},
  author={Singh, Nitesh Kumar and Necoara, Ion},
  journal={IEEE Transactions on Automatic Control},
  year={2024},
  publisher={IEEE}
}

@article{necoara2022stochastic,
  title={Stochastic subgradient for composite convex optimization with functional constraints},
  author={Necoara, Ion and Singh, Nitesh Kumar},
  journal={Journal of Machine Learning Research},
  volume={23},
  number={265},
  pages={1--35},
  year={2022}
}

@article{Ioffe2009,
author = {Ioffe, A. D.},
title = {An Invitation to Tame Optimization},
journal = {SIAM Journal on Optimization},
volume = {19},
number = {4},
pages = {1894-1917},
year = {2009},
doi = {10.1137/080722059},

URL = { 
    
        https://doi.org/10.1137/080722059
    
    

},
eprint = { 
    
        https://doi.org/10.1137/080722059
    
    

}
}

@article{bareilles2025deep,
  title={Deep Learning as the Disciplined Construction of Tame Objects},
  author={Bareilles, Gilles and Gehret, Allen and Aspman, Johannes and Lep{\v{s}}ov{\'a}, Jana and Mare{\v{c}}ek, Jakub},
  journal={arXiv preprint arXiv:2509.18025},
  year={2025}
}

@article{curtis2025stochastic,
  title={A stochastic-gradient-based interior-point algorithm for solving smooth bound-constrained optimization problems},
  author={Curtis, Frank E and Kungurtsev, Vyacheslav and Robinson, Daniel P and Wang, Qi},
  journal={SIAM Journal on Optimization},
  volume={35},
  number={2},
  pages={1030--1059},
  year={2025},
  publisher={SIAM}
}

@article{curtis2026single,
  title={A single-loop stochastic feasible interior-point algorithm for nonlinear inequality-constrained optimization: F. Curtis et al.},
  author={Curtis, Frank E and Jiang, Xin and Wang, Qi},
  journal={Mathematical Programming},
  pages={1--38},
  year={2026},
  publisher={Springer}
}

@article{fang2026trust,
  title={A Trust-Region Interior-Point Stochastic Sequential Quadratic Programming Method},
  author={Fang, Yuchen and Kim, Jihun and Na, Sen and Demmel, James and Lavaei, Javad},
  journal={arXiv preprint arXiv:2603.10230},
  year={2026}
}

@misc{krizhevsky2009learning,
  title={Learning multiple layers of features from tiny images.(2009)},
  author={Krizhevsky, Alex and Hinton, Geoffrey and others},
  year={2009}
}

@article{ding2021retiring,
  title={Retiring Adult: New Datasets for Fair Machine Learning},
  author={Ding, Frances and Hardt, Moritz and Miller, John and Schmidt, Ludwig},
  journal={Advances in Neural Information Processing Systems},
  volume={34},
  year={2021}
}

@inbook{inbook_dutch,
author = {Van der Laan, Paul},
year = {2001},
month = {12},
pages = {39-52},
title = {The 2001 Census in the Netherlands: Integration of Registers and Surveys}
}

@article{de2025penalty,
  title={A penalty barrier framework for nonconvex constrained optimization},
  author={De Marchi, Alberto and Themelis, Andreas},
  journal={Journal of Nonsmooth Analysis and Optimization},
  volume={5},
  number={Original research articles},
  year={2025},
  publisher={Episciences. org}
}

@article{jensen1906fonctions,
  title={Sur les fonctions convexes et les in{\'e}galit{\'e}s entre les valeurs moyennes},
  author={Jensen, Johan Ludwig William Valdemar},
  journal={Acta mathematica},
  volume={30},
  number={1},
  pages={175--193},
  year={1906},
  publisher={Springer}
}

@misc{ramirezPositionAdoptConstraints2025,
  title = {Position: {{Adopt Constraints Over Penalties}} in {{Deep Learning}}},
  shorttitle = {Position},
  author = {Ramirez, Juan and Hashemizadeh, Meraj and {Lacoste-Julien}, Simon},
  year = {2025},
  month = jul,
  number = {arXiv:2505.20628},
  eprint = {2505.20628},
  primaryclass = {cs},
  publisher = {arXiv},
  doi = {10.48550/arXiv.2505.20628},
  urldate = {2025-08-22},
  archiveprefix = {arXiv},
}

@article{boob2023stochastic,
  title={Stochastic first-order methods for convex and nonconvex functional constrained optimization},
  author={Boob, Digvijay and Deng, Qi and Lan, Guanghui},
  journal={Mathematical Programming},
  volume={197},
  number={1},
  pages={215--279},
  year={2023},
  publisher={Springer}
}

@article{kovcvara2003pennon,
  title={PENNON: A code for convex nonlinear and semidefinite programming},
  author={Ko{\v{c}}vara, Michal and Stingl, Michael},
  journal={Optimization methods and software},
  volume={18},
  number={3},
  pages={317--333},
  year={2003},
  publisher={Taylor \& Francis}
}

@incollection{kocvara2012pennon,
  title={PENNON: Software for linear and nonlinear matrix inequalities},
  author={Ko{\v{c}}vara, Michal and Stingl, Michael},
  booktitle={Handbook on semidefinite, conic and polynomial optimization},
  pages={755--791},
  year={2012},
  publisher={Springer}
}

@inproceedings{ma2020quadratically,
  title={Quadratically regularized subgradient methods for weakly convex optimization with weakly convex constraints},
  author={Ma, Runchao and Lin, Qihang and Yang, Tianbao},
  booktitle={International Conference on Machine Learning},
  pages={6554--6564},
  year={2020},
  organization={PMLR}
}

@incollection{kovcvara2003pennonb,
  title={PENNON: a generalized augmented Lagrangian method for semidefinite programming},
  author={Ko{\v{c}}vara, Michal and Stingl, Michael},
  booktitle={High performance algorithms and software for nonlinear optimization},
  pages={303--321},
  year={2003},
  publisher={Springer}
}

@article{kingma2014adam,
  title={Adam: A method for stochastic optimization},
  author={Kingma, Diederik P and Ba, Jimmy},
  journal={arXiv preprint arXiv:1412.6980},
  year={2014}
}

@inproceedings{he2016deep,
  title={Deep residual learning for image recognition},
  author={He, Kaiming and Zhang, Xiangyu and Ren, Shaoqing and Sun, Jian},
  booktitle={Proceedings of the IEEE conference on computer vision and pattern recognition},
  pages={770--778},
  year={2016}
}

@misc{gallego-posadaCooperLibraryConstrained2025,
  title = {Cooper: {{A Library}} for {{Constrained Optimization}} in {{Deep Learning}}},
  shorttitle = {Cooper},
  author = {{Gallego-Posada}, Jose and Ramirez, Juan and Hashemizadeh, Meraj and {Lacoste-Julien}, Simon},
  year = {2025},
  month = apr,
  number = {arXiv:2504.01212},
  eprint = {2504.01212},
  primaryclass = {cs},
  publisher = {arXiv},
  doi = {10.48550/arXiv.2504.01212},
  urldate = {2025-08-25},
  archiveprefix = {arXiv},
}

@article{chamon2022constrained,
  title={Constrained learning with non-convex losses},
  author={Chamon, Luiz FO and Paternain, Santiago and Calvo-Fullana, Miguel and Ribeiro, Alejandro},
  journal={IEEE Transactions on Information Theory},
  volume={69},
  number={3},
  pages={1739--1760},
  year={2022},
  publisher={IEEE}
}

@article{liKernelDependenceRegularizers2022,
  title = {Kernel Dependence Regularizers and {{Gaussian}} Processes with Applications to Algorithmic Fairness},
  author = {Li, Zhu and {P{\'e}rez-Suay}, Adri{\'a}n and {Camps-Valls}, Gustau and Sejdinovic, Dino},
  year = {2022},
  month = dec,
  journal = {Pattern Recognition},
  volume = {132},
  pages = {108922},
  issn = {0031-3203},
  doi = {10.1016/j.patcog.2022.108922},
  urldate = {2025-05-12},
}

@article{lu2021physics,
  title={Physics-informed neural networks with hard constraints for inverse design},
  author={Lu, Lu and Pestourie, Raphael and Yao, Wenjie and Wang, Zhicheng and Verdugo, Francesc and Johnson, Steven G},
  journal={SIAM Journal on Scientific Computing},
  volume={43},
  number={6},
  pages={B1105--B1132},
  year={2021},
  publisher={SIAM}
}

@inproceedings{buyl2024fairret,
title={fairret: a Framework for Differentiable Fairness Regularization Terms},
author={Buyl, Maarten and Defrance, Marybeth and De Bie, Tijl},
booktitle={International Conference on Learning Representations},
year={2024}}

@book{rockafellar2009variational,
  title={Variational analysis},
  author={Rockafellar, R Tyrrell and Wets, Roger J-B},
  volume={317},
  year={2009},
  publisher={Springer Science \& Business Media}
}

@book{beck2017first,
  title     = {First-Order Methods in Optimization},
  author    = {Beck, Amir},
  year      = {2017},
  publisher = {SIAM},
  address   = {Philadelphia, PA},
  doi       = {10.1137/1.9781611974997}
}

@article{McClenny_2023,
   title={Self-adaptive physics-informed neural networks},
   volume={474},
   ISSN={0021-9991},
   url={http://dx.doi.org/10.1016/j.jcp.2022.111722},
   DOI={10.1016/j.jcp.2022.111722},
   journal={Journal of Computational Physics},
   publisher={Elsevier BV},
   author={McClenny, Levi D. and Braga-Neto, Ulisses M.},
   year={2023},
   month=Feb, pages={111722} }

@misc{FacKun2023,
      title={Stochastic Approximation for Expectation Objective and Expectation Inequality-Constrained Nonconvex Optimization}, 
      author={Francisco Facchinei and Vyacheslav Kungurtsev},
      year={2023},
      eprint={2307.02943},
      archivePrefix={arXiv},
      primaryClass={math.OC},
      url={https://arxiv.org/abs/2307.02943}, 
}

@Article{NaAnKo2022,
	author={Na, Sen and Anitescu, Mihai and Kolar, Mladen},
	title={An adaptive stochastic sequential quadratic programming with differentiable exact augmented Lagrangians},
	journal={Mathematical Programming},
	year={2023},
	month={May},
	day={01},
	volume={199},
	number={1},
	pages={721-791},
	doi={10.1007/s10107-022-01846-z},
	url={https://doi.org/10.1007/s10107-022-01846-z}
}

@article{FaNaMaKo2024,
author = {Fang, Yuchen and Na, Sen and Mahoney, Michael W. and Kolar, Mladen},
title = {Fully Stochastic Trust-Region Sequential Quadratic Programming for Equality-Constrained Optimization Problems},
journal = {SIAM Journal on Optimization},
volume = {34},
number = {2},
pages = {2007-2037},
year = {2024},
doi = {10.1137/22M1537862},
URL = { https://doi.org/10.1137/22M1537862},
eprint = { https://doi.org/10.1137/22M1537862},
}

@Article{Curtis2024,
author={Curtis, Frank E.  and O'Neill, Michael J.  and Robinson, Daniel P.},
title={Worst-case complexity of an SQP method for nonlinear equality constrained stochastic optimization},
journal={Mathematical Programming},
year={2024},
month={May},
day={01},
volume={205},
number={1},
pages={431--483},
issn={1436-4646},
doi={10.1007/s10107-023-01981-1},
url={https://doi.org/10.1007/s10107-023-01981-1}
}

@article{Ozto2023,
author = {Oztoprak, Figen and Byrd, Richard and Nocedal, Jorge},
title = {Constrained Optimization in the Presence of Noise},
journal = {SIAM Journal on Optimization},
volume = {33},
number = {3},
pages = {2118-2136},
year = {2023},
doi = {10.1137/21M1450999},
URL = { https://doi.org/10.1137/21M1450999 },
eprint = { https://doi.org/10.1137/21M1450999 },
}

@misc{BerCur2023,
      title={A Stochastic Sequential Quadratic Optimization Algorithm for Nonlinear Equality Constrained Optimization with Rank-Deficient Jacobians}, 
      author={Albert S. Berahas and Frank E. Curtis and Michael J. O'Neill and Daniel P. Robinson},
      year={2023},
      eprint={2106.13015},
      archivePrefix={arXiv},
      primaryClass={math.OC},
      url={https://arxiv.org/abs/2106.13015}, 
}

@misc{NaAnKo2023,
      title={Inequality Constrained Stochastic Nonlinear Optimization via Active-Set Sequential Quadratic Programming}, 
      author={Sen Na and Mihai Anitescu and Mladen Kolar},
      year={2023},
      eprint={2109.11502},
      archivePrefix={arXiv},
      primaryClass={math.OC},
      url={https://arxiv.org/abs/2109.11502}, 
}

@article{Bolla2023,
author = {Raghu Bollapragada and Cem Karamanli and Brendan Keith and Boyan Lazarov and Socratis Petrides and Jingyi Wang},
title = {An adaptive sampling augmented Lagrangian method for stochastic optimization with deterministic constraints},
journal = {Computers and Mathematics with Applications},
volume = {149},
pages = {239-258},
year = {2023},
issn = {0898-1221},
doi = {https://doi.org/10.1016/j.camwa.2023.09.014},
url = {https://www.sciencedirect.com/science/article/pii/S0898122123003991},
}

@article{CurRob2024,
author = {Curtis, Frank E. and Robinson, Daniel P. and Zhou, Baoyu},
title = {Sequential Quadratic Optimization for Stochastic Optimization with Deterministic Nonlinear Inequality and Equality Constraints},
journal = {SIAM Journal on Optimization},
volume = {34},
number = {4},
pages = {3592-3622},
year = {2024},
doi = {10.1137/23M1556149},

URL = { https://doi.org/10.1137/23M1556149 },
eprint = { https://doi.org/10.1137/23M1556149 },
}

@inproceedings{HuaLin2023,
 author = {Huang, Yankun and Lin, Qihang},
 booktitle = {Advances in Neural Information Processing Systems},
 editor = {A. Oh and T. Naumann and A. Globerson and K. Saenko and M. Hardt and S. Levine},
 pages = {61327--61340},
 publisher = {Curran Associates, Inc.},
 title = {Oracle Complexity of Single-Loop Switching Subgradient Methods for Non-Smooth Weakly Convex Functional Constrained Optimization},
 url = {https://proceedings.neurips.cc/paper_files/paper/2023/file/c132c02176577c4319a878f6417a331a-Paper-Conference.pdf},
 volume = {36},
 year = {2023}
}

@InProceedings{pmlr-v267-huang25au,
  title = 	 {Stochastic Smoothed Primal-Dual Algorithms for Nonconvex Optimization with Linear Inequality Constraints},
  author =       {Huang, Ruichuan and Zhang, Jiawei and Alacaoglu, Ahmet},
  booktitle = 	 {Proceedings of the 42nd International Conference on Machine Learning},
  pages = 	 {26106--26142},
  year = 	 {2025},
  editor = 	 {Singh, Aarti and Fazel, Maryam and Hsu, Daniel and Lacoste-Julien, Simon and Berkenkamp, Felix and Maharaj, Tegan and Wagstaff, Kiri and Zhu, Jerry},
  volume = 	 {267},
  series = 	 {Proceedings of Machine Learning Research},
  month = 	 {13--19 Jul},
  publisher =    {PMLR},
  url = 	 {https://proceedings.mlr.press/v267/huang25au.html}
}

@article{ShiWaWa2022,
author = {Shi, Qiankun and Wang, Xiao and Wang, Hao},
title = {A momentum-based linearized augmented Lagrangian method for nonconvex constrained stochastic optimization},
journal = {Optimization Online},
year = {2022},
URL = {https://optimization-online.org/?p=19870},
}

@article{Cotter2019,
    title = {Optimization with Non-Differentiable Constraints with Applications to Fairness, Recall, Churn, and Other Goals},
    author = {Andrew Cotter and Heinrich Jiang and Serena Wang and Taman Narayan and Seungil You and Karthik Sridharan and Maya R. Gupta},
    volume={20},   
    number={172},   
    pages={1--59},
    year = {2019},
    journal = {Journal of Machine Learning Research}
}

@misc{DacDruKakLee2018,
      title={Stochastic subgradient method converges on tame functions}, 
      author={Damek Davis and Dmitriy Drusvyatskiy and Sham Kakade and Jason D. Lee},
      year={2018},
      eprint={1804.07795},
      archivePrefix={arXiv},
      primaryClass={math.OC},
      url={https://arxiv.org/abs/1804.07795}, 
}

@article{xu2004viscosity,
  title={Viscosity approximation methods for nonexpansive mappings},
  author={Xu, Hong-Kun},
  journal={Journal of Mathematical Analysis and Applications},
  volume={298},
  number={1},
  pages={279--291},
  year={2004},
  publisher={Elsevier}
}

@book{van1998tame,
  author    = {van den Dries, Lou},
  title     = {Tame Topology and O-minimal Structures},
  series    = {London Mathematical Society Lecture Note Series},
  volume    = {248},
  publisher = {Cambridge University Press},
  address   = {Cambridge},
  year      = {1998},
  isbn      = {9780521598385}
}

@article{fusek2002examples,
  title={Examples and counterexamples in Lipschitz analysis},
  author={Fusek, Peter and Klatte, Diethard and Kummer, Bernd},
  journal={Control and Cybernetics},
  volume={31},
  number={3},
  pages={471--492},
  year={2002}
}

@article{daniilidis2020pathological,
  title={Pathological subgradient dynamics},
  author={Daniilidis, Aris and Drusvyatskiy, Dmitriy},
  journal={SIAM Journal on Optimization},
  volume={30},
  number={2},
  pages={1327--1338},
  year={2020},
  publisher={SIAM}
}

@article{rios2022examples,
  title={Examples of pathological dynamics of the subgradient method for Lipschitz path-differentiable functions},
  author={Rios-Zertuche, Rodolfo},
  journal={Mathematics of Operations Research},
  volume={47},
  number={4},
  pages={3184--3206},
  year={2022},
  publisher={INFORMS}
}

@incollection{van1984remarks,
  title={Remarks on Tarski's problem concerning (R,+,*, exp)},
  author={van den Dries, Lou},
  booktitle={Studies in Logic and the Foundations of Mathematics},
  volume={112},
  pages={97--121},
  year={1984},
  publisher={Elsevier}
}

@article{pillay1984definable,
  title={Definable sets in ordered structures},
  author={Pillay, Anand and Steinhorn, Charles},
  year={1984}
}

@book {Tarski48,
    AUTHOR = {Tarski, Alfred},
     TITLE = {A {D}ecision {M}ethod for {E}lementary {A}lgebra and
              {G}eometry},
 PUBLISHER = {The Rand Corporation, Santa Monica, CA},
      YEAR = {1948},
     PAGES = {iii+60},
   MRCLASS = {02.0X},
  MRNUMBER = {28796},
MRREVIEWER = {A.\ Heyting},
}

@article{wilkie1996model,
    AUTHOR = {Wilkie, A. J.},
     TITLE = {Model completeness results for expansions of the ordered field
              of real numbers by restricted {P}faffian functions and the
              exponential function},
   JOURNAL = {J. Amer. Math. Soc.},
  FJOURNAL = {Journal of the American Mathematical Society},
    VOLUME = {9},
      YEAR = {1996},
    NUMBER = {4},
     PAGES = {1051--1094},
      ISSN = {0894-0347,1088-6834},
   MRCLASS = {03C62 (03C60 03C65 14P15)},
  MRNUMBER = {1398816},
MRREVIEWER = {Luc\ B\'elair},
       DOI = {10.1090/S0894-0347-96-00216-0},
       }

@article {Speissegger1999Pfaff,
    AUTHOR = {Speissegger, Patrick},
     TITLE = {The {P}faffian closure of an o-minimal structure},
   JOURNAL = {J. Reine Angew. Math.},
  FJOURNAL = {Journal f\"ur die Reine und Angewandte Mathematik. [Crelle's
              Journal]},
    VOLUME = {508},
      YEAR = {1999},
     PAGES = {189--211},
      ISSN = {0075-4102,1435-5345},
   MRCLASS = {14P10 (03C64)},
  MRNUMBER = {1676876},
MRREVIEWER = {A.\ J.\ Wilkie},
       DOI = {10.1515/crll.1999.026},
       URL = {https://doi.org/10.1515/crll.1999.026},
}

@article{borwein2000lipschitz,
  title={Lipschitz functions with maximal Clarke subdifferentials are generic},
  author={Borwein, Jonathan and Wang, Xianfu},
  journal={Proceedings of the American Mathematical Society},
  volume={128},
  number={11},
  pages={3221--3229},
  year={2000}
}

@article{daniilidis2024all,
  title={All convex bodies are in the subdifferential of some everywhere differentiable locally Lipschitz function},
  author={Daniilidis, Aris and Deville, Robert and Tapia-Garc{\'\i}a, Sebasti{\'a}n},
  journal={Proceedings of the London Mathematical Society},
  volume={129},
  number={5},
  pages={e70007},
  year={2024},
  publisher={Wiley Online Library}
}

@article{kranz2026sad,
  title={Sad neural networks: Divergent gradient flows and asymptotic optimality via o-minimal structures},
  author={Kranz, Julian and Gallon, Davide and Dereich, Steffen and Jentzen, Arnulf},
  journal={Advances in Neural Information Processing Systems},
  volume={38},
  pages={3930--3966},
  year={2026}
}

@article{ji2020directional,
  title={Directional convergence and alignment in deep learning},
  author={Ji, Ziwei and Telgarsky, Matus},
  journal={Advances in Neural Information Processing Systems},
  volume={33},
  pages={17176--17186},
  year={2020}
}

@article{kratsios2026every,
  title={Every Feedforward Neural Network Definable in an o-Minimal Structure Has Finite Sample Complexity},
  author={Kratsios, Anastasis and Cousins, Gregory and Borde, Haitz S{\'a}ez de Oc{\'a}riz and Kim, Bum Jun and Brugiapaglia, Simone},
  journal={arXiv preprint arXiv:2605.07097},
  year={2026}
}

@article{kinn2026fast,
  title={Fast approximation and learning of binary classification tasks in o-minimal structures using ReLU neural networks},
  author={Kinn, Clemens and Petersen, Philipp},
  journal={arXiv preprint arXiv:2607.01266},
  year={2026}
}

@article{aschenbrenner2015asymptotic,
  title={Asymptotic differential algebra and model theory of transseries},
  author={Aschenbrenner, Matthias and Dries, Lou van den and van der Hoeven, Joris},
  journal={arXiv preprint arXiv:1509.02588},
  year={2015}
}

@book{enderton2001mathematical,
  title={A mathematical introduction to logic},
  author={Enderton, Herbert B},
  year={2001},
  publisher={Elsevier}
}

@book{hils2019first,
  title={A first journey through logic},
  author={Hils, Martin and Loeser, Fran{\c{c}}ois},
  volume={89},
  year={2019},
  publisher={American Mathematical Soc.}
}

@book{marker2002model,
  title={Model theory: an introduction},
  author={Marker, David},
  year={2002},
  publisher={Springer}
}

@article{van1994elementary,
    AUTHOR = {Dries, Lou van den and Macintyre, Angus and Marker, David},
     TITLE = {The elementary theory of restricted analytic fields with
              exponentiation},
   JOURNAL = {Ann. of Math. (2)},
  FJOURNAL = {Annals of Mathematics. Second Series},
    VOLUME = {140},
      YEAR = {1994},
    NUMBER = {1},
     PAGES = {183--205},
      ISSN = {0003-486X,1939-8980},
   MRCLASS = {12L12 (03C10 03C62)},
  MRNUMBER = {1289495},
MRREVIEWER = {Thanases\ Pheidas},
       DOI = {10.2307/2118545},
       }

@misc{kunc2024decadesactivationscomprehensivesurvey,
      title={Three Decades of Activations: A Comprehensive Survey of 400 Activation Functions for Neural Networks},
      author={Vladimír Kunc and Jiří Kléma},
      year={2024},
      eprint={2402.09092},
      archivePrefix={arXiv},
      primaryClass={cs.LG},
      url={https://arxiv.org/abs/2402.09092},
}

@article{le1998verdier,
    AUTHOR = {Ta L\^e{} Loi},
     TITLE = {Verdier and strict {T}hom stratifications in o-minimal
              structures},
   JOURNAL = {Illinois J. Math.},
  FJOURNAL = {Illinois Journal of Mathematics},
    VOLUME = {42},
      YEAR = {1998},
    NUMBER = {2},
     PAGES = {347--356},
      ISSN = {0019-2082,1945-6581},
   MRCLASS = {32S60 (14P10)},
  MRNUMBER = {1612771},
MRREVIEWER = {Karim\ Bekka},
       }
\bibliographystyle{iclr2027_conference}


\newpage
\appendix

\begin{center}
    {\Large \bf Supplementary material}
\end{center}

\startcontents[appendix]
\section*{Outline of the appendix}
\printcontents[appendix]{}{1}{\setcounter{tocdepth}{2}}
\bigskip

\newpage
\section{List of Notation}
\makenomenclature
\renewcommand{\nomname}{}
\renewcommand\nomgroup[1]{%
  \item[\bfseries
    \ifstrequal{#1}{A}{Probability, data and sampling}{%
    \ifstrequal{#1}{B}{Problem, algorithm and iterates}{%
    \ifstrequal{#1}{C}{Convergence analysis}{%
    \ifstrequal{#1}{D}{General notation}{}}}}%
  ]\vspace{3pt}}
\nomenclature[A01]{$(\Omega,\mathcal{F},P)$}{Probability space carrying the mini-batch draws}
\nomenclature[A02]{$\Omega$}{Sample space; realised canonically as $(\{1,\dots,M\}^B)^{\mathbb{N}_0}$}
\nomenclature[A03]{$P$}{Probability measure on $\Omega$; the product measure $\bigotimes_{k\ge0}\mathrm{Unif}(\{1,\dots,M\}^B)$, i.e.\ its factors are uniform}
\nomenclature[A04]{$\mathcal{F}^k$}{$\sigma$-algebra generated by the batches drawn before iteration $k$, $\sigma(\mathbf{j}^0,\dots,\mathbf{j}^{k-1})$}
\nomenclature[A05]{$\Omega_0$}{Event of full measure on which the convergence analysis is carried out}
\nomenclature[A06]{$\mathbb{E}$}{Expectation operator; $\mathbb{E}[\,\cdot\mid\mathcal{F}^k]$ the conditional expectation given $\mathcal{F}^k$}
\nomenclature[A07]{$\xi$}{Random variable; a single data sample}
\nomenclature[A08]{$\Xi$}{Distribution of $\xi$, i.e.\ a probability measure on the data space;}
\nomenclature[A09]{$\operatorname{supp}\Xi$}{Support of $\Xi$; the dataset $\{\xi^{(1)},\dots,\xi^{(M)}\}$ in case of no repetitions}
\nomenclature[A10]{$\xi^{(l)}$}{$l$-th element of the dataset, $l\in\{1,\dots,M\}$}
\nomenclature[A11]{$M$}{Number of samples in the dataset}
\nomenclature[A12]{$B$}{Mini-batch size}
\nomenclature[A13]{$\xi_b$}{$b$-th sample of a mini-batch, $b\in\{1,\dots,B\}$; the batch is written $\xi_{1:B}$}
\nomenclature[A14]{$\mathbf{j}$}{Multi-index encoding a mini-batch, $\mathbf{j}=(j_1,\dots,j_B)\in\{1,\dots,M\}^B$; the $b$-th sample of the batch is $\xi^{(j_b)}$}
\nomenclature[A15]{$\mathbf{j}^k$}{Mini-batch index drawn at iteration $k$; $P(\mathbf{j}^k=\mathbf{j})=M^{-B}$}

\nomenclature[B01]{$n$}{Dimension of the primal variable $x$}
\nomenclature[B02]{$m$}{Number of constraints}
\nomenclature[B03]{$f:\mathbb{R}^n\to\mathbb{R}$}{Objective function}
\nomenclature[B04]{$g:\mathbb{R}^n\to\mathbb{R}^m$}{Constraint function}
\nomenclature[B05]{$\bar f(\cdot\,;\mathbf{j}),\ \bar g_i(\cdot\,;\mathbf{j})$}{Mini-batch estimators on batch $\mathbf{j}$, e.g.\ $\bar g_i(x;\mathbf{j})=\frac1B\sum_{b=1}^{B}g_i(x,\xi^{(j_b)})$}
\nomenclature[B06]{$\varphi$}{Penalty/barrier function, $\varphi\in\{\varphi_{\mathrm{QL}},\varphi_{\mathrm{QR}}\}$}
\nomenclature[B07]{$\overline{\mathcal{L}}_p(\cdot,\lambda;\mathbf{j})$}{Mini-batched Lagrangian}
\nomenclature[B08]{$\mathcal{L}^{\mathrm{emp}}_p(\cdot,\lambda)$}{Expectation of Lagrangian, $M^{-B}\sum_{\mathbf{j}}\overline{\mathcal{L}}_p(\cdot,\lambda;\mathbf{j})$}
\nomenclature[B09]{$x^k$}{Primal iterate at iteration $k$}
\nomenclature[B10]{$\lambda^k,\ \lambda_i^k$}{Dual variables; $\lambda_i^k$ is associated with constraint $i$ at iteration $k$}
\nomenclature[B11]{$p^k$}{Penalty parameters at iteration $k$}
\nomenclature[B12]{$s^k$}{Prox-centre at iteration $k$}
\nomenclature[B13]{$\bar y^k,\ y^k,\ \zeta^k$}{Stochastic search direction, its conditional mean $\mathbb{E}[\bar y^k\mid\mathcal{F}^k]$, and the noise $\bar y^k-y^k$}
\nomenclature[B14]{$\alpha^k$}{Primal step size; square-summable but not summable}
\nomenclature[B15]{$\gamma^k,\ \kappa^k$}{Dual and penalty averaging coefficients; $\gamma^k,\kappa^k\to1$}
\nomenclature[B16]{$\mu$}{Weight of the Moreau/proximal term}
\nomenclature[B17]{$\delta$}{Prox-centre averaging coefficient, $\delta\in[0,1)$}
\nomenclature[B18]{$\rho$}{Penalty parameter of the ALM updating rule}
\nomenclature[B19]{$\Lambda$}{Upper bound of the dual box, $\lambda^k\in[0,\Lambda]^m$}
\nomenclature[B20]{$\ell_c,\ u_c$}{Lower and upper penalty clipping bounds, $p^k\in[\ell_c,u_c]^m$}
\nomenclature[B21]{$\hat\pi$}{Penalty-updating function, $\hat\pi\in\{\hat\pi^{\mathrm{ALM}}_\rho,\hat\pi^{\mathrm{ADAPT}}\}$}
\nomenclature[B22]{$\odot,\ \oslash$}{Coordinate-wise (Hadamard) product and division}

\nomenclature[C01]{$\mathcal{X}$}{Bounded primal set; a closed ball with a bounded radius $R$}
\nomenclature[C02]{$\hat\lambda,\ \hat p$}{Almost-sure limits of $\lambda^k$ and $p^k$}
\nomenclature[C03]{$\mathcal{D}_{\hat p}(\cdot,\hat\lambda)$}{Conservative field of the potential $-\mathcal{L}^{\mathrm{emp}}_{\hat p}(\cdot,\hat\lambda)$}
\nomenclature[C04]{$G$}{Restriction of $\mathcal{D}_{\hat p}(\cdot,\hat\lambda)$ to $\mathcal{X}$}
\nomenclature[C05]{$\Phi_{\hat p}(\cdot,\hat\lambda)$}{Lyapunov function $\max\{\mathcal{L}^{\mathrm{emp}}_{\hat p}(\cdot,\hat\lambda),\,c\}$}
\nomenclature[C06]{$\partial f(x)$}{Clarke subdifferential of $f$ at $x$ as in \citet{clarke1990optimization}; see \cref{def:clarke}}
\nomenclature[C07]{$\operatorname{Lip}_A(f)$}{Lipschitz constant of $f\mid_A$}
\nomenclature[C08]{$\operatorname{dist}(x,A)$}{Distance from $x$ to the set $A$}
\nomenclature[C09]{$\mathcal{O}$}{Fixed o-minimal structure containing $\mathbb{R}_{\exp}$}
\nomenclature[C10]{$\mathbb{R}_{\mathrm{alg}},\ \mathbb{R}_{\exp}$}{Semi-algebraic and exponential o-minimal structures, $\mathbb{R}_{\mathrm{alg}}\subseteq\mathbb{R}_{\exp}\subseteq\mathcal{O}$}
\nomenclature[C11]{$F:\mathbb{R}^n\rightrightarrows\mathbb{R}^n$}{Set-valued map}

\nomenclature[D01]{$\mathbb{R}^n$}{Euclidean space of dimension $n$}
\nomenclature[D02]{$\mathbb{R}_{+}$}{Set of non-negative real numbers, i.e., $\{x\in\mathbb{R}\mid x\ge0\}$}
\nomenclature[D03]{$\mathbb{R}_{++}$}{Set of strictly positive real numbers, i.e., $\{x\in\mathbb{R}\mid x>0\}$}
\nomenclature[D04]{$\mathbb{N},\ \mathbb{N}_0$}{Positive integers $\{1,2,\dots\}$ and non-negative integers $\{0,1,2,\dots\}$}
\nomenclature[D05]{$\mathbb{B}$}{Closed unit ball}
\nomenclature[D06]{$[a,b]$}{Closed interval}
\nomenclature[D07]{$(a,b)$}{Open interval}
\nomenclature[D8]{$\mathbf{1}_A$}{Indicator function of a set $A$}
\nomenclature[D9]{$\nabla f(x)$}{Gradient of $f$ at $x$}
\nomenclature[D10]{$f\mid_{\mathcal{X}}:\mathcal{X}\to\mathbb{R}$}{Restriction of $f$ to a domain $\mathcal{X}\subseteq\mathbb{R}^n$}
\nomenclature[D11]{$\operatorname{Proj}_{\mathcal{X}}$}{Projection operator onto the set $\mathcal{X}$}
\nomenclature[D12]{$[t]_+$}{Positive part, $[t]_+ := \max\{t,0\} = \operatorname{Proj}_{\mathbb{R}_+}(t)$, applied coordinate-wise}
\nomenclature[D13]{$\operatorname{prox}_f$}{Proximal operator of a function $f$}
\nomenclature[D14]{$\mathcal{R}$}{Set of groups in the fairness tasks; $\mathcal{R}_i$ denotes the $i$-th group}

\setlength{\nomlabelwidth}{3.2cm}
\setlength{\nomitemsep}{0.2cm}
\renewcommand{\nomlabel}[1]{#1\hfill}

\begin{list}{}{\setlength{\leftmargin}{1.5cm}}
\item[] 
\input{nomenclature.nls}
\end{list}



\newpage

\section{Experimental Setup and Implementation details}\label{appendix:experimental_setup}

\subsection{Data and Experiments}

\paragraph{Datasets and problem tasks.} Our fairness-based experiments are based on three datasets.
For the ACSIncome dataset \cite{ding2021retiring}, we use a subset corresponding to the state of Florida, which includes 98,925 samples, and consider the binary classification task of predicting whether an individual's income exceeds \$50{,}000.
For the Dutch Census dataset \cite{inbook_dutch}, which represents aggregated groups of people in the Netherlands and includes 60,000 samples, the binary classification task is to predict whether an individual has a high-level or low-level profession.
The CIFAR-10 dataset~\cite{krizhevsky2009learning} contains 60,000 32x32 color images in 10 different classes, whereas the CIFAR-100 dataset contains 100 different classes with 600 images per class. For both datasets, the task is to learn an accurate classifier of the classes.
PINNs are data efficient, allowing us to sample as many data as needed for our three experiments. We sample 2601 training data points.
The task in PINNs experiments is to approximate a solution to a given differential equation.

\subsubsection{PDEs used for PINNs Experiments}\label{sec:PINNs_appendix}
The \emph{Helmholtz equation} reads:
\begin{align}
    &\Delta u + u = q(z_1,z_2), &&\text{ for every } (z_1, z_2) \in [-1,1]^2, \\
    &u(z_1,z_2) = 0, &&\text{ for every } (z_1,z_2) \in \operatorname{Bd} [-1,1]^2, 
\end{align}
where the function $q$ is chosen as
\begin{align}\label{eq:q_function}
    q(z_1,z_2) &= -\pi^2 \sin(\pi z_1)\sin(4\pi z_2) \nonumber \\
    &- (4\pi)^2 \sin(\pi z_1)\sin(4\pi z_2) + \sin(\pi z_1)\sin(4\pi z_2). \nonumber
\end{align}

The  \emph{viscous Burgers equation} reads: 
\begin{align}\label{burgers_equation}
        &\partial_{t} u + \partial_z(\tfrac{1}{2}u^2 - c\partial_zu) = 0, &&\text{ for } (t,z)\in[0,1]\times[-1,1], \\
        \label{burgers_constr1}
        &u(0,z) = -\sin(\pi z), &&\text{ for } z\in[-1,1], \\
         \label{burgers_constr2}
        &u(t, -1) = u(t, 1) = 0, &&\text{ for } t\in[0,1],
\end{align}
for $(t,z) \in [0,1]\times[-1,1]$, and $c=\frac{0.01}{\pi}$. This setting is the same as in \cite{McClenny_2023}.

The \emph{Klein--Gordon equation} we consider is:
\begin{align}\label{Klein_Gordon_equation}
    &\partial_{t}^2u - \partial_{z}^2u + u^3 = f(t, z), &&\text{ for } (t,z)\in[0,1]\times[0,1], \\
    \label{Klein_Gordon_constr1}
    &u(0, z) = g_1(z), &&\text{ for } z\in[0,1], \\
    \label{Klein_Gordon_constr2}
    &\partial_{t} u(0, z) = g_2(z), &&\text{ for } z\in[0,1], \\
    \label{Klein_Gordon_constr3}
    &u(t,z) = h(t,z), &&\text{ for } (t,z)\in[0,1]\times\{0,1\},
\end{align}
where $f, g_1, g_2, h$ are computed using a fabricated solution, as in \cite{wang2021understanding}:
\begin{equation}
    u(t,z) = z\cos(5\pi t) + (tz)^3. \nonumber 
\end{equation}

\newpage
\subsection{Implementation}

\paragraph{Implementation Details}\label{app:implementation} 
\begin{enumerate}
    \item \emph{(SSL-ALM, SSw implementation)} The implementation of SSw and SSL-ALM are based on the repository \citet{kliachkin2025humancompatible}. 
\item \emph{(PINN implementation)} We base our implementation of the PINN experiments \Exp{6}, \Exp{7}, \Exp{8} on the code repository of \citet{son2023enhanced}. 
\item  \emph{(Primal update step used)} The SPBM algorithm uses Adam as the \texttt{PrimalUpdate} function specified in  Algorithm~\ref{alg:spbm}. 
\item \emph{(SSL-ALM adaptation to our setting)} SSL-ALM \cite{pmlr-v267-huang25au} is designed for equality constraints; we adapt it to our inequality-constrained setting by projecting the dual variables onto the nonnegative orthant.

\item \emph{(Loss functions)} Specifically, the loss function of \Exp{1}--\Exp{5}, for a neural network $u_{\text{nn}}$ with weights $x$, and one-hot label $\operatorname{Target}$, 
\[
 f(x, \xi_j) = -  \sum_{i \in \text{Classes}} \mathbf{1}[\operatorname{Target}(i, j) = 1] \cdot \log \left(  \frac{\exp \bigl( u_{\text{nn}}({x,\xi_j})[i]\bigr)}{\sum_{c \in \text{Classes}} \exp\bigl(u_{\text{nn}}({x,\xi_j})[c]\bigr) } \right).
\]
And the loss function of \Exp{6}--\Exp{8} is:
\[
f(x, \xi_j) = \bigl( N u_{\text{nn}}(x, \xi_j) - h(\xi_j) \bigr)^2.
\]
\item \emph{(Schedules)} To schedule primal learning rate ($\alpha^k$), we use linear warmup over the first 5\% of training steps, followed by a cosine-annealing decay of the learning rate for the remainder~\cite{loshchilov2016sgdr}. We do not implement scheduling neither for the dual learning rate in SSL-ALM nor for the dual update parameters $\gamma, \kappa, \rho$ in SPBM.

\item \emph{(Note on objective and constraint sampling)}\label{paragraph:sampling_constraint} To ensure that we are able to compute constraints at each iteration, we construct minibatches by taking $\frac{B}{|\mathcal{R}|}$ samples from each group independently. This approach produces an unbiased estimate of the constraints as defined in \cref{eq:eqop}; however, this introduces a bias into the objective, as it is defined as the expectation over the entire population. To obtain an unbiased estimator, one could scale each group's samples according to the group's size with respect to the size of the entire dataset. Assume equally-sized batches per demographic groups $b_i$ such that $|b_i| = B/|\mathcal{R}|$ is the number of $R_i$-samples per batch, and $P(R_i) = |R_i|/M$, then define the estimator
\begin{equation}
\begin{aligned}
\overline{f}(x) & =
\sum_{R_i\in\mathcal{R}}
\sum_{l \in b_i} \frac{\left|R_i\right|}{M} \cdot \frac{1}{|b_i|} f(x, \xi_{i, l}) \\ & =
\sum_{R_i\in\mathcal{R}}
P(R_i) \cdot \frac{1}{|b_i|}\sum_{l \in b_i} f(x, \xi_{i,l}).
\end{aligned}
\end{equation}
By the law of total expectation,
$$
\mathbb{E}\big[\bar{f}(x)\big] = \sum_{R_i \in \mathcal{R}} P(R_i)\mathbb{E}_{\xi \sim \operatorname{Unif}(R_i)}[f(x,\xi)] = \mathbb{E}_{\xi}[f(x,\xi)],
$$
hence $\bar{f}(x)$ is an unbiased estimator of the population objective. Note that when the size of each subgroup is equally large, the estimator $\bar f$ coincides with a plain batch average and reweighting has no contribution. We find the size of each subgroup in each of our experiments roughly equally large -- specifically, for CIFAR-10 and CIFAR-100, the size of the groups is exactly the same; for Dutch, the largest group has 3 times the samples of the smallest one (4393 vs. 1488); for Folktables, the largest group has $5.2$ more samples than the smallest one (28697 vs. 5488). We choose to omit the reweighting of the samples in experiments \Exp{2} and \Exp{3} (based on the Folktables and Dutch datasets, respectively) due to the relative simplicity of the task. Therefore, we do \emph{not} implement the reweighting. 



\end{enumerate}



\paragraph{Architectures and epochs.}
To get a reliable estimate of the methods' performance, we conduct five 40-epoch runs with five different seeds per each experiment for \Exp{1}--\Exp{5}; for \Exp{6}--\Exp{8}, we conduct five 6000-epoch runs with different seeds. For experiment \Exp{1}, we use an MLP with two hidden layers of size 64 and 32. For experiment \Exp{2} we use an MLP with 2 hidden layers with sizes 64 and 16, and with sizes 128 and 64 for \Exp{3}. The experiment \Exp{4} is benchmarked using a convolutional neural network with two convolutional layers of respective depths 6 and 16 and a window size of 5, each followed by a 2×2 max-pooling layer, and three fully-connected layers. The experiment \Exp{5}  is benchmarked using the ResNet-18 architecture~\cite{he2016deep}. For experiments \Exp{6}--\Exp{8}, we use a deep narrow MLP with 8 hidden layers of size 64. 

\paragraph{Data split and processing.}
In all experiments, we use the entire dataset for training and hyperparameter selection, omitting a train/test/validation split. We believe that this allows us to fairly compare SPBM against the state-of-the-art purely on optimization performance. In experiments \Exp{1}--\Exp{3}, we preprocess the data to have zero mean and unit variance; for CIFAR experiments \Exp{4, 5} we rescale the data to $[-1,1]$ using default PyTorch tools for CIFAR data. 

\paragraph{Mini-batching}
The mini-batch size varied by problem: for experiments \Exp{1}--\Exp{3} the mini-batch sizes ranged from 32 to 72 samples, whereas for \Exp{4} and \Exp{5} the mini-batch sizes were 120 and 400, respectively. For experiments \Exp{6}--\Exp{8} the mini-batch sizes were set to 1000 samples.

\paragraph{Numerical setup.}\label{numerical_setup} We ran experiments \Exp{1}--\Exp{4}, \Exp{6}--\Exp{8} on a single Tesla V100 GPU; we ran \Exp{5} on a single Tesla H200 141GB NVLink (reference will be added for a camera-ready version).


\paragraph{Choice of hyperparameters.} To determine the hyperparameters, we performed a grid search for each algorithm averaging the results over five runs to account for random initialization; the process in more detail is described in \cref{sec:numexp}.
The hyperparameter grids are listed in Appendix~\ref{hyperparam_gridsearch}, and the final hyperparameters are listed in Appendix~\ref{hyperparam_selection}.

\subsection{Hyperparameter grid search}\label{hyperparam_gridsearch}


In all experiments, we use the PyTorch implementation of the Adam optimizer for the minimization of the objective (Adam, SSw) and of the Lagrangian (SSL-ALM, SPBM).




\paragraph{Fairness experiments (\Exp{1}--\Exp{5}).} The per-method grids are:
\begin{itemize}
    \item \textbf{Adam:} objective learning rate in $\{0.001, 0.005, 0.01, 0.02, 0.05\}$.
    \item \textbf{SSw:} objective and constraint learning rate, each in $\{0.001, 0.005, 0.01, 0.02, 0.05\}$, and Moreau regularization multiplier $\mu$ in $\{0, 1, 2\}$.
    \item \textbf{SSL-ALM:} objective learning rate in $\{0.001, 0.005, 0.01, 0.02, 0.05\}$, dual learning rate in $\{0.01, 0.02, 0.05, 0.1, 0.5\}$, penalty multiplier $\rho \in \{0, 1\}$, and Moreau regularization multiplier in $\{1, 2\}$. 
    \item \textbf{SPBM:} objective learning rate in $\{0.001, 0.005, 0.01, 0.02, 0.05\}$, dual EMA parameter $\gamma$ in $\{0.9, 0.99, 0.999\}$, penalty-barrier function in $\{\varphi_{QL}, \varphi_{QR}\}$, penalty updating function $\hat \pi$ in $\{\hat\pi_{\operatorname{ALM}},\hat\pi_{\operatorname{ADAPT}}\}$~\eqref{penalty_updating_formulas},\eqref{penalty_updating_formulas_alm} and corresponding penalty update multipliers $\kappa \in \{ 0.1, 0.5, 0.9\}$ ($\hat\pi_{\operatorname{ADAPT}}$) and $\rho$ in $\{0.5, 1.0, 2.0\}$ ($\hat\pi_{\operatorname{ALM}}$), and $\mu$ in $\{1, 2\}$.
\end{itemize}

\paragraph{PINN experiments (\Exp{6}--\Exp{8}).} Because PINN training differs substantially from classification in scale and conditioning (long horizons, few minibatches per epoch, stiff residual losses), we use a separate per-regime grid for each method. 

The per-method grids are:
\begin{itemize}
    \item \textbf{Adam:} objective learning rate in $\{0.001, 0.005, 0.01, 0.02, 0.05\}$, and regularization multiplier $\beta$ on the initial/boundary-condition loss terms in $\{0.5, 1, 2, 5, 10\}$.
    \item \textbf{SSw:} objective learning rate in $\{0.001, 0.005, 0.01, 0.02, 0.05\}$ and switching (constraint) learning rate in $\{0.0005, 0.001, 0.005, 0.01, 0.05\}$; no Moreau regularization ($\mu=0$).
    \item \textbf{SSL-ALM:} objective learning rate in $\{0.001, 0.005, 0.01, 0.02, 0.05\}$, dual learning rate in $\{0.0005, 0.001, 0.005, 0.008, 0.01, 0.05\}$, penalty multiplier $\rho \in \{0, 0.1, 1, 2\}$, and Moreau regularization multiplier $\mu$ in $\{0, 1, 2\}$. 
    \item \textbf{SPBM:} objective learning rate in $\{0.001, 0.005, 0.01, 0.02\}$, dual EMA parameter $\gamma$ in $\{0, 0.1, 0.9, 0.999\}$, penalty-barrier function in $\{\varphi_{QL}, \varphi_{QR}\}$, penalty updating function $\hat\pi$ in $\{\hat\pi_{\operatorname{ALM}},\hat\pi_{\operatorname{ADAPT}}\}$ and corresponding penalty update multipliers $\kappa \in \{0.1, 0.9, 0.99\}$ ($\hat\pi_{\operatorname{ADAPT}}$) and $\rho$ in $\{0.1, 1.0, 2.0\}$ ($\hat\pi_{\operatorname{ALM}}$), and $\mu$ in $\{0, 2\}$.
\end{itemize}

\newpage
\subsection{Hyperparameter selection}\label{hyperparam_selection}

In this section, we list the final selected hyperparameters per experiment. Additionally, in all experiments the Quadratic-Logarithmic penalty-barrier function $\varphi_{QL}$ showed better performance as part of the grid, except for \Exp{3}, where the Quadratic-Reciprocal penalty barrier function $\varphi_{QR}$ won out. We mention this here to streamline the tables. LR stands for the learning rate.

\begin{table}[h]
\centering
\label{tab:selected_hparams}
\begin{tabular}{l l c c c c c}
\toprule
Exp. & Method & LR & Dual LR or $\gamma$ & \shortstack{Penalty \\ (ALM: $\rho$)} & \shortstack{Penalty update \\ (SPBM: $\kappa$)} & $\mu$ \\
\midrule
\multirow{4}{*}{\Exp{1}} & Adam    & 0.001 & --    & --   & --    & --    \\
                         & SSw      & 0.001 & 0.01 & --    & --    & 1.0   \\
                         & SSL-ALM  & 0.001 & 0.5 & 0.0   & --    & 2.0   \\
                         & SPBM     & 0.001 & 0.9 & --    & $\hat\pi_{\operatorname{ADAPT}}$, $\kappa = 0.9$  & 2.0   \\
\midrule
\multirow{4}{*}{\Exp{2}} & Adam    & 0.001 & --    & --   & --    & --    \\
                         & SSw      & 0.001 & 0.01 & --    & --    & 0.0   \\
                         & SSL-ALM  & 0.005 & 0.01 & 0.0   & --    & 2.0   \\
                         & SPBM     & 0.005 & 0.9 & --    & $\hat\pi_{\operatorname{ADAPT}}$, $\kappa = 0.9$  & 2.0   \\
\midrule
\multirow{4}{*}{\Exp{3}} & Adam    & 0.005 & --    & --   & --    & --    \\
                         & SSw      & 0.05 & 0.001 & --    & --    & 0.0   \\
                         & SSL-ALM  & 0.02 & 0.01 & 0.0   & --    & 1.0   \\
                         & SPBM     & 0.02 & 0.99 & --    & $\hat\pi_{\operatorname{ADAPT}}$, $\kappa = 0.5$  & 1.0   \\
\midrule
\multirow{4}{*}{\Exp{4}} & Adam    & 0.001 & --    & --   & --    & --    \\
                         & SSw      & 0.001 & 0.001 & --    & --    & 0.0   \\
                         & SSL-ALM  & 0.01 & 0.01 & 1.0   & --    & 2.0   \\
                         & SPBM     & 0.005 & 0.9 & --    &$\hat\pi_{\operatorname{ADAPT}}$, $\kappa = 0.9$ & 2.0   \\
\midrule
\multirow{4}{*}{\Exp{5}} & Adam    & 0.01 & --    & --   & --    & --    \\
                         & SSw      & 0.001 & 0.001 & --    & --    & 1.0   \\
                         & SSL-ALM  & 0.01 & 0.01 & 1.0   & --    & 1.0   \\
                         & SPBM     & 0.005 & 0.999 & --    & $\hat\pi_{\operatorname{ADAPT}}$, $\kappa = 0.9$  & 1.0  \\
\bottomrule
\end{tabular}
\end{table}

\begin{table}[h]
\centering
\label{tab:selected_hparams_pinn}
\begin{tabular}{l l c c c c c}
\toprule
Exp. & Method & LR & Dual LR or $\gamma$ & \shortstack{Penalty \\ (Adam (reg.): $\beta$, \\ SSL-ALM: $\rho$)} & \shortstack{Penalty update \\ (SPBM: $\kappa$ or $\rho$)} & $\mu$ \\
\midrule
\multirow{4}{*}{\Exp{6}} & Adam (reg.)   & 0.005 & --    & 10.0   & --    & --    \\
                         & SSw      & 0.01 & 0.01 & --    & --    & 0.0     \\
                         & SSL-ALM  & 0.01 & 0.05 & 0.1   & --    & 2.0   \\
                         & SPBM     & 0.01 & 0.1 & --    & $\hat\pi_{\operatorname{ALM}}$, $\rho = 2.0$  & 0.0   \\
\midrule
\multirow{4}{*}{\Exp{7}} & Adam  (reg.)  & 0.05 & --    & 10.0   & --    & --    \\
                         & SSw      & 0.005 & 0.0005 & --    & --    & 0.0     \\
                         & SSL-ALM  & 0.005 & 0.01 & 2.0   & --    & 0.0   \\
                         & SPBM     & 0.005 & 0.9 & --    & $\hat\pi_{\operatorname{ADAPT}}$, $\kappa = 0.9$  & 0.0   \\
\midrule
\multirow{4}{*}{\Exp{8}} & Adam  (reg.)  & 0.001 & --    & 0.5   & --    & --    \\
                         & SSw      & 0.001 & 0.01 & --    & --    & 0.0     \\
                         & SSL-ALM  & 0.005 & 0.01 & 0.0   & --    & 0.0   \\
                         & SPBM & 0.005 & 0.0 & --    & $\hat\pi_{\operatorname{ALM}}$, $\rho = 2.0$  & 2.0   \\
\bottomrule
\end{tabular}
\end{table}

\newpage

\section{Remaining Results}\label{appendix:remaining_results}

\subsection{Ablation Study - SPBM}\label{subsec:ablation_study}
We perform an ablation study of SPBM to determine the impact of each component of the method on its performance. We identify $\gamma$, $\kappa$, and $\mu$ as parameters of interest, and run experiments analogous to \Exp{1}--\Exp{8} while setting each to 0. The results for fairness experiments are shown in \cref{tab:best_results_old}, and the results for PINN experiments are shown in \cref{tab:ablation_pinn}.

\begin{table}[h]
\centering
\caption{Results of an ablation study of SPBM on the fairness benchmark. As earlier, we pick the hyperparameter configuration that produces the lowest training loss while being feasible within 10\% tolerance; if no such configurations exist, we pick one with the lowest constraint violation.}
\label{tab:best_results_old}
\begin{tabular}{l l c c c}
\toprule
Exp. & Method & Best loss & Max constraint viol. & Mean constraint \\
\midrule
\midrule
\multirow{5}{*}{\Exp{1}} & SPBM ($\hat{\pi}_\rho^{ALM}$) & 0.373 \footnotesize{$\pm$ 0.001} & \textbf{0.000} \footnotesize{$\pm$ 0.002} & 1.402 \footnotesize{$\pm$ 0.036} \\
 & SPBM ($\hat{\pi}_\rho^{ADAPT}$) & \textbf{0.372} \footnotesize{$\pm$ 0.001} & 0.001 \footnotesize{$\pm$ 0.003} & 1.433 \footnotesize{$\pm$ 0.024} \\
 & SPBM ($\gamma=0$) & \textbf{0.372} \footnotesize{$\pm$ 0.001} & \textbf{0.000} \footnotesize{$\pm$ 0.001} & 1.425 \footnotesize{$\pm$ 0.035} \\
 & SPBM ($\kappa=0$) & 0.374 \footnotesize{$\pm$ 0.001} & \textbf{0.000} \footnotesize{$\pm$ 0.002} & 1.399 \footnotesize{$\pm$ 0.041} \\
 & SPBM ($\mu=0$) & 0.376 \footnotesize{$\pm$ 0.002} & \textbf{0.000} \footnotesize{$\pm$ 0.001} & 1.430 \footnotesize{$\pm$ 0.068} \\

\midrule
\multirow{5}{*}{\Exp{2}} & SPBM ($\hat{\pi}_\rho^{ALM}$) & 0.359 \footnotesize{$\pm$ 0.001} & \textbf{0.000} \footnotesize{$\pm$ 0.002} & 0.000 \footnotesize{$\pm$ 0.002} \\
 & SPBM ($\hat{\pi}_\rho^{ADAPT}$) & \textbf{0.350} \footnotesize{$\pm$ 0.003} & 0.002   \footnotesize{$\pm$ 0.007} & 0.000 \footnotesize{$\pm$ 0.005} \\
 & SPBM ($\gamma=0$) & 0.386 \footnotesize{$\pm$ 0.004} & \textbf{0.000} \footnotesize{$\pm$ 0.009} & 0.000 \footnotesize{$\pm$ 0.006} \\
 & SPBM ($\kappa=0$) & \underline{0.356} \footnotesize{$\pm$ 0.002} & \textbf{0.000} \footnotesize{$\pm$ 0.005} & 0.000 \footnotesize{$\pm$ 0.003} \\
 & SPBM ($\mu=0$) & 0.366 \footnotesize{$\pm$ 0.003} & 0.004 \footnotesize{$\pm$ 0.004} & 0.000 \footnotesize{$\pm$ 0.003} \\

\midrule
\multirow{5}{*}{\Exp{3}} & SPBM ($\hat{\pi}_\rho^{ALM}$) & 0.487 \footnotesize{$\pm$ 0.009} & 0.001 \footnotesize{$\pm$ 0.020} & 0.000 \footnotesize{$\pm$ 0.010} \\
 & SPBM ($\hat{\pi}_\rho^{ADAPT}$) & \textbf{0.467} \footnotesize{$\pm$ 0.003} & 0.001 \footnotesize{$\pm$ 0.004} & 0.000 \footnotesize{$\pm$ 0.006} \\
 & SPBM ($\gamma=0$) & 0.487 \footnotesize{$\pm$ 0.009} & 0.001 \footnotesize{$\pm$ 0.020} & 0.000 \footnotesize{$\pm$ 0.010} \\
 & SPBM ($\kappa=0$) & \underline{0.475} \footnotesize{$\pm$ 0.002} & \textbf{0.000} \footnotesize{$\pm$ 0.006} & 0.000 \footnotesize{$\pm$ 0.004} \\
 & SPBM ($\mu=0$) & \textbf{0.467} \footnotesize{$\pm$ 0.003} & 0.001 \footnotesize{$\pm$ 0.004} & 0.000 \footnotesize{$\pm$ 0.006} \\

\midrule
\multirow{5}{*}{\Exp{4}} & SPBM ($\hat{\pi}_\rho^{ALM}$) & 0.750 \footnotesize{$\pm$ 0.041} & \textbf{0.000} \footnotesize{$\pm$ 0.011} & 0.000 \footnotesize{$\pm$ 0.014} \\
 & SPBM ($\hat{\pi}_\rho^{ADAPT}$) & \textbf{0.016} \footnotesize{$\pm$ 0.004} & \textbf{0.000} \footnotesize{$\pm$ 0.013} & 0.000 \footnotesize{$\pm$ 0.007} \\
 & SPBM ($\gamma=0$) & 2.117 \footnotesize{$\pm$ 0.106} & \textbf{0.000} \footnotesize{$\pm$ 0.024} & 0.000 \footnotesize{$\pm$ 0.025} \\
 & SPBM ($\kappa=0$) & \underline{0.187} \footnotesize{$\pm$ 0.017} & \textbf{0.000} \footnotesize{$\pm$ 0.018} & 0.000 \footnotesize{$\pm$ 0.015} \\
 & SPBM ($\mu=0$) & 0.235 \footnotesize{$\pm$ 0.012} & \textbf{0.000} \footnotesize{$\pm$ 0.052} & 0.000 \footnotesize{$\pm$ 0.028} \\

\midrule
\multirow{6}{*}{\Exp{5}} 
& SPBM ($\hat{\pi}_\rho^{ALM}$) & 4.487 \footnotesize{$\pm$ 0.041} & \textbf{0.000} \footnotesize{$\pm$ 0.029} & 0.000 \footnotesize{$\pm$ 0.018} \\
 & SPBM ($\hat{\pi}_\rho^{ADAPT}$) & \textbf{0.001} \footnotesize{$\pm$ 0.000} & \textbf{0.000} \footnotesize{$\pm$ 0.002} & 0.000 \footnotesize{$\pm$ 0.000} \\
 & SPBM ($\gamma=0$) & 4.604 \footnotesize{$\pm$ 0.001} & \textbf{0.000} \footnotesize{$\pm$ 0.042} & 0.000 \footnotesize{$\pm$ 0.013} \\
 & SPBM ($\kappa=0$) & 0.017 \footnotesize{$\pm$ 0.006} & \textbf{0.000} \footnotesize{$\pm$ 0.012} & 0.000 \footnotesize{$\pm$ 0.005} \\
 & SPBM ($\mu=0$) & \underline{0.002} \footnotesize{$\pm$ 0.000} & \textbf{0.000} \footnotesize{$\pm$ 0.002} & 0.000 \footnotesize{$\pm$ 0.000} \\
 \midrule
 \end{tabular}
\end{table}

\begin{table}[h]
\centering
\caption{Results of an ablation study of SPBM on the PINN benchmark. As earlier, we pick the hyperparameter configuration that produces the lowest training loss while being feasible within 10\% tolerance; if no such configurations exist, we pick one with the lowest constraint violation.}
\label{tab:ablation_pinn}
\begin{tabular}{l l c c c}
\toprule
Exp. & Method & Best loss & Max constraint viol. & Mean constraint \\
\midrule
\midrule
\multirow{5}{*}{\Exp{6}}
 & SPBM ($\hat{\pi}_\rho^{ALM}$) & \textbf{4010.463} \footnotesize{$\pm$ 3274.490} & \textbf{0.0000} \footnotesize{$\pm$ 0.0002} & 0.0001 \footnotesize{$\pm$ 0.0002} \\
 & SPBM ($\hat{\pi}_\rho^{ADAPT}$) & 6680.601 \footnotesize{$\pm$ 7.048} & \textbf{0.0000} \footnotesize{$\pm$ 0.0002} & 0.0001 \footnotesize{$\pm$ 0.0002} \\
 & SPBM ($\gamma=0$) & \underline{5347.308} \footnotesize{$\pm$ 2673.634} & \textbf{0.0000} \footnotesize{$\pm$ 0.0002} & 0.0001 \footnotesize{$\pm$ 0.0002} \\
 & SPBM ($\kappa=0$) & 6684.120 \footnotesize{$\pm$ 0.008} & \textbf{0.0000} \footnotesize{$\pm$ 0.0000} & 0.0000 \footnotesize{$\pm$ 0.0000} \\
 & SPBM ($\mu=0$) & \textbf{4010.463} \footnotesize{$\pm$ 3274.490} & \textbf{0.0000} \footnotesize{$\pm$ 0.0002} & 0.0001 \footnotesize{$\pm$ 0.0002} \\
\midrule
\multirow{5}{*}{\Exp{7}}
 & SPBM ($\hat{\pi}_\rho^{ALM}$) & \textbf{0.000} \footnotesize{$\pm$ 0.000} & \textbf{0.0000} \footnotesize{$\pm$ 0.0000} & 0.0000 \footnotesize{$\pm$ 0.0000} \\
 & SPBM ($\hat{\pi}_\rho^{ADAPT}$) & \textbf{0.000} \footnotesize{$\pm$ 0.000} & \textbf{0.0000} \footnotesize{$\pm$ 0.0000} & 0.0000 \footnotesize{$\pm$ 0.0000} \\
 & SPBM ($\gamma=0$) & \textbf{0.000} \footnotesize{$\pm$ 0.000} & \textbf{0.0000} \footnotesize{$\pm$ 0.0000} & 0.0000 \footnotesize{$\pm$ 0.0000} \\
 & SPBM ($\kappa=0$) & 0.004 \footnotesize{$\pm$ 0.005} & 0.0001 \footnotesize{$\pm$ 0.0001} & 0.0001 \footnotesize{$\pm$ 0.0001} \\
 & SPBM ($\mu=0$) & \textbf{0.000} \footnotesize{$\pm$ 0.000} & \textbf{0.0000} \footnotesize{$\pm$ 0.0000} & 0.0000 \footnotesize{$\pm$ 0.0000} \\
\midrule
\multirow{5}{*}{\Exp{8}}
 & SPBM ($\hat{\pi}_\rho^{ALM}$) & \textbf{0.252} \footnotesize{$\pm$ 0.109} & \textbf{0.0003} \footnotesize{$\pm$ 0.0004} & 0.0002 \footnotesize{$\pm$ 0.0002} \\
 & SPBM ($\hat{\pi}_\rho^{ADAPT}$) & 1815.751 \footnotesize{$\pm$ 3577.152} & 0.0607 \footnotesize{$\pm$ 0.1055} & 0.0352 \footnotesize{$\pm$ 0.0605} \\
 & SPBM ($\gamma=0$) & \textbf{0.252} \footnotesize{$\pm$ 0.109} & \textbf{0.0003} \footnotesize{$\pm$ 0.0004} & 0.0002 \footnotesize{$\pm$ 0.0002} \\
 & SPBM ($\kappa=0$) & 4163.698 \footnotesize{$\pm$ 5079.520} & 0.1199 \footnotesize{$\pm$ 0.1423} & 0.0573 \footnotesize{$\pm$ 0.0691} \\
 & SPBM ($\mu=0$) & \underline{3.079} \footnotesize{$\pm$ 2.051} & \underline{0.0018} \footnotesize{$\pm$ 0.0008} & 0.0009 \footnotesize{$\pm$ 0.0005} \\
\bottomrule

\end{tabular}
\end{table}

\newpage
\subsection{Convergence Plots}\label{subsec:convergence_plots}
\begin{figure}[H]
    \centering
    \begin{minipage}{0.32\linewidth}
        \centering
        \includegraphics[width=\linewidth]{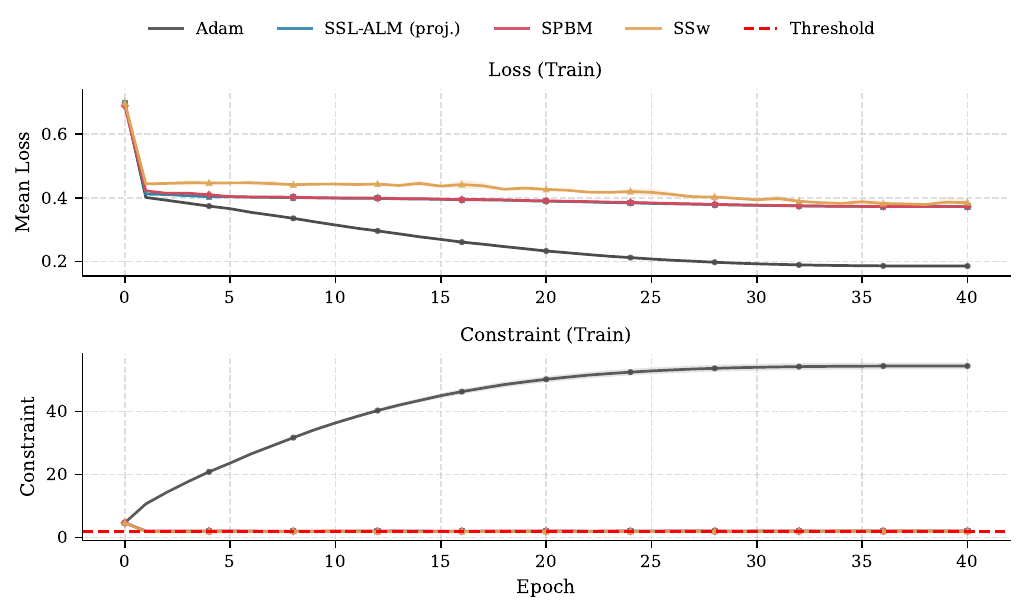}
        \caption{\Exp{1}: Convergence plot.}
        \label{fig:e1_convergence}
    \end{minipage}
    \begin{minipage}{0.32\linewidth}
        \centering
        \includegraphics[width=\linewidth]{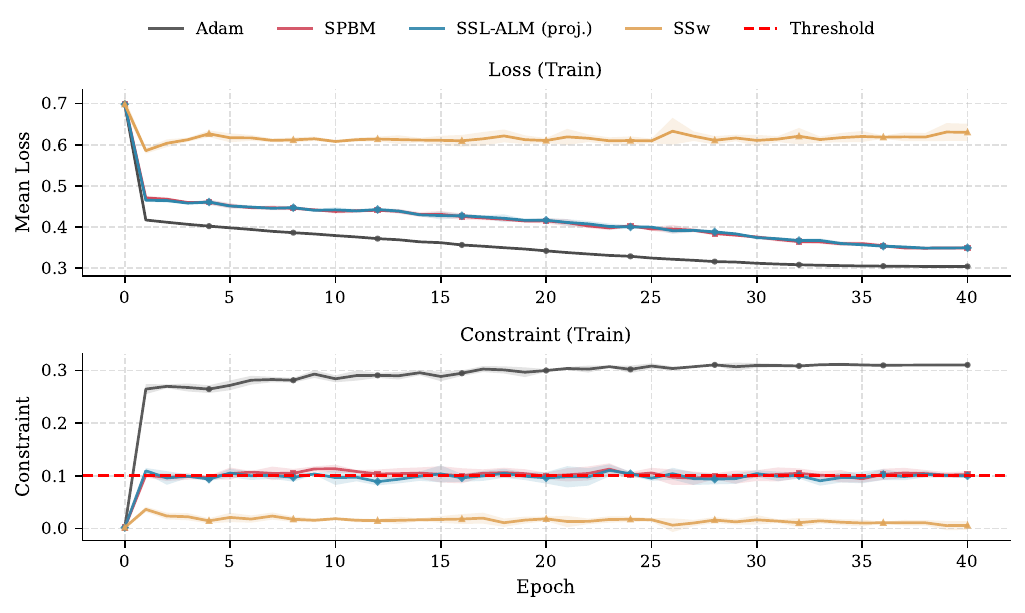}
        \caption{\Exp{2}: Convergence plot.}
        \label{fig:e2_convergence}
    \end{minipage}
    \hfill
    \begin{minipage}{0.32\linewidth}
        \centering
        \includegraphics[width=\linewidth]{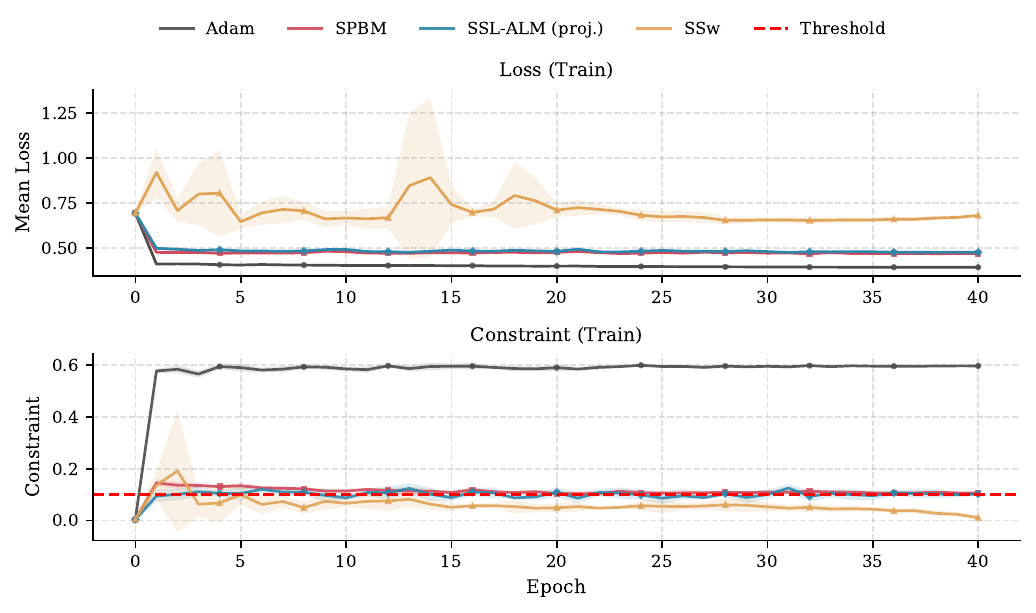}
        \caption{\Exp{3}: Convergence plot.}
        \label{fig:e3_convergence}
    \end{minipage}
\end{figure}

\begin{figure}[H]
\begin{minipage}{0.49\textwidth}
    \includegraphics[width=1.0\columnwidth]{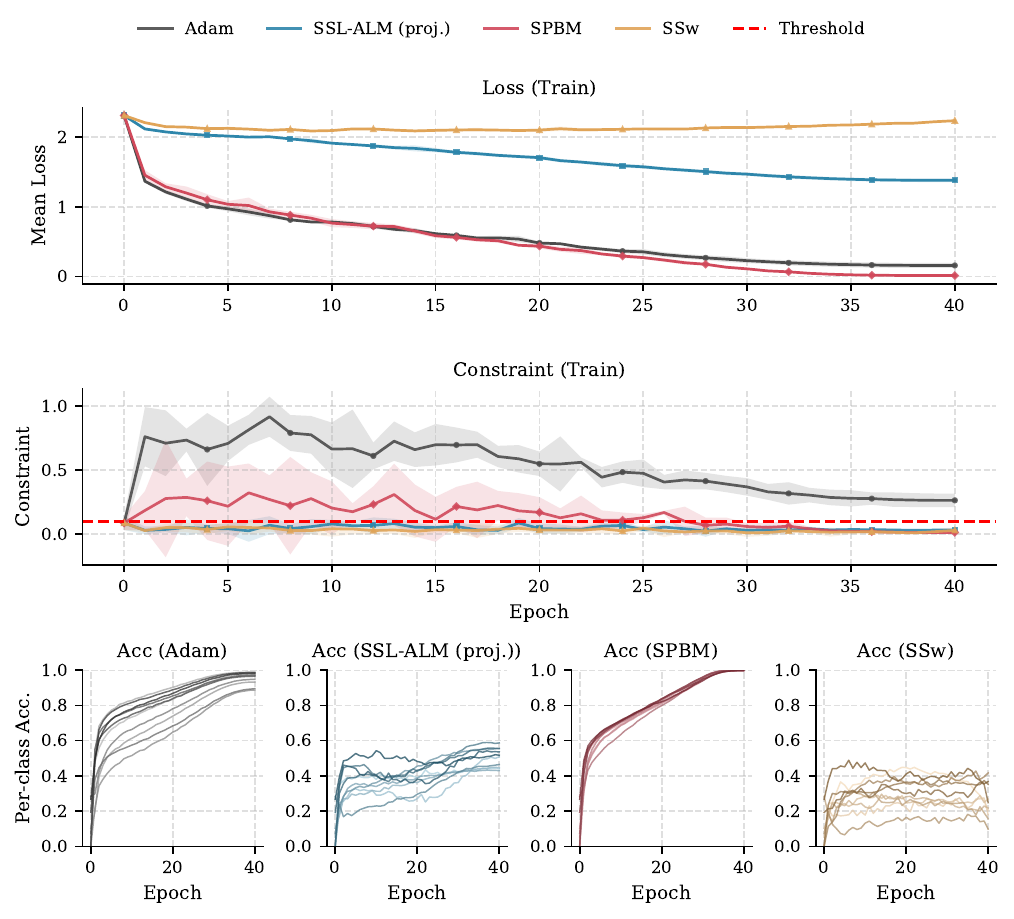}
    \caption{\Exp{4}: Convergence plot. Bottom row shows per-class accuracy convergence (i.e. $10$ accuracy curves per subplot). 
    }\label{fig:e4_convergence}
\end{minipage}
\hfill
\begin{minipage}{0.49\textwidth}
    \includegraphics[width=1.0\columnwidth]{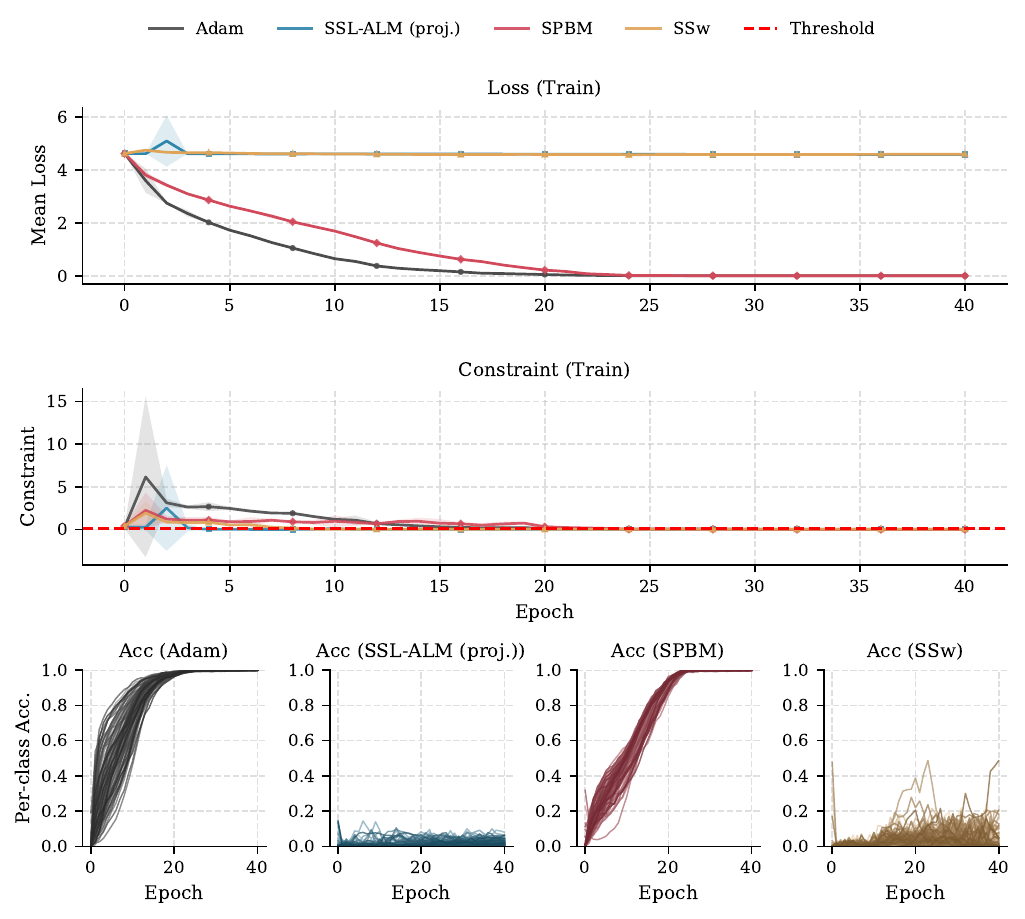}
    \caption{\Exp{5}: Convergence plot. Bottom row shows per-class accuracy convergence (i.e. $100$ accuracy curves per subplot). 
    }\label{fig:e5_convergence}
\end{minipage}
\end{figure}

\begin{figure}[H]
\begin{minipage}{0.32\textwidth}
    \includegraphics[width=1.0\columnwidth]{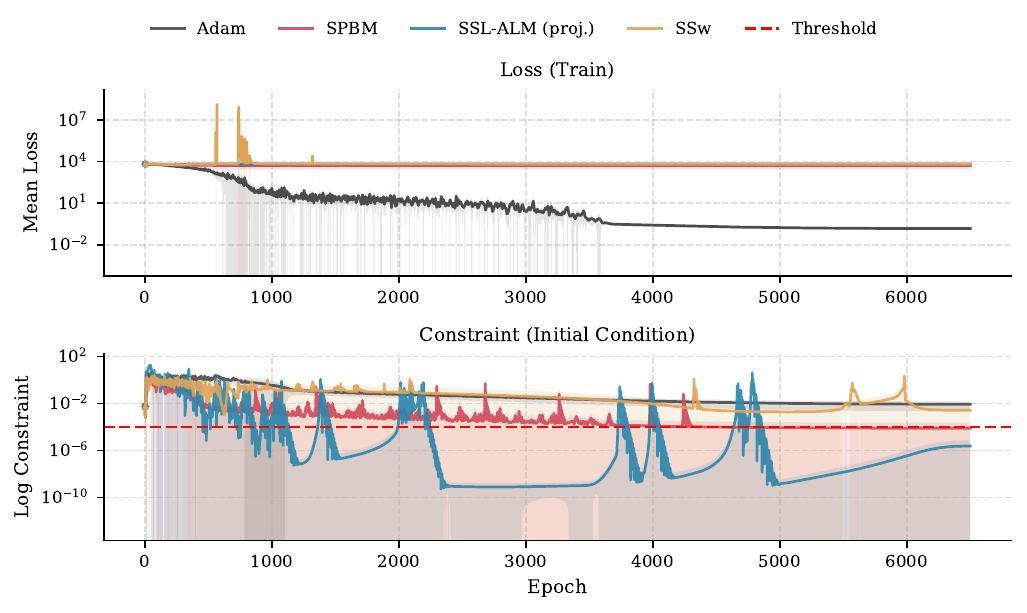}
    \caption{\Exp{6}: Convergence plot.}\label{fig:e7_convergence}
\end{minipage}
\begin{minipage}{0.32\textwidth}
    \includegraphics[width=1.0\columnwidth]{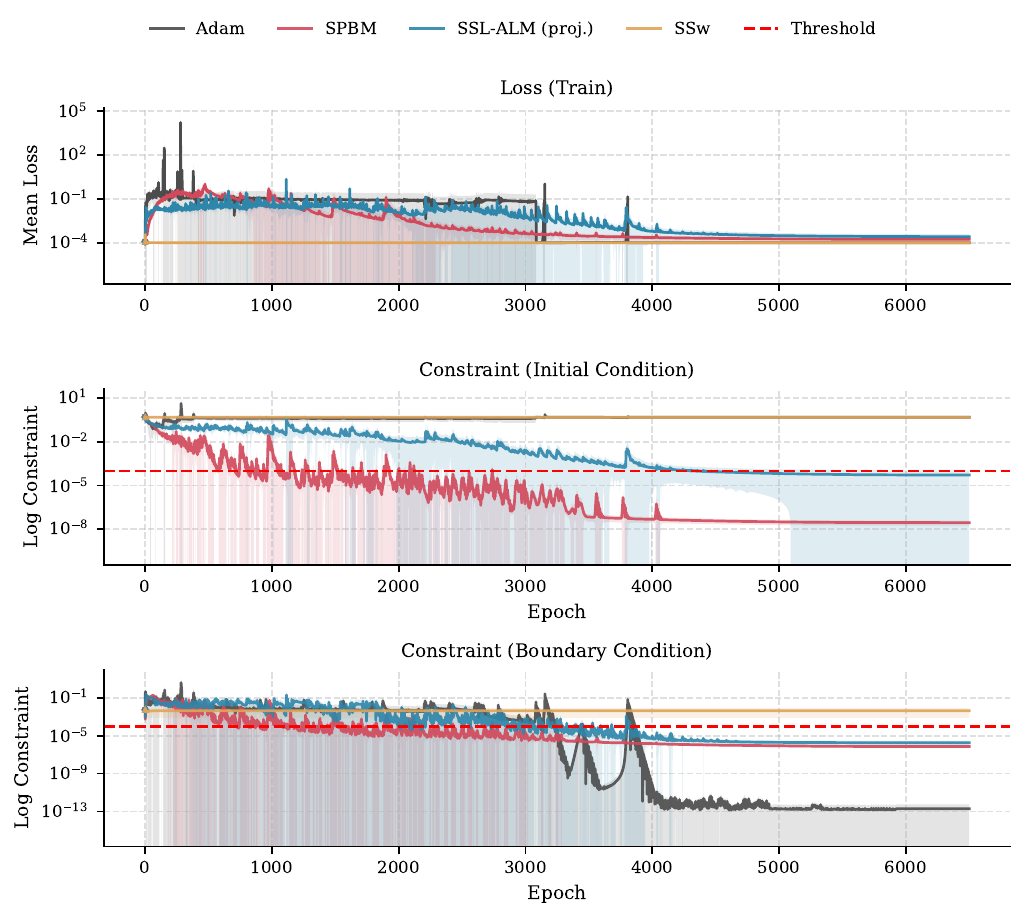}
    \caption{\Exp{7}: Convergence plot.}\label{fig:e8_convergence}
\end{minipage}
\begin{minipage}{0.32\textwidth}
    \includegraphics[width=1.0\columnwidth]{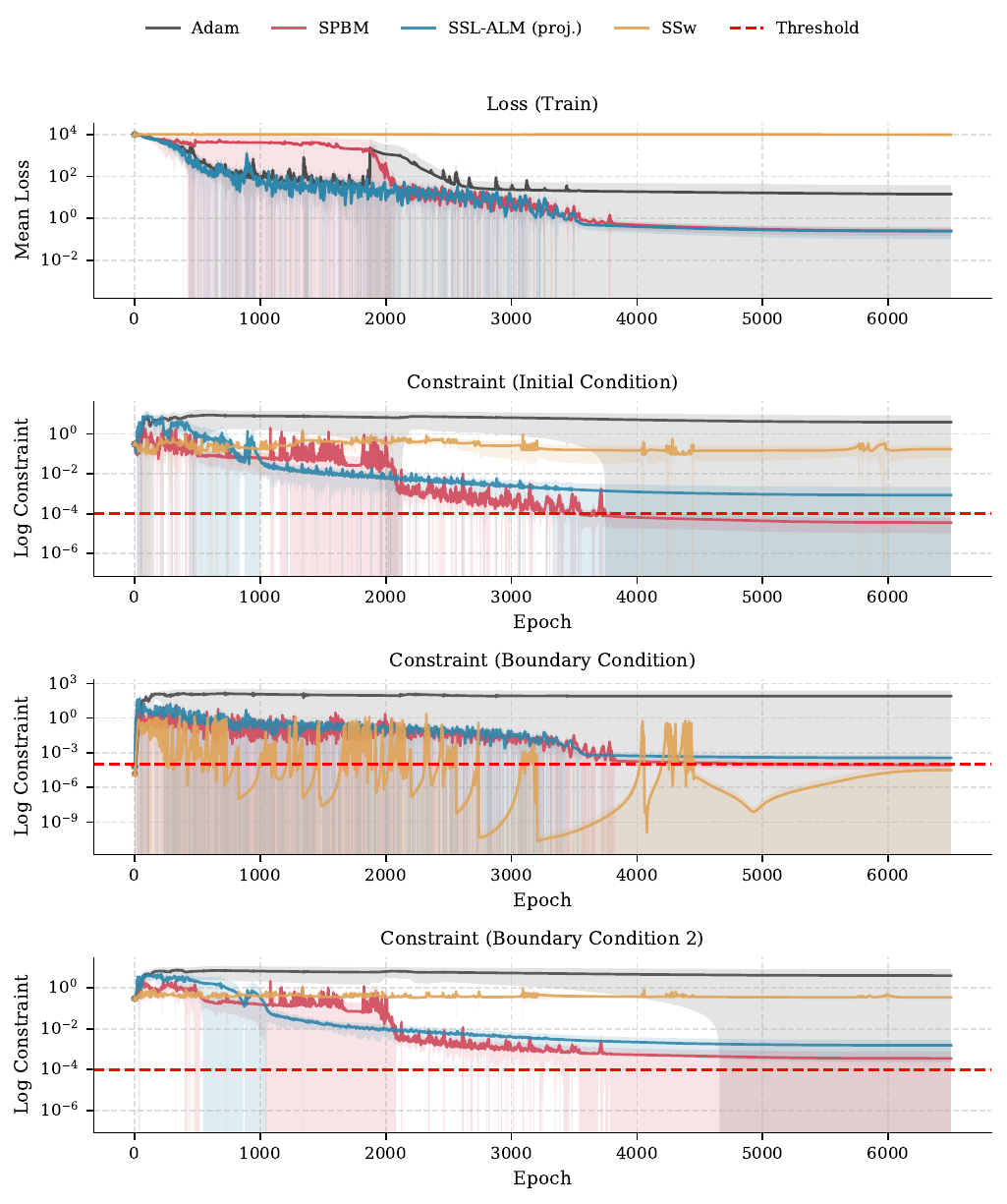}
    \caption{\Exp{8}: Convergence plot.}\label{fig:e9_convergence}
\end{minipage}
\end{figure}

\subsection{Computational Benchmark}

We provide a full table of the computational benchmark in \cref{tab:cifar100_runtime_full}. The shortened version of the table was presented in \cref{tab:cifar100_runtime}.
\begin{table}[H]
\centering
\caption{Computational benchmark based on \Exp{5}: per-epoch runtime (s), mean $\pm$ std over epochs and init seeds.}
\label{tab:cifar100_runtime_full}
\footnotesize
\begin{tabular}{l|lllll}
\toprule
\# constraints ($m$) & Adam & Adam + Reg. & SPBM & SSL-ALM (proj.) & SSw \\
\midrule
100 & 1.74 $\pm$ 0.10 & 1.79 $\pm$ 0.09 & 2.20 $\pm$ 0.01 & 1.98 $\pm$ 0.01 & 1.98 $\pm$ 0.09 \\
1000 & 1.72 $\pm$ 0.01 & 1.84 $\pm$ 0.04 & 2.23 $\pm$ 0.01 & 1.99 $\pm$ 0.01 & 1.95 $\pm$ 0.02 \\
2000 & 1.71 $\pm$ 0.01 & 1.79 $\pm$ 0.02 & 2.23 $\pm$ 0.01 & 1.98 $\pm$ 0.01 & 1.95 $\pm$ 0.02 \\
3000 & 1.71 $\pm$ 0.01 & 1.81 $\pm$ 0.02 & 2.25 $\pm$ 0.01 & 1.98 $\pm$ 0.01 & 1.96 $\pm$ 0.02 \\
4000 & 1.72 $\pm$ 0.01 & 1.81 $\pm$ 0.03 & 2.26 $\pm$ 0.01 & 1.98 $\pm$ 0.01 & 1.96 $\pm$ 0.02 \\
5000 & 1.71 $\pm$ 0.00 & 1.82 $\pm$ 0.03 & 2.26 $\pm$ 0.01 & 1.99 $\pm$ 0.01 & 1.96 $\pm$ 0.03 \\
6000 & 1.71 $\pm$ 0.01 & 1.85 $\pm$ 0.03 & 2.26 $\pm$ 0.01 & 1.99 $\pm$ 0.01 & 1.97 $\pm$ 0.05 \\
7000 & 1.72 $\pm$ 0.01 & 1.82 $\pm$ 0.03 & 2.25 $\pm$ 0.01 & 1.99 $\pm$ 0.01 & 1.96 $\pm$ 0.02 \\
8000 & 1.73 $\pm$ 0.04 & 1.83 $\pm$ 0.05 & 2.26 $\pm$ 0.01 & 1.98 $\pm$ 0.01 & 1.96 $\pm$ 0.03 \\
9000 & 1.72 $\pm$ 0.01 & 1.81 $\pm$ 0.01 & 2.26 $\pm$ 0.01 & 1.99 $\pm$ 0.01 & 1.96 $\pm$ 0.02 \\
9900 & 1.71 $\pm$ 0.00 & 1.84 $\pm$ 0.02 & 2.26 $\pm$ 0.01 & 1.98 $\pm$ 0.01 & 1.96 $\pm$ 0.02 \\
\bottomrule
\end{tabular}
\end{table}

\newpage

\section{Preliminaries}\label{sec:preliminaries}

In this section we give an overview of some of the mathematical ingredients used in the theoretical results in section~\ref{app:theoretical} below. We do not aim to give here an exhaustive account of all prerequisites used; instead we focus on some topics at the interface of \emph{o-minimality} and \emph{nonsmooth nonconvex analysis/optimization}, especially as it relates to the convergence results below.

Section~\ref{tame_theorem_subsection} introduces o-minimality and explains its relevance to our setting. Sections~\ref{Minkowski_sums_subsection}-\ref{conservative_fields_subsection} recall Minkowski sums, Clarke subgradients, and conservative fields, leading to the chain-rule and critical-value properties used in the convergence analysis. Section~\ref{Xu_lemma_subsection} records the definition of weak convexity and an auxiliary sequence convergence lemma.

\subsection{Tame Theory}\label{tame_theorem_subsection}

From a certain birds-eye point of view, the primary objects of study here are objects that live in Euclidean space; e.g., subsets of $\mathbb{R}^n$, functions $\mathbb{R}^n\to\mathbb{R}^m$, and set-valued maps $\mathbb{R}^n\rightrightarrows\mathbb{R}^m$. Indeed, the main convergence result Theorem~\ref{main_theorem} below is ultimately a statement about the expected performance of the algorithm SPBM given the fixed locally Lipschitz functions $\bar f(\cdot;\mathbf{j})$ and
$\bar g_i(\cdot;\mathbf{j})$. From this perspective, the following two comments are in order:
\begin{enumerate}
\item Completely arbitrary locally Lipschitz functions and their corresponding subgradient dynamics can be highly pathological. For example, see~\cite{borwein2000lipschitz,fusek2002examples,daniilidis2020pathological,rios2022examples,daniilidis2024all}.
Consequently, we wouldn't expect Theorem~\ref{main_theorem} to be true in this most general setting, although we make no claims in this regard.
\item We are content with establishing Theorem~\ref{main_theorem} for a very broad class of functions which actually show up in practice. In particular, the functions $\bar{f}$ and $\bar{g}_i$ we have in mind are not arbitrary functions, but instead are functions (NNs) built as a large composition of familiar pieces: affine functions, activation functions such as $\mathsf{ReLU}$, $\mathsf{GeLU}$, and $\mathsf{Mish}$, common loss functions, etc.
\end{enumerate}

To address these, we adopt the framework of \emph{o-minimality}, also called \emph{tame geometry}. This framework arrived in 1984 in~\cite{van1984remarks} and~\cite{pillay1984definable} as a means of extending semialgebraic geometry to broader settings (including a setting containing the exponential function). 

Below our primary reference for connections between o-minimality and deep learning is the survey~\cite{bareilles2025deep}; although these ideas are also discussed in, e.g.,~\cite{DacDruKakLee2018,ji2020directional,bolte2021conservative,kranz2026sad,kinn2026fast,kratsios2026every}.

\begin{definition}\label{omin_structure_def}
A \textbf{structure} on $\mathbb{R}$ is a sequence $\mathcal{O}=(\mathcal{O}_n)_{n\geq 0}$ such that for each $n\geq 0$:
\begin{enumerate}[label=(S\arabic*),leftmargin=*,labelsep=0.6em,align=left]
\item $\mathcal{O}_n$ is a boolean algebra of subsets of $\mathbb{R}^n$;
\item if $A\in\mathcal{O}_n$, then $\mathbb{R}\times A$ and $A\times\mathbb{R}$ belong to $\mathcal{O}_{n+1}$;
\item $\{(x_1,\ldots,x_n)\in\mathbb{R}^n:x_1=x_n\}\in\mathcal{O}_n$;
\item if $A\in\mathcal{O}_{n+1}$, then $\pi(A)\in\mathcal{O}_n$, where $\pi:\mathbb{R}^{n+1}\to\mathbb{R}^n$ is the projection map on the first $n$ coordinates.
\end{enumerate}
A structure $\mathcal{O}$ is \textbf{o-minimal} if it satisfies:
\begin{enumerate}[label=(O\arabic*),leftmargin=*,labelsep=0.6em,align=left]
\item $\{(x,y)\in\mathbb{R}^2:x<y\}\in\mathcal{O}_2$;
\item the sets in $\mathcal{O}_1$ are exactly the finite unions of intervals and points.
\end{enumerate}
Finally, we will always additionally assume that all structures satisfy:
\begin{enumerate}[label=(S\arabic*),start=5,leftmargin=*,labelsep=0.6em,align=left]
\item $\{a\}\in\mathcal{O}_1$ for every $a\in\mathbb{R}$;
\item $\mathcal{O}_3$ contains the graphs of addition and multiplication $+,\cdot:\mathbb{R}^2\to\mathbb{R}$.
\end{enumerate}
Given a structure $\mathcal{O}$ (o-minimal or otherwise) and a set $A\subseteq\mathbb{R}^n$, we say that $A$ is \emph{definable} (in $\mathcal{O}$) if $A\in\mathcal{O}_n$. Likewise, given a function $f:X\to\mathbb{R}^m$ with $X\subseteq \mathbb{R}^n$, we say that the function $f$ is \emph{definable} if its graph $\Gamma(f)\subseteq\mathbb{R}^{n+m}$ is a definable set; a similar convention holds also for set-valued maps $\mathbb{R}^n\rightrightarrows\mathbb{R}^m$.
\end{definition}

The rest of~\ref{tame_theorem_subsection} is organized around following three ideas:
\begin{enumerate}
\item Definability in a structure is preserved by many relevant mathematical constructions (\ref{construction_closure_subsubsection})
\item Many (most?) functions which are built from standard NN primitives are definable in some known o-minimal structure (\ref{common_functions_subsubsection})
\item The o-minimality axiom (O2) has powerful consequences for definable sets, including higher-dimensional definable sets (\ref{omin_consequences_subsubsection})
\end{enumerate}

\subsubsection{Closure under common constructions}\label{construction_closure_subsubsection}

One of the advantages of assuming our Euclidean objects are definable in a common structure $\mathcal{O}$ (as we do in Assumption~\ref{asm:reg} below for $\bar{f}$ and $\bar{g}_i$) is that many related mathematical constructions from the objects will also be definable in $\mathcal{O}$. To give some basic examples of what we mean by that, we record the following facts from~\cite[1.\S2]{van1998tame}: Suppose $\mathcal{O}$ is a structure, then:
\begin{itemize}
\item if functions $f:X\to Y$ and $g:Y\to Z$ are definable, then so is the composition $g\circ f:X\to Z$;
\item if $A\subseteq\mathbb{R}^m$ and $B\subseteq\mathbb{R}^n$ are definable, then so is the cartesian product $A\times B$;
\item if $f:\mathbb{R}\to\mathbb{R}$ is definable, then so is the set $\mathcal{R}_f:=\{x\in\mathbb{R}:\text{$f$ is differentiable at $x$}\}$, as well as the function $x\mapsto f'(x):\mathcal{R}_f\to\mathbb{R}$.
\end{itemize}
More generally, we quote from~\cite{Ioffe2009}:
\begin{quote}
\emph{...the class [of definable sets and functions] is closed with respect to practically all operations used in variational analysis...}
\end{quote}

Why do \emph{structures} support such a large amount of mathematical constructions? The short answer is that the S-axioms of Definition~\ref{omin_structure_def} are expressive enough to permit a vast arsenal of mathematical constructions from their judicious application. A more complete answer is that Definition~\ref{omin_structure_def} encodes the geometric/semantic content of the concept \emph{definability in first-order logic}; namely, a set is definable if and only if there is a valid \emph{first-order formula} which defines it. A first-order formula is a syntactic object built in a disciplined way from variables $x,y,z,\ldots$, primitive relations and functions ($=,+,\cdot,\ldots$), logical connectives $\land,\lor,\neg,\ldots$, and first-order quantifiers $\forall x,\exists x$ (where $x$ may only range over $\mathbb{R}$). 

A rigorous definition and overview of this connection is outside the scope of this paper; our main reference is~\cite[Appendix B]{aschenbrenner2015asymptotic}; see also standard texts in first-order logic and model theory, e.g., ~\cite{enderton2001mathematical,hils2019first,marker2002model}.

Regardless, we can already see how the axioms relate to their syntactic counterparts, e.g.,
\begin{itemize}
\item Axiom (S1) reflects the fact that we have the basic logical connectives $\land,\lor,\neg,\ldots$ at our disposal.
\item Axioms (S3) and (S6) give us the basic primitives $=,+,\cdot$.
\item Axiom (S4) expresses that we may do existential quantification $\exists x_{n+1}$ over the last variable; in fact, quantification over any of the variables can be done, although this requires a small argument.
\end{itemize}

Here is a typical example of the usage of first-order formulas:

\begin{example}
Suppose $\mathcal{O}$ is an arbitrary structure satisfying (S1)-(S4), (S5)-(S6). Then $\mathcal{O}$ also satisfies (O1). Indeed, we may define the binary relation $<$ with the following first-order formula:
\[
\varphi(x,y) \ := \ \exists z\big[\neg(x=y) \land x+z^2=y\big]
\]
We claim that $\varphi$ is a valid first-order formula which defines $<$, which thus establishes that axiom (O1) holds for $\mathcal{O}$. Implicit here is the claim that the theory of first-order definability permits us to ``compile down'' the formula into an argument which only uses the assumed axioms (S1)-(S4), (S5)-(S6).
\end{example}

Thus in practice, in order to identify a construction as being definable, one usually needs to simply write the definition of the construction as a valid first-order formula. This requires learning the careful rules of \emph{first-order definability} and some practice, but otherwise it is usually a routine matter. The constructions we are most concerned with come from topology, differential calculus, and variational analysis; for this, the following insight usually suffices:
\begin{itemize}
\item if a definition involves quantifying over infinite sequences $x_k$, an equivalent $\varepsilon$-$\delta$-style definition should be used instead since first-order quantifiers $\forall x,\exists x$ cannot quantify over sequences.
\end{itemize}

Specifically, we make claims that definability is preserved by a given construction in
Lemmas~\ref{definability_of_Minkowski_sum},~\ref{definability_Clarke_subgradient},~\ref{lem:reg}, and~\ref{proposition_definability_D_uppersem}, and Proposition~\ref{prop:counterexample_is_stosto}; the goal of the preceding discussion was to give some intuition as to \emph{why} such claims ought to be true.

We conclude by emphasizing that nothing in the preceding discussion in~\ref{construction_closure_subsubsection} required or used the o-minimality axiom (O2).

\subsubsection{Definability of functions used in practice}\label{common_functions_subsubsection}

We now briefly survey three o-minimal structures which are perhaps most relevant to the NN examples we have in mind:
\[
\mathbb{R}_{\operatorname{alg}} \ \subseteq \ \mathbb{R}_{\exp} \ \subseteq \ \mathbb{R}_{\operatorname{Pfaff}}
\]

\begin{example}[$\mathbb{R}_{\operatorname{alg}}$ - the semialgebraic sets]
The structure $\mathbb{R}_{\operatorname{alg}}$ can be defined as the smallest structure satisfying axioms (S1)-(S4) and (S5)-(S6). This structure is easily seen to also satisfy (O1) and it is essentially a theorem of~\cite{Tarski48} that this structure is o-minimal, i.e., satisfies (O2). The definable sets of $\mathbb{R}_{\operatorname{alg}}$ are those which can be defined by boolean combinations of sets of the form $\{f(x)\diamond 0\}$, where $f(x)$ is a multivariable polynomial and $\diamond$ is either $=$ or $<$. Consequently, any NN built as a composition of affine layers and semialgebraic activations (such as $\mathsf{ReLU}$ and $\mathsf{Softsign}$) will be semialgebraic; likewise, the barrier function $\varphi_{\operatorname{QR}}$ is also semialgebraic.
\end{example}

As it turns out, no restriction of the real-valued functions $\exp$ or $\log$ will be semialgebraic. For this we turn to:

\begin{example}[$\mathbb{R}_{\exp}$ - the subexponential sets]
The structure $\mathbb{R}_{\exp}$ can be defined as the smallest structure expanding $\mathbb{R}_{\operatorname{alg}}$ and containing the (graph of) the real exponential function $\exp:\mathbb{R}\to\mathbb{R}$. The o-minimality of $\mathbb{R}_{\exp}$ was established by~\cite{wilkie1996model}. In $\mathbb{R}_{\exp}$, the barrier function $\varphi_{\operatorname{QL}}$ becomes definable, as does any NN which now uses activations involving $\exp$ and $\log$ (e.g., $\operatorname{tanh}$, $\mathsf{Logistic}$, $\mathsf{Swish}$, and $\mathsf{Softplus}$).
\end{example}

One notable drawback of $\mathbb{R}_{\exp}$ is that the activation function $\mathsf{GELU}(x)=\tfrac{x}{2}\bigl(1+\operatorname{erf}(x/\sqrt{2})\bigr)$ is not definable; this is because the \emph{error function} $\operatorname{erf}(t)=\tfrac{2}{\sqrt{\pi}}\int_0^t e^{-s^2}\,ds$ is not definable. For this we turn to:

\begin{example}[$\mathbb{R}_{\operatorname{Pfaff}}$ - the Pfaffian closure of $\mathbb{R}_{\operatorname{alg}}$]
This is a further expansion of $\mathbb{R}_{\exp}$ whose o-minimality was shown in~\cite{Speissegger1999Pfaff}. We forego the definition here; for our purposes it suffices to know that some additional activation functions such as $\arctan$ and $\mathsf{GELU}$ become available here.
\end{example}

We also have the following main obstruction to being not-definable in an o-minimal structure:

\begin{nonexample}[Infinite oscillation]
If an activation function exhibits infinite oscillation, then it cannot be definable in any o-minimal structure. For example, if the sine function $\sin:\mathbb{R}\to\mathbb{R}$ were definable, then so would the set $\{x:\sin(x)=0\}$ of zeros; although this clearly contradicts the o-minimality axiom (O2). Even in this case, if the sine function is only being evaluated on a fixed compact set (e.g., $[0,2\pi]$), then the restriction would still be definable in the o-minimal structures $\mathbb{R}_{\operatorname{an}}\subseteq\mathbb{R}_{\operatorname{an},\exp}$ (see e.g.,~\cite{van1994elementary} for definitions of these structures).
\end{nonexample}

In conclusion, as a rule of thumb, if a function is constructed from familiar constructions alluded to in~\ref{construction_closure_subsubsection} above, using familiar pieces which do not exhibit infinite oscillation, then most likely that function will be definable in some known o-minimal structure. This is the basis for our belief that \emph{definable in an o-minimal structure} is an appropriate hypothesis for our theoretical results. We make no general claims about the ubiquity of this hypothesis in practice, although as a heuristic we point to~\cite[Remark 3.19]{bareilles2025deep} where it is estimated that around 89\% of the activation functions listed in the survey~\cite{kunc2024decadesactivationscomprehensivesurvey} are definable in some known o-minimal structure.

\subsubsection{Consequences of the o-minimality axiom}\label{omin_consequences_subsubsection}

To avoid giving the wrong impression, we emphasize the following:

\begin{quote}
\emph{Although it might be trivial to recognize that a given object is definable in a known o-minimal structure $\mathcal{O}$; the proof of o-minimality of $\mathcal{O}$ is non-trivial.}
\end{quote}

In this subsection we briefly address the question:

\begin{quote}
\emph{What can be gained from knowing an object is definable in an o-minimal structure?}
\end{quote}

The short answer is that an o-minimal structure gives rise to a coherent world of Euclidean objects free from undesirable pathologies (e.g., infinite oscillation, or worse): where all definable objects exhibit tame asymptotic and finiteness properties, and the world is closed under all first-order mathematical constructions.

As a simple example, we mention that \emph{non-convergence} is impossible in the following sense:
\begin{itemize}
\item if $f:[0,\infty)\to\mathbb{R}^n$ is a bounded curve definable in an o-minimal structure, then $\lim_{t\to\infty}f(t)$ always exists (cf.~\cite[3.\S1]{van1998tame}).
\end{itemize}

For ``higher dimensional'' definable sets in $\mathbb{R}^n$ ($n\geq 1$), we have the following:
\begin{itemize}
\item Every definable subset $X\subseteq\mathbb{R}^n$ can be assigned a unique and well-defined \emph{dimension} $\dim X$, where \emph{dimension} satisfies many reasonable properties of a \emph{dimension function}~\cite[4.\S1]{van1998tame},~\cite[3.34]{bareilles2025deep}; this dimension agrees with the usual notions of dimension where applicable, e.g., linear dimension and manifold dimension.
\item For every $r\geq 1$, every definable subset $X\subseteq\mathbb{R}^n$ can be partitioned into finitely many definable $C^r$-submanifolds of $\mathbb{R}^n$. There are various results of this form, most notably \emph{smooth cell decomposition}~\cite[7.\S3]{van1998tame} and various \emph{stratification theorems} (e.g., Whitney-(a)/(b), Verdier); see~\cite{le1998verdier}.
\end{itemize}

For our purposes, we point out that the stratification results enable~\cite{bolte2007clarke} to prove the \emph{projection formula} for the set-valued \emph{Clarke subgradient}, and this was key for~\cite{DacDruKakLee2018} to prove the convergence of the \emph{stochastic subgradient method} (SSM). This was later extended to the set-valued notion of \emph{conservative field} in~\cite{bolte2021conservative}, partially to justify convergence even though standard AD implementations such as PyTorch and TensorFlow need not always return an element of the Clarke subgradient.

Our convergence proof (Theorem~\ref{main_theorem}) follows this same playbook.
Namely, we first define an enveloping subgradient-like set-valued map (see Lemma~\ref{proposition_definability_D_uppersem}):
\[
\mathcal{D}_{p}(\cdot, \lambda) \ : \  \mathbb{R}^n \rightrightarrows \mathbb{R}^n.
\]
Next, using the tame stratification machinery outlined above, we show that this map is in fact a \emph{definable conservative field} in the sense of~\citet{bolte2021conservative} (cf.~Lemmas~\ref{lem:conservative_field_clarke} and~\ref{proposition_definability_D_uppersem}). This permits us to adapt the main stochastic approximation / differential inclusion convergence arguments of~\citet{DacDruKakLee2018} and~\citet{bolte2021conservative}; after verifying a descent property and a Weak Sard property (via Theorem~\ref{thm:morse_sard}), we can conclude Theorem~\ref{main_theorem}.


We conclude by emphasizing that everything in the preceding discussion in~\ref{omin_consequences_subsubsection} assumed the o-minimality axiom (O2). When dropping this axiom, cracks appear in the world just described.

\subsection{Minkowski sums}\label{Minkowski_sums_subsection}

Minkowski sums provide the notation for combining set-valued subgradients, in particular when averaging over mini-batches in Section~\ref{sec:convergence_analysis}.

\begin{definition}[Minkowski Sum]
    The Minkowski sum of sets $A,B\subseteq\mathbb{R}^n$ is defined as 
    \[
    A + B = \{ a + b \mid a \in A, b \in B \}.
    \]
\end{definition}

Constructing the Minkowski sum preserves definability:

\begin{lemma}\label{definability_of_Minkowski_sum}
If $A,B\subseteq\mathbb{R}^n$ are definable in a structure, then so is the Minkowski sum $A+B$.
\end{lemma}
\begin{proof}[Proof sketch]
Let $\varphi_A(x)$, $\varphi_B(y)$ be first-order formulas which define $A$ and $B$ in $\mathcal{O}$. Then the following formula defines $A+B$ in $\mathcal{O}$:
\[
\varphi_{A+B}(z) \ := \ \exists x\exists y\big[\varphi_A(x) \ \land \ \varphi_B(y) \ \land \ x+y=z\big]
\]
We emphasize that this lemma and the argument do not depend on the o-minimality axiom.
\end{proof}

We also need the following two facts about Minkowski sums: 

\begin{lemma}\label{lem:minkowski_preserves_inclusion}
    Minkowski sum preserves inclusion, if $A_j\subseteq B_j\subseteq\mathbb{R}^n$ for $j=1,\ldots,N$, then 
    \[
    \sum_{j=1}^N A_j \subseteq \sum_{j=1}^N B_j.
    \]
\end{lemma}
\begin{proof}
This is an instance of the following basic fact about functions: given a function $f:X\to Y$ and an inclusion $A\subseteq B\subseteq X$ of subsets of the domain, we have an inclusion $f[A]\subseteq f[B]$ of direct images in the codomain. In this case, the function $f:=\sum_{j=1}^N:\mathbb{R}^{nN}\to\mathbb{R}^{n}$ is multilinear vector addition, and the domain inclusion is $A:=A_1\times\cdots\times A_N\subseteq B:=B_1\times\cdots\times B_N$.
\end{proof}

\begin{lemma}\label{lem:minkowski_ball}
    Let $C\subseteq\mathbb{R}^n$ be convex, $N\geq 1$, and $c_1,\dots,c_N\ge 0$. Then
    $\sum_{j=1}^{N}c_j C=\bigl(\sum_{j=1}^{N}c_j\bigr)C$.
    In particular, $\tfrac1N\sum_{j=1}^{N}K\,\mathbb{B}=K\,\mathbb{B}$ for $K\ge0$.
\end{lemma}
\begin{proof}
    The inclusion $\bigl(\sum_j c_j\bigr)C\subseteq\sum_j c_j C$ holds for any set, since
    $\bigl(\sum_j c_j\bigr)u=\sum_j c_j u$ for $u\in C$. Conversely, let
    $v=\sum_j c_j u_j$ with $u_j\in C$ and put $s:=\sum_j c_j$. If $s=0$ then $v=0$ and the
    claim is trivial; otherwise $v=s\sum_j (c_j/s)u_j$, and $\sum_j (c_j/s)u_j\in C$ because
    the coefficients $c_j/s$ are nonnegative and sum to one and $C$ is convex.
\end{proof}

\subsection{The Clarke subgradient}\label{Clarke_subgradient_subsection}
\emph{In this subsection we fix a locally Lipschitz function $f:\mathbb{R}^n\to\mathbb{R}$.}

An arbitrary locally Lipschitz function need not have a well-defined gradient at every point. However, Rademacher's theorem (see \cite{nekvindaSimpleProofRademacher1988} for the statement with simple proof) guarantees that the set:
\[
\mathcal{R}_f := \{x\in\mathbb{R}^n:\text{$f$ is differentiable at $x$}\} \subseteq \mathbb{R}^n
\]
has full measure, i.e., the complement $\mathbb{R}^n\setminus\mathcal{R}_f$ has Lebesgue measure zero. In particular, at a given $x\in\mathbb{R}^n$, the gradient of $f$ exists at almost every point near $x$; this motivates considering the set-valued \emph{Clarke subgradient} $\partial f(x)$ of $f$ at $x$ since the actual gradient $\nabla f(x)$ might not actually exist:

\begin{definition}[Clarke subgradient]\label{def:clarke}
The \emph{Clarke subgradient} $\partial f(x)$ of $f$ at $x$ is the set:
\[
  \partial f(x)=\operatorname{conv}
  \Bigl\{\,v\in\mathbb{R}^n,\ \exists\, x^k\underset{k\to\infty}{\to}x
  \ \text{with}\ x^k\in\mathcal{R}_f,\ v_k=\nabla f(x_k)\underset{k\to\infty}{\to}v \Bigr\} \subseteq \mathbb{R}^n
\]
Thus we may construe the Clarke subgradient of $f$ as a set-valued map $\partial f:\mathbb{R}^n\rightrightarrows\mathbb{R}^n$.
\end{definition}

We next recall two continuity notions for set-valued maps that will be used when passing to limits in subgradient inclusions.

\begin{definition}[Outer-Semicontinuity by \citet{rockafellar2009variational}]
    A set-valued map $G: \mathcal{X} \to \mathbb{R}^n$ is outer-semicontinuous at a point $x \in \mathcal{X}$ if for any sequence $x^{i} \to x$ and $v^{i} \in G(x^{i})$ converging to a vector $v \in \mathbb{R}^n$, it holds $v \in G(x)$.
\end{definition}

Outer semicontinuity on $\mathcal{X}$ is equivalent to the graph of $G$ being closed in $\mathcal{X}\times\mathbb{R}^n$.

\begin{definition}[Upper-Semicontinuity by \citet{clarke1990optimization}]
Let $G: X \to Y$ be a set-valued map; when $X, Y$ are Banach spaces, the upper-semicontinuity property at $x$ is defined as: for all $\varepsilon > 0$, there exists $\eta > 0$ such that
\[
G(x') \subseteq G(x) + \varepsilon \mathbb{B}_Y ,\quad \forall x' \in x + \eta \mathbb{B}_X.
\]
\end{definition}

Here are the basic properties of the Clarke subgradient:

\begin{lemma}\label{properties_of_Clarke_subgradient}
$\partial f$ has nonempty compact convex values and is locally bounded, outer semicontinuous, and upper semicontinuous. More precisely, if $f$ is $L$-Lipschitz on an open set $U\subseteq\mathbb{R}^n$, then
\[
\partial f(x)\subseteq L\mathbb{B}\quad\text{for every $x\in U$,}
\]
where $\mathbb{B}$ is the closed Euclidean unit ball. These conclusions require no definability assumption.
\end{lemma}
\begin{proof}
Nonemptiness, compactness, convexity, and the stated bound follow from~\cite[Proposition 2.1.2(a)]{clarke1990optimization}; applying the bound on a Lipschitz neighbourhood of each point gives local boundedness. Outer semicontinuity and uppersemicontinuity follow from~\cite[Proposition 2.1.5(b) and (d)]{clarke1990optimization}, respectively.
\end{proof}

\begin{lemma}\label{definability_Clarke_subgradient}
If $f:\mathbb{R}^n\to\mathbb{R}$ is definable in a structure $\mathcal{O}$, then so is its Clarke subgradient $\partial f$.
\end{lemma}
\begin{proof}[Proof idea]
This is a routine exercise in \emph{first-order definability}; see~\cite[Lemma B.8]{ji2020directional} for an argument. We also emphasize here that the lemma and its proof do not depend on the o-minimality axiom (O2).
\end{proof}

When $f$ (hence also $\partial f$) is definable in an \emph{o-minimal} structure, then we can say much more about the properties of $\partial f$ (cf. Lemma~\ref{lem:conservative_field_clarke} below); for this it is convenient to first discuss the more general notion of \emph{conservative fields}.

\subsection{Conservative Fields}\label{conservative_fields_subsection}

A conservative field of a function $f$ is a set-valued mapping which generalizes the notion of derivates (such as Clarke subgradients \cite{clarke1990optimization}) introduced in \citet{bolte2021conservative}. From the practical point of view, as stated in \citet{bolte2021conservative}, Automatic Differentiation (AD) is proven to yield conservative fields, making the idea of conservative fields a very useful tool.

\begin{definition}[Conservative Fields, Definition 1 \cite{bolte2021conservative}]
    Let $\mathcal{D}: \mathbb{R}^n \to \mathbb{R}^n$ be a set-valued map. $\mathcal{D}$ is a conservative (set-valued) field whenever it is outer-semicontinuous, has nonempty compact values, and for any absolutely continuous loop $x: [0, 1] \to \mathbb{R}^n$ such that $x(0) = x(1)$, we have
    \[
    \int_0^1 \max_{v \in \mathcal{D}(x(t))} \langle \dot x(t), v \rangle dt = 0
    \]
    or equivalently
    \[
     \int_0^1 \min_{v \in \mathcal{D}(x(t))} \langle \dot x(t), v \rangle dt = 0.
    \]
\end{definition}

Conservative fields determine an underlying function (unique up to a constant), which \citet{bolte2021conservative} call a \emph{potential function}.
\begin{definition}[Definition 2 \citet{bolte2021conservative}]\label{def:potential_function}
    Let $\mathcal{D}: \mathbb{R}^n \rightrightarrows \mathbb{R}^n$ be a conservative field. For any absolutely continous curve $x$ with $x(0) = 0, x(1) = s$, a function $\Phi$ defined as 
    \[
    \Phi(s) - \Phi(0) = \int_0^1\langle \dot x(t), \mathcal{D}(x(t)) \rangle dt
    \]
    is called a potential function for $\mathcal{D}$, or $\mathcal{D}$ is a conservative field for $\Phi$.
\end{definition}

\begin{remark}
    There can be many conservative fields for a given potential function, whereas, a conservative field determines a unique potential function up to a constant. 
\end{remark}

Also, if a function is a potential function of some conservative field, that function is called \emph{path differentiable}; this terminology is motivated by the following: 

\begin{lemma}[Lemma 2 \citet{bolte2021conservative}]\label{lem:chain_rule}
    Let $\mathcal{D}: \mathbb{R}^n \rightrightarrows \mathbb{R}^n$ be a locally bounded, outer-semicontinuous set valued map and $\Phi: \mathbb{R}^n \to \mathbb{R}$ be a locally Lipschitz function. Then $\mathcal{D}$ is a conservative field for $\Phi$, if and only if for any absolutely continous curve $x: [0, 1] \to \mathbb{R}^n$, the function $t \mapsto \Phi(x(t))$ satisfies for almost all $t \in [0, 1]$
    \[
    \frac{d}{dt} \Phi(x(t)) = \langle v, \dot x(t) \rangle \quad \forall v \in \mathcal{D}(x(t)).
    \]
\end{lemma}

The following lemma is the key step in the proof of the descent property. It is a restatement of \citep[Lemma~5.2]{DacDruKakLee2018} for an arbitrary set-valued map $G$ admitting a chain rule for $f$, rather than for the Clarke subdifferential; the argument is theirs.
\begin{lemma}\label{lem:general_lemma_chain_rule_dist_eq}
Let $\mathcal{X}\subseteq\mathbb{R}^n$, let $G:\mathcal{X}\rightrightarrows\mathbb{R}^n$ be a
set-valued map with nonempty values, and let $f:\mathbb{R}^n\to\mathbb{R}$. Let
$z:\mathbb{R}_+\to\mathcal{X}$ be absolutely continuous and suppose that for a.e.\ $t\ge0$
\[
  \dot z(t)\in G(z(t))
  \qquad\text{and}\qquad
  \tfrac{d}{dt}f(z(t))=\langle v,\dot z(t)\rangle
  \quad\text{for all } v\in G(z(t)).
\]
Then for a.e.\ $t\ge0$,
\[
  \|\dot z(t)\|=\dist\bigl(0,G(z(t))\bigr)
  \qquad\text{and}\qquad
  \tfrac{d}{dt}f(z(t))=\|\dot z(t)\|^{2}=\dist^{2}\bigl(0,G(z(t))\bigr).
\]
\end{lemma}
\begin{proof}
Fix $t\ge0$ at which $\dot z(t)$ exists and both hypotheses hold, and write
$S:=G(z(t))$, $y:=\dot z(t)$, $W:=\operatorname{span}(S-S)$.

Taking $v=y\in S$ in the chain rule gives $\tfrac{d}{dt}f(z(t))=\langle y,y\rangle=\|y\|^{2}$.

Since the chain rule assigns the same value $\langle v,y\rangle$ to every $v\in S$, we get
$\langle s-s',y\rangle=0$ for all $s,s'\in S$, hence $\langle w,y\rangle=0$ for every
$w\in W$; that is, $y\in W^{\perp}$.

Consequently, for every $w\in W$,
\[
  \|y+w\|^{2}=\|y\|^{2}+2\langle y,w\rangle+\|w\|^{2}=\|y\|^{2}+\|w\|^{2}\;\ge\;\|y\|^{2},
\]
and since $y\in y+W$ we conclude $\|y\|=\dist(0,y+W)$.

Moreover $y\in S$, so $s-y\in S-y\subseteq W$ for every $s\in S$, i.e.\ $S\subseteq y+W$.
For a distance function, it holds that
\[
  \dist(0,S)\;\ge\;\dist(0,y+W)\;=\;\|y\|=\|\dot z(t)\| .
\]
The reverse inequality $\dist(0,G(z(t)))\le\|\dot z(t)\|$ is immediate from $\dot z(t)\in G(z(t))$. Combining the two gives the claim.
\end{proof}

For a locally Lipschitz potential $\Phi$, Rademacher's theorem gives almost-everywhere differentiability; conservativity further ensures that $D(x)=\{\nabla\Phi(x)\}$ Lebesgue almost everywhere \citep[Theorem 1]{bolte2021conservative}. Moreover, the Clarke subgradient is contained in the convex hull of any conservative field for $\Phi$ \citep[Corollary 1]{bolte2021conservative}: 
\begin{equation}\label{cor:clarke_is_a_minimial_conserv_field}
\partial \Phi(x) \subseteq \operatorname{conv}( \mathcal{D}(x) ).
\end{equation}

And as it turns out, any function that is locally Lipschitz and definable in some o-minimal structure, is also path differentiable.

\begin{lemma}\label{lem:conservative_field_clarke}
    Let $\Phi: \mathbb{R}^n \to \mathbb{R}$ be a locally Lipschitz function and definable in some o-minimal structure, then all three hold: 
    \begin{enumerate}
        \item $\Phi$ admits a chain rule for Clarke subdifferential,
        \item $\partial \Phi$ is a conservative field,
        \item $\Phi$ path differentiable.
    \end{enumerate} 
\end{lemma}
\begin{proof}
    By \cite{van1996geometric}, a definable function admits Whitney $C^1$-stratification. And so by [Theorem 5.8.]\cite{DacDruKakLee2018}, we get that a locally Lipschitz function that is Whitney $C^1$-stratifiable, admits a chain rule for Clarke subdifferential. Finally, by [Corollary 2]\cite{bolte2021conservative}, and by the fact that $f$ has a chain rule for a Clarke subdifferential, then Clarke of $\Phi$ is a conservative field and $\Phi$ is path differentiable, which finishes the proof.
\end{proof}
We state some other important properties in the following theorem and corollary.
\begin{theorem}[Theorem 5, \citet{bolte2021conservative}]\label{thm:morse_sard}
    Let $\mathcal{D}: \mathbb{R}^n \rightrightarrows \mathbb{R}^n$ be a conservative field
    for $\Phi: \mathbb{R}^n \to \mathbb{R}$ and assume that both $\Phi$ and $\mathcal{D}$
    are definable in some o-minimal structure $\mathcal{O}$. Then the set of
    $\mathcal{D}$-critical values
    \[
        \bigl\{\, \Phi(x) \;:\; x \in \mathbb{R}^n,\; 0 \in \mathcal{D}(x) \,\bigr\}
    \]
    is finite.
\end{theorem}

\begin{corollary}[Corollary 4, \citet{bolte2021conservative}]\label{cor:outer_sum_rule}
    Let $\Phi_1, \dots, \Phi_N : \mathbb{R}^n \to \mathbb{R}$ be locally Lipschitz
    functions and let $\mathcal{D}_i : \mathbb{R}^n \rightrightarrows \mathbb{R}^n$ be a
    conservative field for $\Phi_i$, $i = 1, \dots, N$. Then
    $\sum_{i=1}^{N} \mathcal{D}_i$ is a conservative field for
    $\sum_{i=1}^{N} \Phi_i$.
\end{corollary}

\subsection{Weak convexity and an auxiliary convergence lemma.}\label{Xu_lemma_subsection}

\begin{definition}[Weak Convexity]\label{def:weakly_convex_function}
A function $f:\mathbb{R}^n \to \mathbb{R}$ is $C$-weakly convex for $C \geq 0$, if the function
$x \mapsto f(x) + \tfrac{C}{2}\|x\|^2$ is convex. It is \emph{weakly convex} if it is $C$-weakly convex for some $C$.
\end{definition}

The following sequence lemma will be used in Lemma~\ref{lem:prox_center_collapses} to show that the proximal center asymptotically tracks the primal iterates.

\begin{lemma}[Lemma 2.1 \cite{xu2004viscosity}]\label{lemma:xu_ek_convergence}
Let $\{a^k\}$ be a sequence of non-negative real numbers such that
\[
a^{k+1} \le (1-\gamma^k)\, a^k + \delta^k,
\]
where $\gamma^k \in (0,1)$, and suppose
\begin{enumerate}
\item[(a)] $\displaystyle\sum_{k=0}^\infty \gamma^k = \infty$, and
\item[(b)] either $\displaystyle\limsup_{k\to\infty} \frac{\delta^k}{\gamma^k} \le 0$
or $\displaystyle\sum_{k} |\delta^k| < \infty$.
\end{enumerate}
Then $a^k \to 0$.
\end{lemma}

\newpage
\section{Theoretical results}\label{app:theoretical}

\subsection{Bias Introduced by Penalty/Barrier Functions}\label{subsection:bias}

Let us recall and prove \cref{theorem:feasible_subset}.
{\toggletrue{restatingtheorem}\theoremFeasibleTransformed*\togglefalse{restatingtheorem}}
\begin{proof}[Proof of \cref{theorem:feasible_subset}]
\label{proof:feasible_subset}
    Let $\hat{x} \in \mathcal{S}^B_{\operatorname{transform}}$ and fix an arbitrary
    $i \in \{ 1, \dots, m \}$. By feasibility,
    $$\mathbb{E}_{\xi_{1:B}} \bigl[ p_i \varphi \bigl( \bar{g}_i(\hat{x}) / p_i \bigr) \bigr] \leq 0.$$
    Since $p_i > 0$, dividing by $p_i$ gives
    $$\mathbb{E}_{\xi_{1:B}} \bigl[ \varphi \bigl( \bar{g}_i(\hat{x}) / p_i \bigr) \bigr] \leq 0,$$
    and Jensen's inequality yields
    $$\varphi \bigl( \mathbb{E}_{\xi_{1:B}}\bigl[ \bar{g}_i(\hat{x}) \bigr] / p_i \bigr)
    \leq \mathbb{E}_{\xi_{1:B}} \bigl[ \varphi \bigl( \bar{g}_i(\hat{x}) / p_i \bigr) \bigr] \leq 0.$$
    By the definition of $\varphi \in \{ \varphi_{\mathrm{QL}}, \varphi_{\mathrm{QR}} \}$,
    we have $\varphi(t) \leq 0$ if and only if $t \leq 0$; hence
    $\mathbb{E}_{\xi_{1:B}}\bigl[ \bar{g}_i(\hat{x}) \bigr] / p_i \leq 0$, and since
    $p_i > 0$, also $\mathbb{E}_{\xi_{1:B}}\bigl[ \bar{g}_i(\hat{x}) \bigr] \leq 0$.
    Finally, $\bar{g}_i(\hat{x})$ is an unbiased estimator of the constraint, i.e.,
    $\mathbb{E}_{\xi_{1:B}}\bigl[ \bar{g}_i(\hat{x}) \bigr]
    = \mathbb{E}_{\xi}\bigl[ g_i(\hat{x}, \xi) \bigr]$, so
    $\mathbb{E}_{\xi}\bigl[ g_i(\hat{x}, \xi) \bigr] = g_i(\hat x) \leq 0$.
    As $i$ was arbitrary, $\hat{x} \in \mathcal{S}$.
\end{proof}
In the next proof, we use the fact that the two particular penalty/barrier functions have the following second derivatives: 
\begin{equation}\label{eq:der_penalty_barrier_functions}
    \begin{aligned}
    \varphi''_{\text{QL}}(t) =
    \begin{cases}
    1 & \text{if } t \ge -\frac{1}{2}, \\ 
    \frac{1}{4t^2}& \text{if } t < -\frac{1}{2},
    \end{cases}
    \quad
    \varphi''_{\text{QR}}(t) =
    \begin{cases}
    1  & \text{if } t \ge -\frac{1}{3}, \\
    \frac{64}{27(1-t)^3 } & \text{if } t < -\frac{1}{3}.
    \end{cases}
    \end{aligned}
    \end{equation}
For the particular
penalty/barrier functions defined in \cref{eq:penalty_barrier_functions}, the bounded curvature of $\varphi$ yields an
explicit bound on the Jensen gap. We recall here \cref{theorem:jensengap}:
\begin{proof}[Proof of \cref{theorem:jensengap}]
    From \cref{eq:der_penalty_barrier_functions}, we have
    $\varphi''(t) \leq 1$ for all $t \in \mathbb{R}$. Hence
    $$ h(t) = \tfrac{1}{2} t^2 - \varphi(t) $$
    satisfies $h''(t) = 1 - \varphi''(t) \geq 0$ and is therefore convex. Applying
    Jensen's inequality to $h$ with a square-integrable random variable $X$ (finite second moment is guaranteed by a finite variance) gives
    $\mathbb{E}[h(X)] \geq h(\mathbb{E}[X])$. Substituting for $h$ gives
    \begin{align}
        \frac{1}{2}\mathbb{E}[X^2] - \mathbb{E} \bigl[ \varphi(X) \bigr] &\geq \frac{1}{2}\bigl(\mathbb{E}[X]\bigr)^2 - \varphi \bigl( \mathbb{E}[X] \bigr),
    \end{align}
    which directly implies
    \begin{align}
         \frac{1}{2} \operatorname{Var}(X) =  \frac{1}{2} \Bigl( \mathbb{E}[X^2] - \bigl(\mathbb{E}[X]\bigr)^2 \Bigr) \geq
        \mathbb{E} \bigl[ \varphi(X) \bigr] - \varphi \bigl( \mathbb{E}[X] \bigr).
    \end{align}
    Setting $X := \bar{g}_i(x) / p_i$, with the expectation taken over
    $\xi_{1:B}$, yields
    \begin{equation}\label{eq:proof_bias_penaltybar}
        \mathbb{E}_{\xi_{1:B}} \bigl[ \varphi\bigl( \bar{g}_i(x) / p_i \bigr) \bigr]
        - \varphi \bigl( \mathbb{E}_{\xi_{1:B}}[ \bar{g}_i(x)] / p_i \bigr)
        \;\leq\; \frac{1}{2 p_i^2} \operatorname{Var}\bigl( \bar{g}_i(x) \bigr).
    \end{equation}
    Multiplying \eqref{eq:proof_bias_penaltybar} by $p_i$ and using
    $\operatorname{Var}(\bar{g}_i(x)) = \frac{1}{B}\operatorname{Var}(g_i(x, \xi))$,
    which holds since $\xi_1, \dots, \xi_B$ are i.i.d., gives the claim with
    $\mathbb{E}_{\xi_{1:B}}[\bar{g}_i(x)]$ in place of
    $\mathbb{E}_{\xi}[g_i(x,\xi)]$; these coincide because $\bar{g}_i(x)$ is an
    unbiased estimator of the constraint, which finishes the proof.
\end{proof}
\cref{theorem:jensengap} quantifies how the bias introduced by the
penalty/barrier transformation depends on the batch size and the penalty parameters.
Moreover, the bound is tight: on the quadratic branch of $\varphi$, where
$\varphi(t) = t + \frac{1}{2}t^2$, inequality \eqref{eq:proof_bias_penaltybar} holds
with equality (this is immediate by substituting for $\varphi$ in \eqref{eq:proof_bias_penaltybar}), so the Jensen gap equals
$\operatorname{Var}(g_i(x,\xi))/(2 B p_i)$ whenever $\bar{g}_i(x)/p_i$ lies in the
quadratic region almost surely. 

\subsection{SPBM Reduces to SSL-ALM-like Primal-Dual Update}\label{sec:equivalence_to_alm}
\cite{ben1997penalty} showed that PBM reduces to the standard augmented
Lagrangian method~\cite{hestenes1969multiplier, powell1969method} for a particular
choice of penalty-updating rule. We show the reduction survives mini-batching.
Assume \emph{(i)} the penalty-updating function \eqref{penalty_updating_formulas_alm},
\emph{(ii)}  no dual damping, $\gamma^k \equiv 0$, \emph{(iii)} $\rho \lambda^k \in [\ell_c, u_c]^m$, and \emph{(iv)} that $\bar g_i(x^k)/p_i^k$ lies in
the quadratic branch of $\varphi$ (i.e. $\bar g_i(x^k) \ge -p_i^k/2$ for $\varphi_{\mathrm{QL}}$ or $\bar g_i(x^k) \ge -p_i^k/3$ for $\varphi_{\mathrm{QR}}$),
where the quadratic branch of $\varphi\in\{\varphi_{\mathrm{QL}},\varphi_{\mathrm{QR}}\}$ satisfies
\[
\varphi(t) = t + \tfrac12 t^2, \qquad \varphi'(t) = 1 + t.
\]
Then the Lagrangian becomes the augmented Lagrangian with penalty $\rho$:
\begin{align}
\overline{\mathcal{L}}_{p^k}(x^k, \lambda^k)
&= \bar f(x^k) + \sum_{i=1}^m \lambda_i^k p_i^k \varphi\!\left( \bar g_i(x^k) / p_i^k \right)
= \bar f(x^k) + \sum_{i=1}^m \lambda_i^k p_i^k \left( \frac{\bar g_i(x^k)}{p_i^k}
   + \frac{\bar g_i^2(x^k)}{2 (p_i^k)^2} \right) \\
&= \bar f(x^k) + \sum_{i=1}^m \left( \lambda_i^k \bar g_i(x^k)
   + \frac{\lambda_i^k \bar g_i^2(x^k)}{2 p_i^k} \right) \\
&= \bar f(x^k) + (\lambda^k)^\top \bar g(x^k) + \frac{1}{2\rho} \left\| \bar g(x^k) \right\|^2.
\end{align}
The dual update gives
\[
\lambda^{k+1} = \lambda^k \odot \varphi'\!\left( \bar g(x^k) \oslash p^k \right)
= \lambda^k + \lambda^k \odot \frac{\bar g(x^k)}{\lambda^k \rho}
= \lambda^k + \frac{1}{\rho}\, \bar g(x^k),
\]
which is the ALM multiplier update. The reduction is sample-path-wise: the algebra is
pointwise in the realized $\bar g(x^k)$, so it is unaffected by mini-batching.
Mini-batching does, however, bias the augmented Lagrangian itself. Since $\bar f$ is
unbiased and $\lambda^\top \bar g$ is linear in $\bar g$, only the quadratic term
contributes:
\[
\mathbb{E}\bigl[ \tfrac{1}{2\rho}\bar g_i^2(x) \bigr]
- \tfrac{1}{2\rho}\bigl( \mathbb{E}[\bar g_i(x)] \bigr)^2
= \tfrac{1}{2\rho}\operatorname{Var}\bigl(\bar g_i(x)\bigr)
= \frac{ \operatorname{Var}\bigl(g_i(x,\xi)\bigr)}{2\rho B},
\]
recovering the $1/B$ decay of \cref{theorem:jensengap} in this special case.
It is worth noting that the recovered update is that of the \emph{equality}-constrained ALM: for
strongly inactive constraints $\bar g_i < -p_i^k/2$ (or $\bar g_i < -p_i^k/3$ for $\varphi_{\mathrm{QR}}$), $\varphi$ leaves its quadratic
branch and SPBM decays $\lambda_i$ multiplicatively compared to ALM,
so the methods differ there. Lastly, since SPBM utilizes the Moreau envelope, setting the penalty-updating function to \eqref{penalty_updating_formulas_alm} yields the SSL-ALM algorithm up to the order of the dual and primal update steps, the prox-center index, and the coupling of the dual learning rate and the penalty regularizer. Note that $\kappa$ becomes a redundant hyperparameter and is instead replaced by $\rho$. 

\subsection{Convergence Analysis}\label{sec:convergence_analysis}
\providecommand{\Lemp}{\mathcal{L}^{\mathrm{emp}}}
\providecommand{\dist}{\operatorname{dist}}
\providecommand{\Lip}{\operatorname{Lip}}
\providecommand{\conv}{\operatorname{conv}}
\providecommand{\bball}{\mathbb{B}}
\providecommand{\supp}{\operatorname{supp}}

The analysis applies to \cref{alg:spbm_schedules}, a variant of the benchmarked
\cref{alg:spbm} that drives the dual and penalty averaging coefficients to a limit along fixed schedules.

We work on a probability space $(\Omega,\mathcal{F}, P)$ carrying the
mini-batch draws $\{\mathbf{j}^k\}_{k\geq0}$, with the filtration
$\mathcal{F}^{k}:=\sigma(\mathbf{j}^0,\dots,\mathbf{j}^{k-1})$. At step $k$ the algorithm
returns the direction $\bar y^{k}$ of line~4, which we express as a sum of its conditional mean and the noise of the problem:
\[
  y^{k}:=\mathbb{E}\bigl[\bar y^{k}\mid\mathcal{F}^{k}\bigr],
  \qquad
  \zeta^{k}:=\bar y^{k}-y^{k},
\]
so that the primal update of line~5 reads
\begin{equation}\label{SGD_dynamics}
  x^{k+1}=x^{k}+\alpha^{k}\bigl(y^{k}+\zeta^{k}\bigr).
\end{equation}
Here $x^k,\lambda^k,p^k,s^k$ and $y^k$ are $\mathcal{F}^k$-measurable, while $\bar y^k$ and
$\zeta^k$ are $\mathcal{F}^{k+1}$-measurable.
Let $\{x^{k},\lambda^{k},p^{k},s^{k}\}_{k\geq0}$ be generated by
\cref{alg:spbm_schedules}. We show that the duals and the penalty parameters converge
almost surely to limits $\hat\lambda$ and $\hat p$, and that the primal iterates
asymptotically track the differential inclusion
\[
  \dot x(t)\in\mathcal{D}_{\hat p}\bigl(x(t),\hat\lambda\bigr),
\]
whose equilibria are the $\mathcal{D}_{\hat p}(\cdot,\hat\lambda)$-critical points appearing
in \cref{main_theorem}. Here $\mathcal{D}_{\hat p}(\cdot,\hat\lambda)$ is a conservative
field of $-\mathcal{L}^{\mathrm{emp}}_{\hat p}(\cdot,\hat\lambda)$, where
$$\mathcal{L}^{\mathrm{emp}}_{p}(\cdot,\lambda)
:=\mathbb{E}\bigl[\bar{\mathcal{L}}_{p}(\cdot,\lambda)\bigr]
$$
is the expected value of the Lagrangian of the transformed problem \eqref{eq:stoctrsto_transformed}. The schedules are required to satisfy $\gamma^{k},\kappa^{k}\to1$ with
$\sum_{k}(1-\gamma^{k})<\infty$ and $\sum_{k}(1-\kappa^{k})<\infty$, and
$\sum_{k}\alpha^{k}=\infty$ with $\sum_{k}(\alpha^{k})^{2}<\infty$.
For instance, we suggest
\begin{align}\label{eq:kappa_update}
\kappa^{k+1} &=
1-
\frac{1}{
\lfloor k/L\rfloor^{1+\varepsilon_{\kappa}}
+\frac{1}{1-\kappa^0}
},
\qquad &\varepsilon_{\kappa} >0,\\
\label{eq:gamma_update}
\gamma^{k+1} &=
1-
\frac{1}{
\lfloor k/L\rfloor^{1+\varepsilon_{\gamma}}
+\frac{1}{1-\gamma^0}
},
\qquad &\varepsilon_{\gamma} >0,\\
\label{eq:alpha_update}
\alpha^{k+1} &= \frac{1}{(1+k)^{\varepsilon_\alpha}},
\qquad &\varepsilon_{\alpha} \in (\frac{1}{2}, 1].\\
\end{align}
Boundedness of the duals and penalties is enforced rather than assumed. 
The duals are projected onto $[0,\Lambda]^{m}$ in line~6, and the penalties implicitly onto
$[\ell_c,u_c]^{m}$ by both suggested updating methods \eqref{penalty_updating_formulas} and \eqref{penalty_updating_formulas_alm}.
\begin{algorithm}[h]
  \caption{Stochastic Penalty-Barrier Method (SPBM) with schedules. Schedules: $\gamma^k,\kappa^k$ follow \eqref{eq:kappa_update},\eqref{eq:gamma_update} ($\to 1$); $\alpha^k$ follows a square-summable, but not summable schedule \eqref{eq:alpha_update}($\to 0$). \PrimalUpdate{} is standard SGD step \eqref{SGD_dynamics}. Since $\kappa^k$ is now scheduled, we redefine \eqref{penalty_updating_formulas} as $\hat{\pi}^{\operatorname{ADAPT}}(\bar{g}, p, \lambda, \kappa^k)$. We assume $\rho$ is fixed, so for consistency we simply redefine \eqref{penalty_updating_formulas_alm} as $\hat \pi^{\operatorname{ALM}}_{\rho}(\bar{g}, p, \lambda, \cdot)$. }
  \label{alg:spbm_schedules}
  \begin{algorithmic}[1]
    \STATE {\bfseries Input:} $x^0\in\mathbb{R}^n$; $\lambda^0\in\mathbb{R}^m_{++}$; $p^0=\mathbf{1} (\text{or } p^0 = \rho \,\lambda^0  \text{ if }\hat\pi = \hat\pi_\rho^{\operatorname{ALM}})$; $s^0=x^0$; schedules $\alpha^0,\gamma^0$ and $\kappa^0 (\text{alternatively }\rho \in \mathbb{R}_{++}, \text{ if }\hat\pi = \hat\pi_\rho^{\operatorname{ALM}})$; $\mu\ge0$, $\delta\in[0,1)$; batch size $B$; 
    $\hat\pi\in\{\hat\pi_\rho^{\operatorname{ALM}},
    \hat\pi^{\operatorname{ADAPT}}\}$; $\Lambda \in \mathbb{R}_{++}$.
    \FOR{$k=0$ {\bfseries to} $K$}
      \STATE Sample $\xi_1,\dots,\xi_B \iid \Xi$
      \STATE $\bar y^{k} = -\overline{\nabla}\mathcal{L}_{p^{k}}(x^k,\lambda^{k}) - \mu(x^k-s^k)$ \hfill\COMMENT{Moreau-augmented grad, \eqref{eq:moreau_env_prob}}
      \STATE $x^{k+1} = x^k + \alpha^k \bar y^k$ \hfill\COMMENT{Primal step, \eqref{eq:scheme1}, \eqref{SGD_dynamics}}
      \STATE $\lambda^{k+1} = \operatorname{Proj}_{[0, \Lambda]^m}\left[\gamma^k\lambda^k + (1-\gamma^k)\,\lambda^k \odot \varphi'(\overline{g}(x^k)\oslash p^k)\right]$ \hfill\COMMENT{Dual variable step, \eqref{eq:scheme2}}
      \STATE $p^{k+1} = \hat\pi(\overline{g}(x^k), p^k, \lambda^{k+1}, \kappa^k)$ \hfill\COMMENT{Penalty param. step, \eqref{penalty_updating_formulas}/\eqref{penalty_updating_formulas_alm}}
      \STATE $s^{k+1} = \delta s^k + (1-\delta)x^{k+1}$ \hfill\COMMENT{Prox-center update}
      \STATE Advance schedules $\alpha^{k},\gamma^{k},\kappa^{k}$ \hfill\COMMENT{Schedule updates}
    \ENDFOR
  \end{algorithmic}
\end{algorithm}


The section is organised as follows. \cref{assumptions_convergence_analysis} states the two standing
assumptions and their direct implications. \cref{main_theorem_section} then proves the convergence
\cref{main_theorem}, pointing to lemmas that are stated and proved afterwards. Lastly, 
\cref{dual_conv_subsection} establishes convergence of the duals and penalties together with
condition on limiting properties of $y^k$, and \cref{sec:lyapunov} the Lyapunov properties behind the descent and the Weak-Sard property. 


\subsubsection{Assumptions and Direct Implications}\label{assumptions_convergence_analysis}

Let $(\Omega,\mathcal{F},P)$ be a probability space carrying the independent
mini-batch draws $\{\mathbf{j}^k\}_{k\ge0}$, such that $\mathbf{j}^k \in \{ 1, \dots, M\}^B$ where $M \in \mathbb{N}$ is the number of samples in our dataset. And let $\{\mathcal{F}^k\}_{k\ge0}$
be the filtration generated by the batches drawn before iteration $k$:
\[
  \mathcal{F}^k:=\sigma\bigl(\mathbf{j}^0, \dots, \mathbf{j}^{k-1}\bigr),
  \qquad \mathcal{F}^0:=\{\emptyset,\Omega\},
  \qquad \mathcal{F}_\infty:=\sigma\Bigl(\bigcup_{k\ge0}\mathcal{F}^k\Bigr),
\]
so that $\mathcal{F}^k\subseteq\mathcal{F}^{k+1}\subseteq\mathcal{F}$. This gives that $x^k, \lambda^k, p^k, s^k$ and $y^k$ are $\mathcal{F}^k$-measurable but $\bar y^k$ need not be, and the draw $\mathbf{j}^k$ is independent of $\mathcal{F}^k$.

Specifically, assume a finite dataset $(\xi^{(1)},\dots, \xi^{(M)})$, possibly with repetitions, and $\Xi$ its empirical measure; i.e. $\Xi$ is the law of $\xi^{(l)}$ for $l \sim \operatorname{Unif}\{1,\dots, M\}$, in particular $\operatorname{supp}\Xi$ is finite. 
So drawing $\xi\sim\Xi$ is the same as drawing an index uniformly from $\{1,\dots,M\}$, and a mini-batch of $B$ i.i.d. draws is encoded by a multi-index
\[
  \mathbf{j}=(j_1,\dots,j_B)\in\{1,\dots,M\}^B.
\]
We accordingly write $\bar g_i(x;\mathbf{j}):=\frac1B\sum_{b=1}^{B}g_i(x,\xi^{(j_b)})$, and likewise $\bar f(\cdot\,;\mathbf{j})$. This connects the notation of \cref{subsection:bias} with the one used below. Concretely, the run may then be realised on the probability space with $\Omega =(\{1,\dots,M\}^B)^{\mathbb{N}_0}$, equipped with its product sigma algebra and the product of uniform measures $P$. In particular, $P(\mathbf{j}^k=\mathbf{j})=M^{-B}$. 

\begin{assumption}[Assumption A.2 in \cite{DacDruKakLee2018}]\label{asm:C}
    There exists a deterministic constant $R$ such that, almost surely,
    \[
    \sup_{k \geq 0} \|x^k\| \leq R.
    \]
\end{assumption}
\begin{remark}
\cref{asm:C} is a standard boundedness requirement in stochastic approximation. We impose no restriction on the size of $R$, only that it is finite. Conditions, under which such assumption can be guaranteed, are discussed in \citep[Section~6.1]{DacDruKakLee2018}.

Unlike the duals and penalties, we do not project the primal iterates. Doing so would change the main statement being proved. By projecting onto
$R\mathbb{B}$, the limiting dynamics would change into a projected inclusion, changing the $D_{\hat p}(\cdot,\hat\lambda)$-criticality being established. 
\end{remark}

\begin{definition}[Bounded primal set]\label{def:working_set}
Let $R$ be the constant from \cref{asm:C} and set
\[
  \mathcal{X}:=\bigl\{x\in\mathbb{R}^{n}:\|x\|\le R+1\bigr\},
\]
a compact convex set. 
\end{definition}
\begin{corollary}[Condition A.1 in \cite{DacDruKakLee2018}]\label{remark:limit_points_are_contained_in_X}
By \cref{asm:C}, $x^k \in R\mathbb{B} \subset \operatorname{int}\mathcal{X}$ for every $k$, almost surely, and hence every limit
point of $\{x^k\}$ lies in $R\mathbb{B} \subset \mathcal{X}$.
\end{corollary}

\begin{remark}\label{remark:assumptionA}
The padding by $1$ is not cosmetic: \cref{proposition_definability_D_uppersem}, \cref{lem:jointlip} and \cref{lem:consistency} bound the Clarke subdifferential at $x$. Since a Clarke subdifferential is determined by the nearby points, this requires a whole neighborhood of each $x$ to lie in $\mathcal{X}$.  Any fixed $\varepsilon > 0$ in place of $1$ would work too.
\end{remark}



\begin{lemma}[Schedules and projection of \cref{alg:spbm_schedules} satisfy A.3]\label{asm:C_instead_as_proposition} \quad 
\begin{enumerate}
    \item The dual variables and penalty variables are within compact boxes almost surely
\[
 \lambda^k\in[0,\Lambda]^m, \quad p^k\in[\ell_c,u_c]^m.
\]
\item (A.3 in \cite{DacDruKakLee2018}) The sequence of primal learning rates $\{\alpha^k\}$  is nonnegative, square summable, but not summable:
    \[
        \alpha^k \geq 0, \qquad \sum_{k=1}^{\infty} \alpha^k = \infty, \qquad \text{and} \qquad \sum_{k=1}^{\infty} (\alpha^k)^2 < \infty.
    \]
\item  The proposed schedule of the dual variable \eqref{eq:gamma_update}, and of the penalty parameter \eqref{eq:kappa_update} have the property
\[
\sum_{k=0}^{\infty}(1-\kappa^k)<\infty ,\qquad \sum_{k=0}^{\infty}(1-\gamma^k)<\infty.
\]
Specifically, in the limit $\gamma^k, \kappa^k \to 1$.
\end{enumerate}
\end{lemma}
\begin{proof}
    \emph{(1)} The duals and penalties are ensured to be 
within compact sets $\lambda^k\in[0,\Lambda]^m$, $p^k\in[\ell_c,u_c]^m$ through projections onto bounded boxes (see \cref{alg:spbm_schedules}). Therefore by \cref{asm:C}, it follows that all three random variables $x^k, p^k, \lambda^k$ lie in a compact set almost surely. \emph{(2)}  Follows from the suggested schedule \eqref{eq:alpha_update} by $p$-series test. \emph{(3)} It holds $1-\gamma^{k+1} = O\left(k^{-1-\varepsilon_{\gamma}}\right)$
and consequently, by $p$-series test, the sequence is summable $
\sum_{k=0}^{\infty}(1-\gamma^k)<\infty. $
The same applies to summability of $\sum_{k=0}^{\infty}(1-\kappa^k)$.
Note that when $k \leq L$, then $(\gamma^k, \kappa^k) = (\gamma^0, \kappa^0)$.
\end{proof}

\newpage
\begin{lemma}[Regularity of penalty-barriers]\label{prop:penalty_reg}
The function $\varphi\in\{\varphi_{\mathrm{QL}},\varphi_{\mathrm{QR}}\}$ is convex, \emph{finite-valued on all of $\mathbb{R}$}, and definable in the o-minimal structure~$\mathbb{R}_{\operatorname{exp}}$. Moreover, $\varphi \in C^2$ and $\varphi' > 0$.
\end{lemma}
\begin{proof}
    By construction $\varphi\in\{\varphi_{\mathrm{QL}},\varphi_{\mathrm{QR}}\}$ is convex and $C^2$, locally Lipschitz and definable. Specifically, using the notation taken from ~\citet{bareilles2025deep}, $\varphi_{\mathrm{QR}}$  is semi-algebraic, hence definable in $\mathbb{R}_{\operatorname{alg}}\subseteq\mathbb{R}_{\exp}$; $\varphi_{\mathrm{QL}}$ involves the logarithm on its second branch and is therefore definable in $\mathbb{R}_{\exp}$. $\varphi' > 0$ follows from \eqref{eq:penalty_barrier_functions}. 
\end{proof}

\begin{assumption}[Definability, finite support, and the subgradient oracle]\label{asm:reg} \quad
\begin{enumerate}
\item For every
$\mathbf{j}\in\{1,\dots,M\}^{B}$ the mini-batch estimators $\bar f(\cdot;\mathbf{j})$ and
$\bar g_i(\cdot;\mathbf{j})$ are locally Lipschitz and definable in the common $\mathcal{O} \supseteq \mathbb{R}_{\operatorname{exp}}$. We fix $\mathcal{O}$ from this point on.
\item 
dataset $(\xi^{(1)},\dots,\xi^{(M)})$ is finite, and $\Xi$ is the law of $\xi^{(l)}$ for $j$ being uniformly distributed over $M$ samples, which gives
        \begin{equation}\label{eq:Lemp}
          \mathcal{L}^{\mathrm{emp}}_{p}(x,\lambda)
:=\mathbb{E}_{\xi_{1:B}}\big[\bar{\mathcal{L}}_{p}(x,\lambda)\big]
=\frac{1}{M^{B}}\sum_{\mathbf{j}\in\{1, \dots, M \}^{B}}\bar{\mathcal{L}}_{p}(x,\lambda;\mathbf{j}).
        \end{equation}
In other words $\mathcal{L}^{\mathrm{emp}}_{p}(\cdot,\lambda)$ is the average of the mini-batch over all possible $M^B$ batches, giving a deterministic value corresponding to the true expectation value of $\overline{\mathcal{L}}_p(\cdot, \lambda)$. 
\item (Subgradient Backpropagation Oracle) For each $\mathbf{j}\in\{1,\dots,M\}^{B}$ there is a Borel map
$\mathrm{ch}(\cdot,\cdot,\cdot\,;\mathbf{j}):\mathbb{R}^{n}\times[0,\Lambda]^{m}\times[\ell_{c},u_{c}]^{m}\to\mathbb{R}^{n}$
with $\mathrm{ch}(x,\lambda,p;\mathbf{j})\in-\partial_x\bar{\mathcal{L}}_{p}(x,\lambda;\mathbf{j})$,
and line~4 of \cref{alg:spbm} (and \cref{alg:spbm_schedules}) returns
$\bar y^{k}=\mathrm{ch}(x^{k},\lambda^{k},p^{k};\mathbf{j}^{k})-\mu(x^{k}-s^{k})$.
\end{enumerate}
\end{assumption}

\begin{remark}
Regarding (1), we argue in \cref{common_functions_subsubsection} that o-minimal setting is natural in the context of deep learning. Moreover, note that from \cref{prop:penalty_reg} it follows that $\varphi$ is definable in $\mathcal{O}$. For the rest of the proof, definable means definable in $\mathcal{O}$.
  Definability is later required for definability of the Lagrangian function, which propagates into the definability of the conservative field of the Lyapunov function; without this the Morse-Sard theorem could not be applied. Regarding~(3), since we treat SPBM as a subgradient method, an oracle that gives the access to an element of the Clarke subdifferential is reasonable. Although many modern ML libraries implementing automatic differentiation, e.g. PyTorch and TensorFlow, do not necessarily output an element of a Clarke subdifferential, there exist implementations that do, such as \citet{kakade2018provably}.
\end{remark}

Later, we analyze a Lyapunov function based on $\mathcal{L}^{\mathrm{emp}}_p(x, \lambda)$ with fixed $p, \lambda$. From \cref{asm:reg}, we can deduce properties of this function:

\begin{lemma}[Definability and regularity of the Lagrangian Estimate]\label{lem:reg}
Under \cref{asm:C},\cref{asm:reg}:
\begin{enumerate}
  \item for every fixed $\lambda\in[0,\Lambda]^m$ and $p\in[\ell_c,u_c]^m$, the map
  \[
    \mathcal{L}^{\mathrm{emp}}_{p}(x,\lambda)
    =\mathbb{E}_{\xi_{1:B}}\big[\bar{\mathcal{L}}_{p}(x,\lambda)\big]
    =\frac{1}{M^{B}}\sum_{\mathbf{j}\in\{1,\dots,M\}^{B}}
      \overline{\mathcal{L}}_{p}(x,\lambda;\mathbf{j})
  \]
  is locally Lipschitz, bounded from below on the compact set $\mathcal{X}$, and definable
  in $\mathcal{O}$; the same holds for each
  $\overline{\mathcal{L}}_{p}(\cdot,\lambda;\mathbf{j})$;

  \item the joint maps
  $(x,\lambda,p)\mapsto\overline{\mathcal{L}}_{p}(x,\lambda;\mathbf{j})$ and
  $(x,\lambda,p)\mapsto\mathcal{L}^{\mathrm{emp}}_{p}(x,\lambda)$
  are locally Lipschitz on
  $\mathbb{R}^{n}\times\mathbb{R}^{m}\times(0,\infty)^{m}$.
\end{enumerate}
\end{lemma}
\begin{proof} \emph{Step 1 (Definable and locally Lipschitz on $\mathcal{X}$).}
Fix $\mathbf{j}$ and $i$. By \cref{asm:reg}(1) and $p_i>0$, the map $ x \mapsto \bar{g}_i(x; \mathbf{j}) / p_i$ is definable and locally Lipschitz. Since $\varphi$ is definable and is $C^2$ (hence
locally Lipschitz; \cref{prop:penalty_reg}),
\[
\text{the composition} \quad
x\mapsto\varphi(\bar{g}_i(x; \mathbf{j})/p_i)
\]
is definable (compositions of
definable maps within the same structure are definable; see \citet{van1998tame}) and locally Lipschitz (compositions of locally Lipschitz
maps are locally Lipschitz). The same holds for
$\bar f(\cdot; \mathbf{j})$. Finite sums over nonnegative combinations (over $i$)
preserve both properties (o-minimal structures are closed under $+$; a local
Lipschitz constant of the sum is bounded by the sum of the constants). Hence
$\mathcal{L}^{\mathrm{emp}}_p(\cdot,\lambda)$ is locally Lipschitz, and definable. Since $\mathcal{X}$ is compact, and $\mathcal{L}^{\mathrm{emp}}_p(x,\lambda)$ is locally Lipschitz, by the extreme value theorem, we get that $\mathcal{L}^{\mathrm{emp}}_p(x,\lambda)$ is bounded from below on $\mathcal{X}$.
Finally, a locally Lipschitz function definable in an o-minimal structure admits a
Whitney $(b)$-stratification compatible with the projection, hence admits a chain rule \citep{DacDruKakLee2018}. Note that all of the above also applies to $\overline{\mathcal{L}}_{p}(x,\lambda;\mathbf{j})$.

\emph{Step 2 (Locally Lipschitz joint maps).}
Fix $\mathbf{j}$ and $i$. The argument of \emph{Step 1} applies once the three maps involving $\lambda$ and $p$ are checked to be locally Lipschitz jointly rather than in $x$ alone. First, $(x,p)\mapsto \bar g_i(x;\mathbf{j})/p_i$ is locally Lipschitz on $\mathbb{R}^n\times(0,\infty)^m$, being a composition of locally Lipschitz maps whose denominator is locally bounded away from zero. Second, $\varphi$ is $C^2$ by
\cref{prop:penalty_reg}, hence locally Lipschitz, so composing preserves the property. Third, $\lambda_i p_i$ is no longer a constant, but a product of locally Lipschitz maps of $(x,\lambda,p)$ is locally Lipschitz. The remaining steps: 
local Lipschitzness of $\bar f(\cdot;\mathbf{j})$, closure of the property under finite sums over $i$, and averaging over $\mathbf{j}$ remain the same as in \emph{Step 1}, which finishes the proof.
\end{proof}

\begin{lemma}[Membership and boundedness of $y^k$]\label{proposition_definability_D_uppersem}
The direction $\bar y^{k}$ generated in
    line~4 of \cref{alg:spbm_schedules} satisfies
    \begin{equation}\label{eq:oracle}
      y^{k}\;:=\;\mathbb{E}\bigl[\bar y^{k}\mid\mathcal{F}^k\bigr]
      \;\in\; \mathcal{D}_{p^k}(x^k, \lambda^k)
             -\mu\,(x^{k}-s^{k}),\quad \text{almost surely},
    \end{equation}
    where $\mathcal{D}_{p^k}(\cdot, \lambda^k): \mathbb{R}^n \rightrightarrows \mathbb{R}^n$ is convex- and compact-valued, upper-semicontinuous, definable conservative field of a potential function $-\Lemp_{p^{k}}(\cdot,\lambda^{k})$. Moreover, $\sup_{k\geq 0}\max\{||y^k||, ||\bar y^k||\}$ is bounded by a deterministic constant $C_y < \infty$ almost surely.
\end{lemma}
\begin{proof} \emph{Step 1 (Membership and properties of the conservative field).}
Let  $\mathrm{ch}(x^k,\lambda^k,p^k;\mathbf{j})$ be the subgradient backpropagation oracle from \cref{asm:reg}(3). Then
\[\bar y^{k}=\mathrm{ch}(x^{k},\lambda^{k},p^{k};\mathbf{j}^{k})-\mu(x^{k}-s^{k}).
\]
Taking the conditional expected value, we get by using measurability of $x^k, s^k$
\[
    \mathbb{E}\bigl[\bar y^{k}\mid\mathcal{F}^k\bigr] =  
    \mathbb{E}[ \operatorname{ch}(x^k, \lambda^k, p^k;\mathbf{j}^{k} ) \mid \mathcal{F}^k ] - \mu(x^k - s^k).
\]
By the finiteness of $\{ 1, \dots, M\}^B$ (\cref{asm:reg}(2)), we rewrite the random variable of choice as
\[
\operatorname{ch}(x^k, \lambda^k, p^k;\mathbf{j}^{k}) = \sum_{\mathbf{j} \in \{ 1, \dots, M\}^B} \operatorname{ch}(x^k, \lambda^k, p^k;\mathbf{j}) \mathbf{1}_{\{ \mathbf{j} = \mathbf{j}^k \} }.
\]
From \cref{asm:reg}(3), we have that \emph{(i)} $\operatorname{ch}(x^k, \lambda^k, p^k; \mathbf{j})$ is $\mathcal{F}^k$-measurable, being the Borel map $\operatorname{ch}(\cdot, \cdot, \cdot; \mathbf{j})$ composed with the $\mathcal{F}^k$-measurable vector $(x^k, \lambda^k, p^k)$, \emph{(ii)} the definition of an indicator random variable $\mathbf{1}_{\{ \mathbf{j} = \mathbf{j}^k \}}$ gives that $\mathbf{1}_{\{ \mathbf{j} = \mathbf{j}^k \}} \in L^1(\Omega, \mathcal{F}, P)$, \emph{(iii)} since $\partial \overline{\mathcal{L}}_{p^k}(x^k, \lambda^k; \mathbf{j})$ is locally bounded (Clarke subdifferential of a locally Lipschitz is locally bounded), $x^k, \lambda^k, p^k$ lie in a compact set almost surely, and the indicator function is bounded by 1, then $\operatorname{ch}(x^k, \lambda^k, p^k;\mathbf{j}) \mathbf{1}_{\{ \mathbf{j} = \mathbf{j}^k \}} \in L^1(\Omega, \mathcal{F}, P)$. Combining \emph{(i)}-\emph{(iii)}, we apply \citep[Proposition 4.2.10]{dembo2019lecture} to pull-out $\operatorname{ch}(x^k, \lambda^k, p^k;\mathbf{j})$, and then in the second step we use the independence of the draw at iterate $k$ of previous draws, to get
\begin{align}
    \mathbb{E}\bigl[\bar y^{k}\mid\mathcal{F}^k\bigr] = 
    &\sum_{\mathbf{j} \in \{ 1, \dots, M\}^B} \operatorname{ch}(x^k, \lambda^k, p^k;\mathbf{j}) \mathbb{E}[ \mathbf{1}_{\{ \mathbf{j} = \mathbf{j}^k \} } \mid \mathcal{F}^k] - \mu(x^k - s^k) \\
    & = \frac{1}{M^B} \sum_{\mathbf{j} \in \{ 1, \dots, M\}^B} \operatorname{ch}(x^k, \lambda^k, p^k;\mathbf{j}) - \mu(x^k - s^k) .
\end{align}
By the definition of the Minkowski sum and by \cref{asm:reg}(3), we have that 
\[
y^{k}\;:=\;\mathbb{E}\bigl[\bar y^{k}\mid\mathcal{F}^k\bigr]
      \;\in\; \frac{1}{M^B} \sum_{\mathbf{j} \in \{ 1, \dots, M\}^B} -\partial_x \overline{\mathcal{L}}_{p^k}(x^k, \lambda^k; \mathbf{j})
             -\mu\,(x^{k}-s^{k}),\quad \text{almost surely}.
\]
From \cref{lem:reg}(1), we have that $-\overline{\mathcal{L}}_{p^k}(\cdot, \lambda^k; \mathbf{j})$ is a locally Lipschitz, definable function, thus $-\frac{1}{M^B}\overline{\mathcal{L}}_{p^k}(\cdot, \lambda^k; \mathbf{j})$ is a definable, locally Lipschitz function. Therefore its Clarke subdifferential is a definable set-valued mapping \citep{bolte2007clarke}. And by Clarke calculus \citep[Theorem 13.10]{clason2026introduction}, it holds
\[
y^{k}\;:=\;\mathbb{E}\bigl[\bar y^{k}\mid\mathcal{F}^k\bigr]
      \;\in\;  \sum_{\mathbf{j} \in \{ 1, \dots, M\}^B} \partial_x \left(-\frac{1}{M^B}\overline{\mathcal{L}}_{p^k}(x^k, \lambda^k; \mathbf{j}) \right)
             -\mu\,(x^{k}-s^{k}),\quad \text{almost surely},
\]
where each element of the sum is a Clarke definable set-valued subdifferential. Moreover, by \cref{lem:conservative_field_clarke}, we have that a Clarke subdifferential of a locally Lipschitz, definable function is also a conservative field. Furthermore, by \cref{cor:outer_sum_rule} \citep[Corollary 4]{bolte2021conservative}, we get that
    \[
    \mathcal{D}_{p^k}(\cdot, \lambda^k) =  \sum_{\mathbf{j} \in \{ 1, \dots, M\}^B} \partial_x \left( -\frac{1}{M^B}\overline{\mathcal{L}}_{p^k}(\cdot, \lambda^k; \mathbf{j}) \right) = -\frac{1}{M^B} \sum_{\mathbf{j} \in \{ 1, \dots, M\}^B} \partial_x \overline{\mathcal{L}}_{p^k}(\cdot, \lambda^k; \mathbf{j}) 
    \]
is a conservative field of a potential function $-\Lemp_{p^k}(\cdot, \lambda^k)$, hence outer-semicontinuous and locally bounded, hence upper-semicontinuous. Moreover, the sum of sets is a Minkowski sum, which preserves convexity, and so $\mathcal{D}_{p^k}(\cdot, \lambda^k)$ is convex-valued. Moreover, Minkowski sum also preserves definability \citep[Ch.~1]{van1998tame} . Lastly $\mathcal{D}_{p^k}(\cdot, \lambda^k)$ is compact as it is a property of a conservative field, which completes the proof.

\emph{Step 2 (Boundedness of $y^k$).} 
We first bound the proximal term. Since $s^0=x^0$ and
$s^{k+1}=\delta s^k+(1-\delta)x^{k+1}$ with $\delta\in[0,1)$, an induction shows
that each $s^k$ is a convex combination of $x^0,\dots,x^k$. By \cref{asm:C} these
lie in the convex set $R\bball$, hence $\|s^k\|\le R$ almost surely and
\begin{equation}\label{eq:bounded_prox}
  \|\mu(x^k-s^k)\| \;\le\; 2\mu R \;<\;\infty .
\end{equation}

Moreover, $\mathcal{D}_{p^k}(x^k,\lambda^k)$ is a Minkowski sum of Clarke subdifferentials of the maps $\overline{\mathcal{L}}_{p^k}(\cdot,\lambda^k;\mathbf{j})$ evaluated at $x^k$. By \cref{lem:reg}(2) these are locally Lipschitz on $\mathbb{R}^n\times\mathbb{R}^m\times(0,\infty)^m$, which contains the compact set $\mathcal{X}\times[0,\Lambda]^m\times[\ell_c,u_c]^m$.  Hence the constant
\[
S := \max_{\mathbf{j}\in\{1,\dots,M\}^B}\ \sup_{\lambda\in[0,\Lambda]^m,\,p\in[\ell_c,u_c]^m}
      \Lip_{\mathcal{X}}\bigl(\overline{\mathcal{L}}_{p}(\cdot,\lambda;\mathbf{j})\bigr)
\]
is finite. By \cref{remark:limit_points_are_contained_in_X}, almost surely
$x^k\in\operatorname{int}\mathcal{X}$, and by \cref{asm:C_instead_as_proposition}(1) $p^k, \lambda^k$ are within compact sets $[0,\Lambda]^m, [\ell_c,u_c]^m$ almost surely. So each
$\overline{\mathcal{L}}_{p^k}(\cdot,\lambda^k;\mathbf{j})$ is Lipschitz near $x^k$ with
constant $S$, and by \citep[Lemma 13.2]{clason2026introduction} every element of $\mathcal{D}_{p^k}(x^k,\lambda^k)$ has a norm at most $S$. Together with \eqref{eq:bounded_prox}, it holds $\|y^k\|\le C_y<\infty$ almost surely for the
deterministic $C_y=S+2\mu R$. Note that by using the same arguments the same applies to the oracle value $\mathrm{ch}(x^k,\lambda^k,p^k;\mathbf{j}^k)$ for any $k$, hence by \cref{asm:reg}(3), also $\|\bar y^k\|\le C_y$ which finishes the proof.
\end{proof}
\begin{corollary}[Dependence of a conservative field $\mathcal{D}$ on $\lambda, p$]\label{remark_conservative_field_D}
Note that we did not assume anything specific about $\lambda^k, p^k$, hence the conservative field $\mathcal{D}_{p}(\cdot, \lambda)$ has the desired properties for any $\lambda, p$. Also, note that none of the properties rely on the negative sign of each element, hence
\[
    -\mathcal{D}_{p^k}(\cdot, \lambda^k) = \frac{1}{M^B}\sum_{\mathbf{j} \in \{ 1,\dots, M \}^B} \partial \overline{\mathcal{L}}_{p^k}(\cdot, \lambda^k; \mathbf{j})
\]
is a convex-valued, upper-semicontinuous and definable conservative field of the potential function $\Lemp_{p^{k}}(\cdot,\lambda^{k})$.
\end{corollary}
\begin{proof}
    Follows immediately from  the proof of \cref{proposition_definability_D_uppersem} showing no assumptions on the fixed $p, \lambda$, and from the sign properties of a Clarke subdifferential.
\end{proof}

\begin{lemma}[Bounded conditional second moment of $\bar y^k$ ]\label{lem:bounded_variance_of_yk}
    Let \cref{asm:C}, \cref{asm:reg} hold. Then the subgradient element approximation has a bounded variance
        \begin{equation}
        \mathbb{E}\left[||\bar y^k - y^k||^2 \mid \mathcal{F}^k \right] \leq \sigma^2_{\bar y} , \quad a.s.
        \end{equation}
\end{lemma}
\begin{proof}
    By using the triangle inequality,  $\mathcal{F}^k$-measurability of $y^k$, and by the boundedness property given by \cref{proposition_definability_D_uppersem}, it holds
    \begin{align}
            \mathbb{E}\left[||\bar y^k - y^k||^2 \mid \mathcal{F}^k \right] 
            &\leq \mathbb{E}\left[ || \bar y^k ||^2 \mid \mathcal{F}^k \right] + 2 \mathbb{E}\left[ || \bar y^k || \mid \mathcal{F}^k \right]\,||y^k||+ ||y^k||^2\\
            &\leq 4C_y^2 := \sigma^2_y < \infty.
    \end{align}
\end{proof}

We express the noise $\zeta^{k}:=\bar y^{k}-y^{k}$ of the recursion \eqref{SGD_dynamics}. We show that the noise sequence is a martingale sequence difference.
\begin{lemma}[$\zeta^k$ is a martingale difference sequence]\label{remark:martingale_sequence_difference}
        Let \cref{asm:C}, \cref{asm:reg} hold. Then $\{ \zeta^k \}$ is a martingale difference sequence w.r.t. to the increasing $\sigma$-fields $\mathcal{F}^k$. Specifically, there exists a function $h : \mathbb{R}^n \to [0, \infty)$ bounded on bounded sets, so that almost surely, for all $k \in \mathbb{N}$, we have
        $$\mathbb{E}[\zeta^{k}\mid\mathcal{F}^k]=0 \quad \text{and} \quad \mathbb{E}[ \| \zeta^k \|^2 \mid \mathcal{F}^k ] \leq h(x^k) \quad a.s.$$
\end{lemma}
\begin{proof}
    The first condition follows from $\mathcal{F}^k$-measurability of $y^k$:
    \[
    \mathbb{E}[ \zeta^k \mid \mathcal{F}^k ] = \mathbb{E}[ \bar y^k \mid \mathcal{F}^k] - y^k = 0 \quad a.s.
    \]
    By definition of $\zeta^k$ and \cref{lem:bounded_variance_of_yk}
    \[
    \mathbb{E}[\| \zeta^k \|^2 \mid \mathcal{F}^k] = \mathbb{E}[ \| \bar y^k - y^k  \|^2 \mid \mathcal{F}^k ] \leq \sigma^2_{\bar y},
    \]
    hence by taking $h=\sigma^2_y$, the second condition holds almost surely.
\end{proof}

\begin{lemma}[Condition A.4 in \cite{DacDruKakLee2018}]\label{cor:A4}
    Let \cref{asm:C}, \cref{asm:reg} hold. There exists a finite random vector $q$ such that the following holds almost surely
    \begin{equation}\label{eq:A4}
    \lim_{j \to \infty} \sum_{k=1}^{j} \alpha^k \zeta^k = q.
    \end{equation}
\end{lemma}
\begin{proof}
    Since $\{ \alpha^k \}$ is square summable (\cref{asm:C_instead_as_proposition}(2)), and from \cref{lem:bounded_variance_of_yk}
    it follows
    \[
    \sum_{i=0}^\infty (\alpha^i)^2 \mathbb{E}[ \| \zeta^i \|^2 \mid \mathcal{F}^i ] \leq  \sigma_{\bar y}^2 \sum_{i=0}^\infty (\alpha^i)^2 < \infty.
    \]
     Hence
    $X_k=\sum_{i=0}^{k}\alpha^i\zeta^i$ satisfies
    $\mathbb{E}\|X_k\|^2<\infty$ for every $k$. Since
    $\mathbb{E}[\zeta^k\mid\mathcal{F}^k]=0$ and $\alpha^k$ is deterministic,
    $\{X_k\}$ is a square-integrable, $\{ \mathcal{F}^{k+1} \}$-measurable martingale.
    Therefore the martingale $X_k$ is in $L^2$.
    Its predictable compensator is
    \[
    \langle X \rangle_j = \sum_{k=1}^j \mathbb{E} \left[ \| \sum_{i=0}^k \alpha^i\zeta^i - \sum_{i=0}^{k-1} \alpha^i\zeta^i \|^2 \mid \mathcal{F}^k \right] = \sum_{k=1}^j \mathbb{E} \left[ \| \alpha^k\zeta^k\|^2 \mid \mathcal{F}^k \right] \leq \sigma_{\bar y}^2 \sum_{k=1}^j (\alpha^k)^2.
    \]
    And so $\langle X \rangle_{\infty}$ is finite almost surely. By \citep[Theorem 5.3.33]{dembo2019lecture} this gives that $X_k$ converges to a finite limit almost surely, which completes the proof. The arguments of this lemma are mostly taken from \citep[Lemma 4.1.]{DacDruKakLee2018} and extended to our setting.
\end{proof}

\subsubsection{Proof of the Main Theorem}\label{main_theorem_section}
\begin{theorem}[Primal criticality and transformed Lagrangian convergence]\label{main_theorem}
Let \cref{asm:C} and \cref{asm:reg} hold. Let $\{ x^k, \lambda^k, p^k, s^k \}_{k \geq 0}$ be generated by \cref{alg:spbm_schedules}. Then almost surely there exist random $\hat \lambda \in  [0, \Lambda]^m$, $\hat p \in [\ell_c, u_c]^m$ such that
\begin{enumerate}
    \item $\lambda^k\to\hat\lambda$, $p^k\to\hat p$;
    \item $\|x^k-s^k\|\to0$;
    \item the function values $\{ \mathcal L^{\mathrm{emp}}_{\hat p}(x^k,\hat\lambda) \}_{k \geq 0}$ converge;
    \item every limit point $\hat x$ of $\{ x^k \}_{k \geq 0}$ satisfies $0 \in \mathcal{D}_{\hat p}(\hat x, \hat \lambda)$,
\end{enumerate}
where $\mathcal{D}_{\hat p}(\cdot, \hat \lambda): \mathbb{R}^n \rightrightarrows \mathbb{R}^n$ is a conservative field of the potential function $-\mathcal L^{\mathrm{emp}}_{\hat p}(\cdot,\hat\lambda)$.
\end{theorem}


\begin{proof}[Proof of \cref{main_theorem}]
The proof reduces SPBM to the stochastic approximation framework of
\citet{DacDruKakLee2018}. Since the duals and penalties change only slightly in the limit, the iteration asymptotically tracks SGD applied to the fixed function
$\mathcal{L}^{\mathrm{emp}}_{\hat p}(\cdot,\hat\lambda)$, and we may verify their
conditions [A.1]--[A.5] and [B.1]--[B.2] for the pair
$\bigl(\Phi_{\hat p}(\cdot,\hat\lambda),\,\mathcal{D}_{\hat p}(\cdot,\hat\lambda)|_{\mathcal{X}}\bigr)$
consisting of a Lyapunov function and a restricted conservative field. All references to
[A.$\cdot$] and [B.$\cdot$] below are assumptions of \citet{DacDruKakLee2018}.

\emph{Step 1 (Preliminaries).} By \cref{lem:dual} the duals and penalties converge almost
surely to some $\hat\lambda\in[0,\Lambda]^m$ and $\hat p\in[\ell_c,u_c]^m$, and by
\cref{lem:prox_center_collapses} the proximal term vanishes, $\|x^k-s^k\|\to0$ almost surely. This proves items \emph{(1)} and \emph{(2)} of the theorem.

\emph{Step 2 (The limiting objects).} The limits $\hat\lambda,\hat p$ are random, hence so are the objects built from them, whereas the framework of \citet{DacDruKakLee2018} is
stated for a fixed set-valued map. We therefore argue path-wise. Let $\Omega_0$ be the
intersection of the almost-sure events of \cref{asm:C}, \cref{asm:C_instead_as_proposition}, \cref{proposition_definability_D_uppersem}, \cref{lem:bounded_variance_of_yk}, \cref{remark:martingale_sequence_difference}, \cref{cor:A4}, \cref{lem:dual}, \cref{lem:prox_center_collapses}, \cref{lem:consistency}, \cref{cor:A5}; being a finite intersection of events of full
measure, $\Omega_0$ still maintains full measure: $P(\Omega_0)=1$. We fix $\omega\in\Omega_0$ and write $\hat\lambda=\hat\lambda(\omega)$, $\hat p=\hat p(\omega)$ for the remainder of the proof.

Set $G := \mathcal{D}_{\hat p}(\cdot,\hat\lambda)|_{\mathcal{X}}$ and
$\Phi_{\hat p}(\cdot,\hat\lambda) := \max\{c,\ \mathcal{L}^{\mathrm{emp}}_{\hat p}(\cdot,\hat\lambda)\}$ with $c$ as in \cref{lem:reg_lyapunov}. We cannot take $\mathcal{X} = \mathbb{R}^n$ in the proof: with
$\Phi_{\hat p}(\cdot,\hat\lambda) = \max\{\Lemp_{\hat p}(\cdot,\hat\lambda), c\}$
the Lyapunov function is constant on $\{\Lemp_{\hat p}(\cdot,\hat\lambda) < c\}$,
so the strict descent required by [B.2] is not available on curves entering that
set, whereas without the maximum, Assumption [B] requires
$\Lemp_{\hat p}(\cdot,\hat\lambda)$ to be bounded from below on all of
$\mathbb{R}^n$, which \cref{lem:reg}(1) gives only on compact sets. The
choice $\mathcal{X} = \{x : \|x\| \le R+1\}$ is admissible because [A.1] requires
only that the limit points of $\{x^k\}$ lie in $\mathcal{X}$, which holds by \cref{asm:C}. Following \citet[Section~3.1]{DacDruKakLee2018}, a trajectory of $G$ is an arc
$x : \mathbb{R}_+ \to \mathbb{R}^n$ satisfying $\dot x(t) \in G\big(x(t)\big)$ for a.e.
$t \ge 0$; since the inclusion is meaningful only where $G$ is defined, every
such arc takes its values in $\mathcal{X}$, and [B.1]--[B.2] are accordingly
conditions on curves in $\mathcal{X}$ alone. 
Since \cref{thm:morse_sard} \citep[Theorem 5]{bolte2021conservative} and \cref{lem:chain_rule} \citep[Lemma 2]{bolte2021conservative} are stated for
conservative fields on $\mathbb{R}^n$, we verify [B.1] and [B.2] for the
unrestricted field $\mathcal{D}_{\hat p}(\cdot,\hat\lambda)$ and transfer them to
$G$ in the proof, ultimately proving [B.1]-[B.2] for $G$.

\emph{Step 3 (Conditions [A.1]--[A.5]).} 
\begin{table}[h]
\centering
\caption{Each condition is established by the result named
below.}
\begin{tabular}{@{}lll@{}}
\toprule
& \textbf{Condition} & \textbf{Established by} \\
\midrule
{}[A.1] & limit points of $\{x^k\}$ lie in $\mathcal{X}$ & \cref{remark:limit_points_are_contained_in_X} \\
{}[A.2] & $\sup_k\|x^k\|<\infty$, $\sup_k\|y^k\|<\infty$ & \cref{asm:C}, \cref{proposition_definability_D_uppersem} \\
{}[A.3] & $\sum_k\alpha^k=\infty$, $\sum_k(\alpha^k)^2<\infty$ & \cref{asm:C_instead_as_proposition} (2) \\
{}[A.4] & $\sum_{k}\alpha^k\zeta^k$ converges & \cref{cor:A4} \\
{}[A.5] & Ces\`aro averages of $y^{k_j}$ approach $G(\hat x)$ & \cref{cor:A5} \\
\bottomrule
\end{tabular}
\end{table}

\emph{Step 4 (Conditions [B.1]--[B.2]).} $\Phi_{\hat p}(\cdot,\hat\lambda)$ is continuous
and bounded below by \cref{lem:reg_lyapunov}. The weak Sard property [B.1] is
\cref{lem:morse_sard} and the descent property [B.2] is \cref{lem:descent_property}; both are proven for $G$, as described in \emph{Step 2}.

\emph{Step 5 (Conclusion).} All hypotheses of \citet[Theorem~3.2]{DacDruKakLee2018} hold
for $\omega$, so every limit point $\hat x$ of $\{x^k\}$ satisfies $0\in G(\hat x)$, and
the sequence $\{\Phi_{\hat p}(x^k,\hat\lambda)\}_{k\geq0}$ converges. Since
$x^k\in\mathcal{X}$, \cref{lem:reg_lyapunov} gives
$\Phi_{\hat p}(x^k,\hat\lambda)=\mathcal{L}^{\mathrm{emp}}_{\hat p}(x^k,\hat\lambda)$, which
proves item \emph{(3)}; and $G(\hat x)=\mathcal{D}_{\hat p}(\hat x,\hat\lambda)$ for
$\hat x\in\mathcal{X}$, which proves item \emph{(4)}. As $\omega\in\Omega_0$ was arbitrary and
$P(\Omega_0)=1$, the conclusions hold almost surely.

\end{proof}

\begin{remark}[Dual and penalty paths have finite length by construction]
    The bounds of the dual variables and of penalty parameters are not influenced by the progress of the algorithm. Instead, convergence of the duals and penalty parameters is imposed by our summable schedules, and both $\hat \lambda, \hat p$ end up within a fixed distance from the initial iterates (\cref{lem:dual}). Summability is however critical for the analysis: without it, there are no guarantees that the random $\hat \lambda, \hat p$ exist. The convergence theorem reduces to the framework of \cite{DacDruKakLee2018} which assumes a path-wise fixed set-valued map $G$, that we build with fixed $\hat \lambda, \hat p$. Setting instead $\sum_k(1-\gamma^k)~=~\infty, \sum_k(1-\kappa^k)=\infty$ would leave the duals and penalty variables moving, making the framework unavailable. We leave this extension for future work. 
\end{remark}

\newpage
\begin{corollary}[Meaning of criticality in the smooth setting]\label{cor:criticality_meaning}
Let $\hat x$ be a limit point of $\{x^k\}_{k\geq 0}$ as in \cref{main_theorem}, and suppose that $\bar f(\cdot;\mathbf{j})$ and $\bar g_i(\cdot;\mathbf{j})$ are continuously differentiable on a neighbourhood of $\hat x$ for every
$\mathbf{j}\in\{1,\dots,M\}^{B}$ and every $i\in\{1,\dots,m\}$. Then item (4) of
\cref{main_theorem} gives
\[
0 \;=\; \frac{1}{M^{B}}\sum_{\mathbf{j}\in\{1,\dots,M\}^{B}}
  \Biggl[\,\nabla\bar f(\hat x;\mathbf{j})
  \;+\;\sum_{i=1}^{m}
  \underbrace{\hat\lambda_i\,
    \varphi'\!\bigl(\bar g_i(\hat x;\mathbf{j})/\hat p_i\bigr)}_{
    =:\ \bar\lambda_i^{\mathbf{j}}\ \geq\ 0}
  \,\nabla\bar g_i(\hat x;\mathbf{j})\Biggr].
\]
That is, $\hat x$ is a stationary point of the averaged Lagrangian with non-negative batch-dependent
effective multipliers $\bar\lambda_i^{\mathbf{j}}$.
\end{corollary}
\begin{proof}
Fix $\mathbf{j}$. By \cref{prop:penalty_reg} $\varphi\in C^{2}$, and by the hypothesis on the continuous differentiability of $\bar f(\cdot;\mathbf{j})$ and $\bar g_i(\cdot;\mathbf{j})$, it holds that $\overline{\mathcal{L}}_{\hat p}(\cdot,\hat\lambda;\mathbf{j})$ is continuously
differentiable on a neighbourhood of $\hat x$ with
\[
  \nabla_x\overline{\mathcal{L}}_{\hat p}(\hat x,\hat\lambda;\mathbf{j})
  =\nabla\bar f(\hat x;\mathbf{j})
   +\sum_{i=1}^{m}\hat\lambda_i\,
    \varphi'\!\bigl(\bar g_i(\hat x;\mathbf{j})/\hat p_i\bigr)\,
    \nabla\bar g_i(\hat x;\mathbf{j}).
\]
Since the Lagrangian is continuously differentiable near $\hat x$, it is also strictly differentiable. By strict differentiability and by \citep[Proposition 2.2.4]{clarke1990optimization}, it holds that $\partial_x\overline{\mathcal{L}}_{\hat p}(\hat x,\hat\lambda;\mathbf{j})
=\{\nabla_x\overline{\mathcal{L}}_{\hat p}(\hat x,\hat\lambda;\mathbf{j})\}$, hence $\mathcal{D}_{\hat p}(\hat x,\hat\lambda)$ is a singleton given by the negative of the average and item (4) of \cref{main_theorem} is the stated equation. Finally
$\hat\lambda_i\geq 0$ by \cref{asm:C_instead_as_proposition}(1) and $\varphi'>0$ by
\cref{prop:penalty_reg}, hence $\bar\lambda_i^{\mathbf{j}}\geq 0$.
\end{proof}

For a better understanding, we give \cref{tab:proof} that links each lemma and the proof of the main theorem.
\begin{table}[h]
\centering
\caption{Structure of the convergence proof. Throughout, $\mathcal{D}_{\hat p}(\cdot,\hat\lambda)$
is the conservative field of $-L^{\mathrm{emp}}_{\hat p}(\cdot,\hat\lambda)$ from
\cref{proposition_definability_D_uppersem}, and $G$ is its restriction to $\mathcal{X}$.
The Lyapunov function $\Phi_{\hat p}(\cdot,\hat\lambda)$ of \cref{lem:reg_lyapunov} coincides
with $L^{\mathrm{emp}}_{\hat p}(\cdot,\hat\lambda)$ on $\mathcal{X}$. The upper blocks show the list of fixed-path-wise objects and the
prerequisites specific to SPBM; the lower block lists the conditions of \citet{DacDruKakLee2018}.
Conditions A.1--A.5 concern the recursion and the map $G$; B.1--B.2 concern the pair
$(\Phi_{\hat p}(\cdot,\hat\lambda),G)$. Together they yield \cref{main_theorem} via
\citep[Theorem 3.2]{DacDruKakLee2018}.}
\small
\begin{tabular}{@{}l l l@{}}
\toprule
& \textbf{Statement} & \textbf{Established by} \\
\midrule
\multicolumn{3}{@{}l}{\emph{Objects, fixed path-wise on $\Omega_0$}}\\[2pt]
& $\mathcal{D}_{\hat p}(\cdot,\hat\lambda):\mathbb{R}^n\rightrightarrows\mathbb{R}^n$
      conservative field of $-L^{\mathrm{emp}}_{\hat p}(\cdot,\hat\lambda)$
    & \cref{proposition_definability_D_uppersem} \\
& $G:=\mathcal{D}_{\hat p}(\cdot,\hat\lambda)\big|_{\mathcal{X}}$, \emph{Step 2} of the proof \cref{main_theorem}
    & \cref{main_theorem} \\
& $\Phi_{\hat p}(\cdot,\hat\lambda)=\max\{\mathcal{L}^{\mathrm{emp}}_{\hat p}(\cdot,\hat\lambda),c\}
      =\mathcal{L}^{\mathrm{emp}}_{\hat p}(\cdot,\hat\lambda)$ on $\mathcal{X}$
    & \cref{lem:reg_lyapunov} \\
\midrule
\multicolumn{3}{@{}l}{\emph{Prerequisites for SPBM}}\\[2pt]
& $\lambda^k\!\to\!\hat\lambda$, $p^k\!\to\!\hat p$ a.s.
    & \cref{lem:dual} \\
& $\|x^k-s^k\|\to 0$ a.s.
    & \cref{lem:prox_center_collapses} \\
& $\mathbb{E}[\bar y^k\mid\mathcal{F}^k]\in\mathcal{D}_{p^k}(x^k,\lambda^k)-\mu(x^k\!-\!s^k)$ a.s.
    & \cref{proposition_definability_D_uppersem} \\
& $\dist\bigl(y^k, \mathcal{D}_{\hat p}(x^k, \hat \lambda)\bigr)\to 0$ a.s.
    & \cref{lem:consistency} \\
\midrule
\multicolumn{3}{@{}l}{\emph{Conditions of \citet{DacDruKakLee2018}}}\\[2pt]
{}[A.1] & Limit points lie in $\mathcal{X}$
    & Def.\ of $\mathcal{X}$, \cref{remark:limit_points_are_contained_in_X} \\
{}[A.2] & $\sup_k\|x^k\|,\ \sup_k\|y^k\|<\infty$
    & \cref{asm:C}, \cref{proposition_definability_D_uppersem} \\
{}[A.3] & $\alpha^k\!\ge\!0$, $\sum_k\alpha^k\!=\!\infty$, $\sum_k(\alpha^k)^2\!<\!\infty$
    & \cref{asm:C_instead_as_proposition}(2) \\
{}[A.4] & $\sum_{k}\alpha^k\zeta^k$ convergent
    & \cref{cor:A4} \\
{}[A.5] & $\dist\bigl(\tfrac1N\sum_{j\le N} y^{k_j},\,G(\hat x)\bigr)\to 0$
    & \cref{cor:A5} \\[2pt]
{}[B.1] & Weak Sard: $\Phi^{-1}(v)\cap G^{-1}(0)=\emptyset$ for a dense set of $v$
    & \cref{lem:morse_sard} \\
{}[B.2] & Descent along trajectories of $\dot x\in G(x)$
    & \cref{lem:descent_property} \\[2pt]
& $\Phi_{\hat p}(\cdot,\hat\lambda)$ continuous, bounded below
    & \cref{lem:reg_lyapunov} \\
& $\mathcal{D}_{\hat p}(\cdot,\hat\lambda)$ convex compact valued, definable,
      usc; inherited by $G$ on $\mathcal{X}$
    & \cref{proposition_definability_D_uppersem} \\
\bottomrule
\end{tabular}
\label{tab:proof}
\end{table}

\begin{remark}
    The definability in some common o-minimal structure is critical in exactly one place: to prove the Weak Sard property. Unlike the descent property, which follows from the properties of a general conservative field.  Weak Sard proof, by contrast, needs Morse-Sard \cref{thm:morse_sard} \citep[Theorem 5]{bolte2021conservative}, which requires both the potential function and the conservative field to be definable. The second is not implied by the first. As \citet{bolte2021conservative} shows, by setting a conservative field
    \[
    D(s) = \begin{cases}
        [0, 1], \quad \text{if } s \in \mathbb{Z},\\
        \{ 0 \}, \quad \text{otherwise,}
    \end{cases}
    \]
    $D$ is a conservative field of a zero function, which is definable. Yet, $D$ is not definable as $\mathbb{Z}$ is not definable in any o-minimal structure. We were able to prove that our choice of a conservative field is definable since we take $\mathcal{D}_{\hat p}(\cdot, \hat \lambda)$ to be a sum of Clarke subdifferentials of a definable function, which are always definable \cite{bolte2007clarke}.
\end{remark}
\begin{remark}
    \cref{main_theorem} establishes only $D$-criticality, $0\in\mathcal{D}_{\hat p}(\hat x,\hat\lambda)$ which is weaker than Clarke criticality. Indeed, one can see from inclusion \eqref{cor:clarke_is_a_minimial_conserv_field} \citep[Corollary 1]{bolte2021conservative}, that not every $D$-critical point is necessarily Clarke critical. Nevertheless, the use $D$-criticality has been of interest in the research community recently \citep{bianchi2022convergence, bolte2023subgradient, bolte2021conservative}.
\end{remark}

\subsubsection{Proving  Dual Convergence and A.5 of \citet{DacDruKakLee2018}}\label{dual_conv_subsection}

 As it turns out, dual variables and penalty parameter iterates form Cauchy sequences almost surely, giving that their limit points exist almost surely.
\begin{lemma}[Convergence of the dual and penalty variables]\label{lem:dual}
Suppose \cref{asm:C} and \cref{asm:reg} hold, let $\{ x^k, \lambda^k, p^k, s^k \}_{k\geq0}$ be generated by \cref{alg:spbm_schedules}.
Then, almost surely, $\gamma^k,\kappa^k\to1$ and there exist (random)
$\hat\lambda\in[0,\Lambda]^m$, $\hat p\in[\ell_c,u_c]^m$ with
$\lambda^k\to\hat\lambda$ and $p^k\to\hat p$. Specifically, the path lengths are finite, for a deterministic $C < \infty$:
\[
\sum_{k=0}^\infty \|\lambda^{k+1}-\lambda^k\|\le\Lambda\sqrt m\,C\, \sum_{k=0}^\infty (1-\gamma^k),
\]
for penalty schedule \cref{penalty_updating_formulas}, it holds
\[
\sum_{k=0}^\infty \|p^{k+1}-p^k\|\le u_c \, \sqrt{m}C \, \sum_{k=0}^\infty(1 - \kappa^k),
\]
whereas for penalty schedule \cref{penalty_updating_formulas_alm}, it holds:
\[
\sum_{k=0}^\infty \|p^{k+1}-p^k\|\leq \rho \Lambda\sqrt m\,C\, \sum_{k=0}^\infty (1-\gamma^k).
\]
\end{lemma}
\begin{proof}
$\gamma^k,\kappa^k\to1$ is immediate from \eqref{eq:kappa_update}--\eqref{eq:gamma_update}.
We know that by \cref{asm:C} that $x^k \in \operatorname{int}\mathcal{X}$ almost surely. Moreover, by construction
$p_i^k\ge \ell_c>0$. We can conclude that
\begin{equation}\label{equation_boundedset}
    \text{the arguments} \quad \bar g_i(x^k) \,,\, \bar g_i(x^k)/p_i^k \quad \text{lie in a compact set}.
\end{equation}
Since $\bar g_i(\cdot; \mathbf{j})$ is locally Lipschitz (\cref{asm:reg}(1)) for any $\mathbf{j} \in \{ 1, \dots, M\}^B$, and the derivative $\varphi' > 0$ is also locally Lipschitz (can be seen from $\varphi'' \leq 1$; see \cref{eq:der_penalty_barrier_functions}), following the same arguments from the Pf. of \cref{lem:reg}(1), we can conclude:
\begin{equation}\label{equation_locallyLipschitz}
\text{both}\quad \varphi'\big(\bar g_i(x^k)/p_i^k\big), \varphi'\big(\bar g_i(x^k)\big) \quad \text{ are locally Lipschitz, hence continuous.}
\end{equation}
By using \eqref{equation_boundedset}, \eqref{equation_locallyLipschitz}, and the extreme value theorem, we can conclude that there is a deterministic $C<\infty$ with, for all $k,i$,
\[
\Big|\varphi'\big(\bar g_i(x^k)/p_i^k\big)-1\Big|\le C,
\qquad
\Big|\tfrac{1}{\varphi'(\bar g_i(x^k))+\varepsilon}-1\Big|\le C .
\]

\emph{Dual Norm Bounds.} By the ascent update rule of the dual variable, we can equivalently rewrite the iterative formula as
\[\lambda^{k+1}= \operatorname{Proj}_{[0, \Lambda]^m} \left[ \lambda^k \odot \left ( (1-\gamma^k)\left(\varphi'(\bar g(x^k)/p^k)-1\right) + 1\right ) \right].
\]
Equivalently, we can write (by $\operatorname{Proj}_{[0, \Lambda]^m} \left[ \lambda^k \right] = \lambda^k $)
\[\lambda^{k+1} - \lambda^k= \operatorname{Proj}_{[0, \Lambda]^m} \left[  \lambda^k \odot \left ( (1-\gamma^k)\left(\varphi'(\bar g(x^k)/p^k)-1\right) + 1\right )\right] - \operatorname{Proj}_{[0, \Lambda]^m} \left[ \lambda^k\right].
\]
Note that we enforce the dual random variable $\lambda^k$ to lie within the compact set $[0, \Lambda]^m$. Thus, by taking norms on each side of the equality, we get by the non-expansiveness of the $\operatorname{Proj}$ operator:
\begin{align}
\|\lambda^{k+1}-\lambda^k\|&=\|\operatorname{Proj}_{[0, \Lambda]^m} \left[ \lambda^k \odot \left ( (1-\gamma^k)\left(\varphi'(\bar g(x^k)/p^k)-1\right) + 1\right )\right] - \operatorname{Proj}_{[0, \Lambda]^m} \left[\lambda^k \right]\| \\&\leq
\|\lambda^k \odot \left ( (1-\gamma^k)\left(\varphi'(\bar g(x^k)/p^k)-1\right)\right )\|
\\ \label{equation:lambda_bounds}
&\le \Lambda\sqrt m\, (1-\gamma^k)\,C.
\end{align}
We can take the sums of each side of \eqref{equation:lambda_bounds}
\begin{equation}\label{eq:bound_dual_variable}
    \sum_{k=0}^\infty \|\lambda^{k+1}-\lambda^k\|\le\Lambda\sqrt m\,C\, \sum_{k=0}^\infty (1-\gamma^k),
\end{equation}
which by our assumption of the lemma on summability $\sum_k(1-\gamma^k) < \infty$ gives that the distance traveled by the dual variable is bounded, which gives that $\{\lambda^k\}$ has summable increments. Its tails therefore vanish. Hence for any $\varepsilon_\lambda > 0$, there exists $K_0$ such that
\[
\sum_{k=K_0}^\infty \| \lambda^{k+1} - \lambda^k \| \le \varepsilon_\lambda,
\]
Thus by the triangle inequality for any $q_1, q_2 > K_0$:
\[
\| \lambda^{q_1} - \lambda^{q_2} \| \le \varepsilon_\lambda.
\]
Since all of the preceding steps hold almost surely, we conclude that $\{ \lambda^k \}$ is Cauchy, hence $\lambda^k \to \hat \lambda \in [0, \Lambda]^m$ almost surely.

\emph{Penalty Norm Bounds.} Rewrite the penalty updating formula \eqref{penalty_updating_formulas}
\[
p^{k+1}=\operatorname{Proj}_{[\ell_c,u_c]^m} \left[ \, p^k \odot\big( (1-\kappa^k) ( \tfrac{1}{\varphi'(\bar g(x^k))+\varepsilon}-1  ) + 1\big) \right]
\]
The operator $\operatorname{Proj}$ is nonexpansive  (see the same arguments when proving \eqref{equation:lambda_bounds}) and fixes $p^k$, hence
\begin{align}
\|p^{k+1}-p^k\| &=
\| \operatorname{Proj}_{[\ell_c,u_c]^m} \left[ \, p^k \odot \left( (1-\kappa^k)\big(  \tfrac{1}{\varphi'(\bar g(x^k))+\varepsilon}-1  \big) + 1 \right )\right] - \operatorname{Proj}_{[\ell_c,u_c]^m} \left[ p^k \right] \| \\ &\leq \| p^k \odot (1-\kappa^k)\big(  \tfrac{1}{\varphi'(\bar g(x^k))+\varepsilon}-1  \big) \| \leq u_c \, \sqrt{m}C \, (1 - \kappa^k)
\end{align}
As $\sum_k(1-\kappa^k)<\infty$, the increments are summable, so $\{p^k\}$ is Cauchy almost surely and
$p^k\to\hat p\in[\ell_c,u_c]^m$ almost surely. 

When considering the penalty-updating formula \eqref{penalty_updating_formulas_alm}, we know from the \emph{Dual Norm Bounds}, that $\lambda^k \to \hat \lambda$ almost surely. Hence $p^{k+1}=\operatorname{Proj}_{[\ell_c,u_c]^m}\left[\rho\,\lambda^{k+1}\right]$ for $\rho>0$, so by the continuity of the projection and $\lambda^k\to\hat\lambda$ we get
$p^k\to\hat p$ almost surely. Using the non-expansiveness
\[
\sum_{k=0}^\infty \|p^{k+1}-p^k\|\le\rho \sum_{k=0}^\infty\|\lambda^{k+1}-\lambda^k\| \leq \rho \Lambda\sqrt m\,C\, \sum_{k=0}^\infty (1-\gamma^k),
\]
so the path length is finite as well. In either case of \eqref{penalty_updating_formulas} and \eqref{penalty_updating_formulas_alm}, it holds $p^k\to\hat p$ almost surely. 

\end{proof}

\newpage
\begin{lemma}[Prox center collapses]\label{lem:prox_center_collapses}
Suppose \cref{asm:C} and \cref{asm:reg} hold, let $\{ x^k, \lambda^k, p^k, s^k \}_{k\geq 0}$ be generated by \cref{alg:spbm_schedules}. Then almost surely
\[
\|x^k-s^k\|\to0.
\]
\end{lemma}

\begin{proof}
Let $e^k:=\|x^k-s^k\|\ge0$. From $s^{k+1}=\delta s^k+(1-\delta)x^{k+1}$,
\begin{align}
s^{k+1}-x^{k+1} &=\delta\,(s^k-x^{k+1})=\delta\,(s^k-x^k)+\delta\,(x^{k}-x^{k+1}),
\intertext{so by the triangle inequality}
e^{k+1} &\le \delta\,e^k+\delta\,\|x^{k+1}-x^k\|.
\end{align}
Now $\|x^{k+1}-x^k\|=\|\alpha^k(y^k+\zeta^k)\|\le \alpha^k\|y^k\|+\|\alpha^k\zeta^k\|\to0$,
since $\alpha^k\to0$ (as $\sum_k(\alpha^k)^2<\infty$), $\|y^k\|$ is bounded almost surely
(\cref{proposition_definability_D_uppersem}), and $\alpha^k\zeta^k\to0$ (the terms of the convergent series
by \cref{cor:A4} vanish); the same argument is used in \citep[Lemma 3.3]{DacDruKakLee2018}. Setting $b^k:=\delta\|x^{k+1}-x^k\|\ge0$ we obtain
\begin{equation}\label{eq_prox_center_collapse}
e^{k+1}\le\delta e^k+b^k,\qquad \delta\in[0,1),\qquad b^k\to0 .
\end{equation}
We argue that $e^k \to 0.$ Indeed, by observing \cref{lemma:xu_ek_convergence} \citep[Lemma 2.1]{xu2004viscosity}, we can rewrite \eqref{eq_prox_center_collapse} as
\[
e^{k+1} \le \big(1-(1-\delta)\big)\, e^k + b^k.
\]
hence both
\[
\sum_{k=1}^\infty (1-\delta) = \infty, \qquad \limsup_{k \to \infty }\frac{b^k}{1-\delta} = 0, \qquad \text{hold,}
\]
hence by applying \cref{lemma:xu_ek_convergence}, we have $e^k \to 0$.
All steps
hold on the almost-sure event of \cref{asm:C}, so $e^k\to0$ almost surely. Note that \cref{lemma:xu_ek_convergence} \citep[Lemma 2.1]{xu2004viscosity} requires $\delta \in (0, 1)$. We can see that for $\delta = 0$, the almost sure convergence $e^k\to0$ is immediate.
Consequently, along any subsequence with $x^{k_j}\to\hat x$ we have
$\|s^{k_j}-\hat x\|\le e^{k_j}+\|x^{k_j}-\hat x\|\to0$, so $s^{k_j}\to\hat x$; i.e.\ every
limit point $(\hat x,\hat s)$ of $\{(x^k,s^k)\}$ satisfies $\hat s=\hat x$, and therefore
$\mu(\hat x-\hat s)=0$.
\end{proof}

Now that we have proven the convergence of dual variables, and that the proximal term vanishes, one last lemma that we need is one that states a desired property of the approximate evaluation $y^k$ of $G$. For that, we need to prove that a residual function is locally Lipschitz in $x$. For the next lemma, define the finite constants
\begin{align}
  &G_{\max} :=\max_{i\le m,\; \mathbf{j} \in \{1, \dots, M \}^B}\;\max_{x\in \mathcal{X}}\bigl|\bar{g}_i(x; \mathbf{j})\bigr|, \qquad
  L_g   :=\max_{i\le m,\;\mathbf{j} \in \{1, \dots, M \}^B}\Lip_{\mathcal{X}}\bigl(\bar{g}_i(\cdot; \mathbf{j})\bigr), \\
  & S_1 :=\max_{t\in T}\varphi'(t), \qquad
  S_2 :=\max_{t\in T}\varphi''(t), \qquad T:=\bigl[-G_{\max}/\ell_c,\;G_{\max}/\ell_c\bigr].
\end{align}
Finiteness of $L_g$ and $G_{\max}$ follows from local Lipschitzness on the
compact set $\mathcal{X}$ together with the finiteness of the dataset
(\cref{asm:reg}(1)--(2)); finiteness of $S_1,S_2$ from
$\varphi\in C^{2}$ and compactness of $T$. For both barriers one has in fact
$S_2\le1$, since $\varphi''\equiv1$ on the quadratic branch and $\varphi''<1$
below it.

\newpage
\begin{lemma}[Perturbation in the duals and penalties]\label{lem:jointlip}
Let $(\lambda,p),(\lambda',p')\in[0,\Lambda]^{m}\times[\ell_c,u_c]^{m}$ and set $r_{\mathbf{j}}^{(\lambda, p), (\lambda', p')}: \mathbb{R}^n \to \mathbb{R}$ as
\[
  r_{\mathbf{j}}^{(\lambda, p), (\lambda', p')}
   :=\overline{\mathcal{L}}_{p}(\cdot,\lambda; {\mathbf{j}})-\overline{\mathcal{L}}_{p'}(\cdot,\lambda'; {\mathbf{j}}).
\]
then 
\begin{enumerate}
    \item for all $x,x'\in \mathcal{X}$, and any ${\mathbf{j}} \in \{ 1, \dots, M\}^B$
    $$\bigl|r_{\mathbf{j}}^{(\lambda, p), (\lambda', p')}(x)-r_{\mathbf{j}}^{(\lambda, p), (\lambda', p')}(x')\bigr|
  \;\le\;C_\mathcal{X}\bigl(\|\lambda-\lambda'\|+\|p-p'\|\bigr)\,\|x-x'\|,$$
  \item for all $x\in \operatorname{int}\mathcal{X}$, $\eta \in \mathbb{R}$, and any ${\mathbf{j}} \in \{ 1, \dots, M\}^B$
  $$
  \quad \eta \,\partial_x \overline{\mathcal{L}}_{p}(x,\lambda; \mathbf{j})\;\subseteq\;
  \eta \, \partial_x \overline{\mathcal{L}}_{p'}(x,\lambda'; \mathbf{j})
  \;+\; |\eta| \,C_\mathcal{X}\bigl(\|\lambda-\lambda'\|+\|p-p'\|\bigr)\,\bball.
  $$
\end{enumerate}
Specifically, item 1 gives $\Lip_{\mathcal{X}}\bigl(r_{\mathbf{j}}^{(\lambda, p), (\lambda', p')}\bigr)
\le C_\mathcal{X}\bigl(\|\lambda-\lambda'\|+\|p-p'\|\bigr)$.

\end{lemma}
\begin{proof} 
\emph{Step 1 (Lipschitz constant).}
Write $r_{\mathbf{j}}=r_{\mathbf{j}}^{(\lambda,p),(\lambda',p')}$. Since the objective terms cancel, the fundamental theorem of calculus yields
$$
\begin{aligned}
r_{\mathbf{j}}(x)-r_{\mathbf{j}}(x')
=& \sum_{i=1}^m(\lambda_i-\lambda_i')
\int_{\bar{g}_i(x'; {\mathbf{j}})}^{\bar{g}_i(x; {\mathbf{j}})}
\varphi'(\tau/p_i)\,d\tau\\
&+\sum_{i=1}^m\lambda_i'
\int_{\bar{g}_i(x'; \mathbf{j})}^{\bar{g}_i(x; {\mathbf{j}})}
\left[\varphi'(\tau/p_i)-\varphi'(\tau/p_i')\right]\,d\tau .
\end{aligned}
$$

For every $\tau$ in these intervals, $|\tau|\le G_{\max}$ and
$\tau/p_i,\tau/p_i'\in T$. Hence

$$
|\varphi'(\tau/p_i)|\le S_1.
$$
Since $\varphi'$ is $S_2$-Lipschitz on $T$,
\[
  \bigl|\varphi'(\tau/p_i)-\varphi'(\tau/p_i')\bigr|
  \le S_2\Bigl|\frac{\tau}{p_i}-\frac{\tau}{p_i'}\Bigr|
  = S_2\,|\tau|\,\frac{|p_i-p_i'|}{p_ip_i'}
  \le S_2\,\frac{G_{\max}}{\ell_c^{2}}\,|p_i-p_i'| .
\]

Moreover, by

$$
|\bar{g}_i(x; {\mathbf{j}})-\bar{g}_i(x'; {\mathbf{j}})|\le L_g\|x-x'\|,
$$

and $\lambda_i'\le\Lambda$, we obtain

$$
|r_j(x)-r_j(x')|
\le L_g\left(
S_1\sum_{i=1}^m|\lambda_i-\lambda_i'|
+\frac{\Lambda S_2G_{\max}}{\ell_c^2}
\sum_{i=1}^m|p_i-p_i'|
\right)\|x-x'\|.
$$

Using $\sum_i|a_i|\le\sqrt m\,\|a\|$ (holds by Cauchy-Schwarz) applied to $a=\lambda-\lambda'$ and $a=p-p'$ separately gives the result with 
\[
C_{\mathcal X}
=\sqrt m\,L_g
\max\left\{S_1,\frac{\Lambda S_2G_{\max}}{\ell_c^2}\right\}.
\]
\\ \emph{Step 2 (Subdifferential inclusion).}
Write $\overline{\mathcal{L}}_{p}(\cdot,\lambda; \mathbf{j})= \overline{\mathcal{L}}_{p'}(\cdot,\lambda'; \mathbf{j})+ r_{\mathbf{j}}^{(\lambda,p), (\lambda',p')}(\cdot)$, with residual $r_{\mathbf{j}}$ defined as in \emph{Step 1}. The Clarke sum
rules \cite[Theorem 13.10, Theorem 13.20]{clason2026introduction} give for any $\eta \in \mathbb{R}$
\[
\eta \,\partial_x\overline{\mathcal{L}}_{p}(x,\lambda; \mathbf{j})\subseteq \eta \,\partial_x \overline{\mathcal{L}}_{p'}(x,\lambda'; \mathbf{j}),
+\partial \bigl(\eta \, r_{\mathbf{j}}(x) \bigr).
\] 
Since $x \in \operatorname{int}\mathcal{X}$ (\cref{remark:limit_points_are_contained_in_X}), there is $\varepsilon > 0$ with $\varepsilon\mathbb{B} + x  \subseteq \mathcal{X}$, so $\eta r_j$ is Lipschitz near $x$ with constant $|\eta|\,\Lip_{\mathcal{X}}(r_{\mathbf{j}})$. \citep[Lemma 13.2]{clason2026introduction} gives that the subdifferential is bounded by $|\eta|\,\Lip_{\mathcal{X}}(r_{\mathbf{j}})$. \emph{Step 1} bounds
$\Lip_\mathcal{X}(r_{\mathbf{j}})$ by $C_\mathcal{X}(\|\lambda-\lambda'\|+\|p-p'\|)$ which completes the proof.
\end{proof}

\begin{lemma}[Asymptotic consistency of the search directions]\label{lem:consistency}
Let \cref{asm:C} and \cref{asm:reg} hold, let summability properties as stated in \cref{lem:dual} hold, let $\hat\lambda,\hat p$
be the limits defined by \cref{lem:dual}, and set $D: \mathbb{R}^n \rightrightarrows \mathbb{R}^n$ as
\begin{equation}\label{eq:G}
  D(x):= \mathcal{D}_{\hat p }(x, \hat \lambda).
\end{equation}
Then, almost surely,
\[
  \lim_{k\to\infty}\dist\bigl(y^{k},\, D(x^{k})\bigr)=0 .
\]
\end{lemma}
\begin{proof}
Fix path-wise some $\omega$ on the almost sure event on which \cref{asm:C}, \cref{lem:dual}, \cref{lem:prox_center_collapses}, \cref{proposition_definability_D_uppersem} simultaneously hold; being a finite intersection of events of
full measure, it has also full measure. Fix $k$.

\emph{Step 1 (splitting off the proximal term).}
By \cref{proposition_definability_D_uppersem} there is
$w^{k}\in\mathcal{D}_{p^{k}}(x^{k},\lambda^{k})$ with
$y^{k}=w^{k}-\mu\,(x^{k}-s^{k})$. Hence, for every $t\in D(x^{k})$,
\[
  \|y^{k}-t\|\;=\;\|(w^{k} - \mu \,(x^k - s^k))-t\|\,\le\, \|w^k - t \|+\mu\,\|x^{k}-s^{k}\| ,
\]
and taking the infimum over $t\in D(x^{k})$,
\begin{equation}\label{eq:consistency_split}
  \dist\bigl(y^{k},D(x^{k})\bigr)
  \;\le\;\dist\bigl(w^{k},D(x^{k})\bigr)+\mu\,\|x^{k}-s^{k}\| .
\end{equation}
\emph{Step 2 (bounding the conservative fields).}
Write $K:=C_{\mathcal{X}}\bigl(\|\lambda-\lambda'\|+\|p-p'\|\bigr)$ and, for each
$\mathbf{j}$, $r_{\mathbf{j}}:= \overline{\mathcal{L}}_{p}(\cdot,\lambda; \mathbf{j})- \overline{\mathcal{L}}_{p'}(\cdot,\lambda'; \mathbf{j})$,
so that $\overline{\mathcal{L}}_{p}(\cdot,\lambda; \mathbf{j})=\overline{\mathcal{L}}_{p'}(\cdot,\lambda'; \mathbf{j})+r_{\mathbf{j}}$ as in \cref{lem:jointlip}. Since $x\in\operatorname{int}\mathcal{X}$, by \cref{lem:jointlip} (2) with $\eta = -1$, it holds
\[
  -\partial_x \overline{\mathcal{L}}_{p}(x,\lambda; \mathbf{j})\;\subseteq\;
  -\partial_x \overline{\mathcal{L}}_{p'}(x,\lambda'; \mathbf{j})+K\,\mathbb{B}
  \qquad\text{for every } \mathbf{j} .
\]
Furthermore, Minkowski summation preserves inclusions (\cref{lem:minkowski_preserves_inclusion}), hence
$$
\begin{aligned}
  \mathcal{D}_{p}(x,\lambda)
  =\frac{1}{M^B}\sum_{\mathbf{j} \in \{ 1, \dots, M \}^B}& \bigl(-\partial_x \overline{\mathcal{L}}_{p}(x,\lambda; \mathbf{j})\bigr)
  \;\subseteq\; \\
  &\frac{1}{M^B}\sum_{\mathbf{j} \in \{ 1, \dots, M \}^B}\Bigl(-\partial_x \overline{\mathcal{L}}_{p'}(x,\lambda'; \mathbf{j})+K\,\mathbb{B}\Bigr)
  =\mathcal{D}_{p'}(x,\lambda')+K\,\mathbb{B},
\end{aligned}
$$
where the last equality uses $\frac{1}{M^B}\sum_{j=1}^{M^B}K\,\mathbb{B}=K\,\mathbb{B}$,
valid since $\mathbb{B}$ is convex (see \cref{lem:minkowski_ball}).

\emph{Step 3.} Therefore by setting $(\lambda, p)$ as $(\lambda^k, p^k)$, and $(\lambda', p')$ as $(\hat \lambda, \hat p)$, we bound the distance using inequality \eqref{eq:consistency_split} as
\[
\dist\bigl(y^{k},D(x^{k})\bigr) \leq K + \mu \| x^k -s ^k\| = C_{\mathcal{X}}\bigl(\|\lambda^k-\hat \lambda\|+\|p^k - \hat p\|\bigr) + \mu \| x^k - s^k \| \longrightarrow 0 \quad a.s.
\]
\end{proof}

\begin{lemma}[A.5 \cite{DacDruKakLee2018}]\label{cor:A5}
Let \cref{asm:C}, \cref{asm:reg} hold. Let the hypotheses of \cref{lem:consistency} hold and let $D$ and $G := D|_{\mathcal{X}}$ be as in
\eqref{eq:G}. Then, almost surely, for every increasing unbounded sequence
$\{k_j\}\subset\mathbb{N}$ such that $x^{k_j}\to\hat x$ for some
$\hat x\in\mathbb{R}^{n}$,
\[
  \lim_{N\to\infty}
  \dist\Bigl(\frac1N\sum_{j=1}^{N}y^{k_j},\;G(\hat x)\Bigr)=0 .
\]
\end{lemma}
\begin{proof}
Let $\Omega_0$ be the almost-sure event on which \cref{lem:consistency} holds. Fix $\omega \in \Omega_0$ and a sequence $\{k_j\}$ together with its limit $\hat x\in \mathcal{X}$  and the iterates $x^{k_j} \in \operatorname{int}\mathcal{X}$ (by \cref{remark:limit_points_are_contained_in_X}). Therefore, $G(x^{k_j}) = \mathcal{D}_{\hat p}(x^{k_j}, \hat \lambda)$ for any $k_j$ and also $G(\hat x) = \mathcal{D}_{\hat p}(\hat x, \hat \lambda)$, in other words, the two set-valued maps coincide in our setting, hence we prove the property for $\mathcal{D}_{\hat p}(\cdot, \hat \lambda)$ which gives that the property also holds for $G$.

\emph{Step 1.}
Write $D:= \mathcal{D}_{\hat p}(\cdot, \hat \lambda)$. The set-valued map $D: \mathbb{R}^n \rightrightarrows \mathbb{R}^n$ is nonempty, convex, compact and upper-semicontinuous by \cref{proposition_definability_D_uppersem}. From compactness and upper-semicontinuity we get: for every
$\varepsilon/2>0$ there is $\varrho>0$ such that
\begin{equation}\label{eq:osc}
  \|x-\hat x\|\le\varrho
  \quad\Longrightarrow\quad
  D(x)\subseteq D(\hat x)+\tfrac{\varepsilon}{2}\bball .
\end{equation}

Fix $\varepsilon>0$ and choose $\varrho$ as in \eqref{eq:osc}. Since both
\[
x^{k_j}~\to~\hat x
\]
and also, by \cref{lem:consistency},
\[
\dist(y^{k_j},D(x^{k_j}))~\to~0,
\]
there exists $J$ such that for all $j\ge J$ both
$\|x^{k_j}-\hat x\|\le\varrho$ and
$\dist\bigl(y^{k_j},D(x^{k_j})\bigr)\le\varepsilon/2$. For such $j$,
\eqref{eq:osc} gives
\[
  b^j:=\dist\bigl(y^{k_j},D(\hat x)\bigr)
  \le\dist\bigl(y^{k_j},D(x^{k_j})\bigr)+\tfrac{\varepsilon}{2}
  \le\varepsilon .
\]
As $\varepsilon>0$ was arbitrary, $b^j\to0$.

\emph{Step 2 (averaging).} The set $D(\hat x)$ is nonempty and compact (\cref{proposition_definability_D_uppersem}), so for
each $j$ there is $v^j\in D(\hat x)$ with $\|y^{k_j}-v^j\|=b^j$. Because
$D(\hat x)$ is \emph{convex}, the average $\bar v^N:=\frac1N\sum_{j=1}^{N}v^j$
again lies in $D(\hat x)$, and therefore
\begin{align}
  \dist\Bigl(\frac1N\sum_{j=1}^{N}y^{k_j},\,D(\hat x)\Bigr)
  \;\le\;\Bigl\|\frac1N\sum_{j=1}^{N}\bigl(y^{k_j}\bigr)-\bar v^N\Bigr\|
  \; = \;\Bigl\|\frac1N\sum_{j=1}^{N}\bigl(y^{k_j}- v^j\bigr)\Bigr\|
  \;\le\;\frac1N\sum_{j=1}^{N}b^j .\label{eq:cesaro_average}
\end{align}

It remains to observe that $b^j\to0$ forces its Ces\`aro averages \eqref{eq:cesaro_average} to vanish,
\emph{irrespective of whether the series $\sum_j b^j$ converges}. Let us prove the convergence of its Ces\`aro average to 0. Pick an arbitrary $\varepsilon > 0$. Find $J$ such that $b^j < \frac{\varepsilon}{2}$ for all $j > J$.
 Then, we get for $N > J$
\begin{align}
    \frac{1}{N}\sum_{j=1}^Nb^j =  \frac{1}{N}\sum_{j=1}^J b^j + \frac{1}{N}\sum_{j = J+1}^N b^j < \frac{1}{N}\sum_{j=1}^J b^j + \frac{\varepsilon}{2}.
\end{align}
Now fix $N'$ such that both $N' > J$ and $\frac{1}{N'}\sum_{j=1}^J  b^j < \frac{\varepsilon}{2}$ hold. This is always possible since the sum $\sum_{j=1}^J b^j$ is a finite number and $N'$ can be arbitrarily large. 
Therefore,
\begin{align}
    \frac1N\sum_{j=1}^{N}b^j < \frac \varepsilon2 + \frac{\varepsilon}{2} = \varepsilon, \quad \forall N > N',
\end{align}
and $\varepsilon$ was arbitrary. 
This completes the proof that the Ces\`aro average \eqref{eq:cesaro_average} has a 0 limit with $b^j \to 0$. 
This also finishes the proof of lemma as the sequence $\{ k^j \}$ was arbitrary and $\omega \in \Omega_0$.
\end{proof}

\subsubsection{Proving Boundedness from Below, Descent and Morse-Sard Properties of a Potential Function}\label{sec:lyapunov}

To construct a Lyapunov function, we need the boundedness from below on the entire domain. This is approached in the next lemma.
\begin{lemma}[Continuous, bounded-below Lyapunov function]\label{lem:reg_lyapunov}
 Fix $\lambda\in[0,\Lambda]^{m}$ and $p\in[\ell_c,u_c]^{m}$, set
\[
  c(\lambda,p):=\min_{x\in\mathcal{X}}\mathcal{L}^{\mathrm{emp}}_{p}(x,\lambda),
  \qquad
  \Phi_{p}(x,\lambda):=\max\bigl\{\mathcal{L}^{\mathrm{emp}}_{p}(x,\lambda),\,c(\lambda,p)\bigr\}.
\]
Then
\begin{enumerate}
  \item $c(\lambda,p)\in\mathbb{R}$;
  \item $\Phi_{p}(\cdot,\lambda)$ is locally Lipschitz, hence continuous, on $\mathbb{R}^{n}$,
        and satisfies $\Phi_{p}(\cdot,\lambda)\ge c(\lambda,p)$, hence is bounded from below
        on $\mathbb{R}^{n}$;
  \item $\Phi_{p}(x,\lambda)=\mathcal{L}^{\mathrm{emp}}_{p}(x,\lambda)$ for every
        $x\in\mathcal{X}$.
\end{enumerate}
\end{lemma}

\begin{proof}
By \cref{lem:reg}(1), $\mathcal{L}^{\mathrm{emp}}_{p}(\cdot,\lambda)$ is locally
Lipschitz on $\mathbb{R}^{n}$, hence continuous. Since $\mathcal{X}$ is compact and
nonempty, the extreme value theorem gives that the minimum defining $c(\lambda,p)$ is
finite and attained, proving item~1.

For item~2, a maximum of two locally Lipschitz functions is locally Lipschitz, and a
constant is locally Lipschitz, so $\Phi_{p}(\cdot,\lambda)$ is locally Lipschitz and
therefore continuous. By definition of the maximum,
$\Phi_{p}(x,\lambda)\ge c(\lambda,p)$ for every $x\in\mathbb{R}^{n}$, so
$\Phi_{p}(\cdot,\lambda)$ is bounded from below on $\mathbb{R}^{n}$.

For item~3, let $x\in\mathcal{X}$. By definition of $c(\lambda,p)$ as the minimum over
$\mathcal{X}$ we have $\mathcal{L}^{\mathrm{emp}}_{p}(x,\lambda)\ge c(\lambda,p)$, and
hence $\Phi_{p}(x,\lambda)=\mathcal{L}^{\mathrm{emp}}_{p}(x,\lambda)$.
\end{proof}

Next, we prove the Weak-Sard property of the potential function $\Phi$ together with the conservative field $\mathcal{D}_{\hat p}(\cdot, \hat \lambda)$. 

\begin{lemma}[Weak-Sard, condition B.1 in \cite{DacDruKakLee2018}]\label{lem:morse_sard}
    Let \cref{asm:C} and \cref{asm:reg} hold. Set $G: \mathcal{X} \rightrightarrows \mathbb{R}^n$
    \[
    G(x) = \left(\mathcal{D}_{\hat p}(\cdot, \hat \lambda)|_{\mathcal{X}}\right)(x)
    \]
    and $\Phi(\cdot) := \Phi_{\hat p}(\cdot, \hat \lambda)$ defined in \cref{lem:reg_lyapunov}. Then for a dense set $V$ of values $v \in \mathbb{R}$, the intersection of
    \[
    \Phi^{-1}(v) \cap G^{-1}(0)
    \]
    is empty.
\end{lemma}
\begin{proof}
Since $\partial(-f) = -\partial f$ for a Clarke subdifferential \citep[Theorem 13.10]{clason2026introduction}, and $\sum_j (-A_j) = -\sum_j A_j$ for Minkowski sums, $\mathcal{D}_{\hat p}(x, \hat \lambda)$ is the pointwise negative of the conservative field $- \mathcal{D}_{\hat p}(x, \hat \lambda)$. Since $0 \in A \iff 0 \in -A$, the two critical sets coincide: 
\begin{equation}\label{eq:d_criticality_same}
    \mathcal{D}_{\hat p}(\cdot, \hat \lambda)^{-1}(0) = \left(-\mathcal{D}_{\hat p}(\cdot, \hat \lambda)\right)^{-1}(0).
\end{equation}

\emph{Step 1. (Apply Morse-Sard Theorem and construct a candidate for $V$)}\\
By \cref{lem:reg}(1), we know that $\Lemp_{\hat p}(x, \hat \lambda)$ is definable, and locally Lipschitz. By \cref{proposition_definability_D_uppersem} and \cref{remark_conservative_field_D}, $-\mathcal{D}_{\hat p}(x, \hat \lambda)$ is a definable conservative field of $\Lemp_{\hat p}(x, \hat \lambda)$. Combining these, \cref{thm:morse_sard} \citep[Theorem 5]{bolte2021conservative} gives that the set
\[
\operatorname{CritL} :=
        \bigl\{\, \Lemp_{\hat p}(x, \hat \lambda) \;:\; x \in \mathbb{R}^n,\; 0 \in -\mathcal{D}_{\hat p}(x, \hat \lambda) \,\bigr\}
\]
is finite.
Fix $c:=\min_{x\in\mathcal{X}}\mathcal{L}^{\mathrm{emp}}_{\hat p}(x,\hat \lambda)$ as in \cref{lem:reg_lyapunov}, and define $V = \mathbb{R} \setminus (\operatorname{CritL} \cup \{ c \})$. We argue that
\begin{equation}\label{eq:more_sard_eq1}
\Phi^{-1}(v) \cap \left( -\mathcal{D}_{\hat p}(\cdot, \hat \lambda)\right)^{-1}(0) = \emptyset.
\end{equation}

\emph{Step 2. (Proving \cref{eq:more_sard_eq1})}\\
Indeed, if we write down the definition of $\Phi^{-1}(v)$
\[
\Phi^{-1}(v) = \begin{cases}
    \Lemp_{\hat{p}}(\cdot, \hat \lambda)^{-1}(v), \text{if }v> c,\\
    \{ x : \Lemp_{\hat{p}}(x, \hat \lambda) \leq c \}, \text{if }v= c, \\
    \emptyset , \text{if }v < c.
\end{cases}
\]
Consider an arbitrary $v \in V$. Let us assume first \emph{(i)} $v> c$, then if $\mathcal{L}^{\mathrm{emp}}_{\hat{p}}(\cdot, \hat \lambda)^{-1}(v) \cap \left( -\mathcal{D}_{\hat p}(\cdot, \hat \lambda) \right)^{-1}(0) \neq \emptyset$, that means that $v \in \operatorname{CritL}$ which is a contradiction with the construction of $V$. Next \emph{(ii)} assume that $v = c$ which is not possible by construction of $V$. Lastly, \emph{(iii)} $v < c$, then $\Phi^{-1}(v) = \emptyset$, hence trivially the equality is true.

\emph{Step 3. (Proving density of $V$)}\\
Next, we prove that $V$ is dense. This is the same as proving that the interior of the complement is empty, i.e. the interior of
\[
\mathbb{R} \setminus V = \mathbb{R} \setminus \left( \mathbb{R} \setminus (\operatorname{CritL} \cup \{ c \}) \right) = \operatorname{CritL} \cup \{ c \}.
\]
Since $\operatorname{CritL}$ is finite, $\operatorname{CritL}\cup \{ c\}$ is also finite, hence has empty interior.

\emph{Step 4. (proving Weak-Sard for $G$)}\\
Finally, we show that the property also holds for $G$ and not only for $\mathcal{D}_{\hat p}(\cdot, \hat \lambda)$. Since $G$ is its restricted version, for any $q \in \mathbb{R}^n$
\[
G^{-1}(q) = \{ x \in \mathbb{R}^n \mid q \in \mathcal{D}_{\hat p}(x, \hat \lambda), x\in {\mathcal{X}}\} \subseteq \{ x \in \mathbb{R}^n \mid q \in \mathcal{D}_{\hat p}(x, \hat \lambda)\} = \mathcal{D}_{\hat p}(\cdot, \hat \lambda)^{-1}(q).
\]
But then by \eqref{eq:more_sard_eq1}
\[
\Phi^{-1}(v) \cap G^{-1}(0) \subseteq \Phi^{-1}(v) \cap \mathcal{D}_{\hat p}(\cdot, \hat \lambda)^{-1}(0) = \Phi^{-1}(v) \cap \left(-\mathcal{D}_{\hat p}(\cdot, \hat \lambda)\right)^{-1}(0) = \emptyset,
\]
where the second last equality was proven in the beginning.
\end{proof}

\begin{lemma}[Descent, condition B.2 in \cite{DacDruKakLee2018}]\label{lem:descent_property}
    Let \cref{asm:C} and \cref{asm:reg} hold and fix a realisation of
    $(\hat\lambda,\hat p)$. Let $x:\mathbb{R}_+\to \mathcal{X}$ be absolutely continuous
    with
    \[
        \dot x(t)\in \left(\mathcal{D}_{\hat p}(\cdot, \hat \lambda)|_{\mathcal{X}}\right)(x(t)) = G\big(x(t)\big)
        \qquad\text{for a.e. } t\ge 0 .
    \]
    Then for a.e.\ $t\ge0$ the velocity is the minimal-norm element,
    \[
        \|\dot x(t)\|=\dist\Big(0, G\big(x(t)\big)\Big),
    \]
    and for every $T\ge 0$
    \[
        \Phi_{\hat p}\bigl(x(T),\hat\lambda\bigr)-\Phi_{\hat p}\bigl(x(0),\hat\lambda\bigr)
        \;=\;-\int_0^T \dist^2\Big(0,G\big(x(t)\big)\Big)\,dt
        \;\le\;0 .
    \]
    In particular, if $0\notin G\bigl(x(0)\bigr)$ there exists $T>0$ with
    \[
        \Phi_{\hat p}\bigl(x(T),\hat\lambda\bigr)
        \;<\;\sup_{t\in[0,T]}\Phi_{\hat p}\bigl(x(t),\hat\lambda\bigr)
        \;\le\;\Phi_{\hat p}\bigl(x(0),\hat\lambda\bigr),
    \]
    i.e.\ condition B.2 of \cite{DacDruKakLee2018} holds for the pair
    $\bigl(\Phi_{\hat p}(\cdot,\hat\lambda),\, G(\cdot)\bigr)$.
\end{lemma}
\begin{proof}
    Write $\Phi := \Phi_{\hat p}(\cdot, \hat \lambda)$. By a property of a conservative field $\mathcal{D}_{\hat p}(\cdot,\hat\lambda)$ of a potential function $-\Lemp_{\hat p}(\cdot,\hat\lambda)$, we apply \cref{lem:chain_rule} \citep[Lemma 2]{bolte2021conservative} to the absolutely
    continuous curve $x$, for a.e.\ $t$ to get
    \begin{equation}
        \tfrac{d}{dt}\bigl[-\Lemp_{\hat p}(x(t),\hat\lambda)\bigr]
        =\langle v,\dot x(t)\rangle
        \qquad\text{for \emph{all} } v\in\mathcal{D}_{\hat p}(x(t),\hat\lambda).
    \end{equation}
    Since the absolutely continuous curve $x$ is a solution to a DI defined through $G$, $x(t) \in \mathcal{X}$ must hold, therefore $\left(\mathcal{D}_{\hat p}(\cdot,\hat\lambda)|_{\mathcal{X}}\right)\left(x(t)\right) = \mathcal{D}_{\hat p}(x(t),\hat\lambda)$ for all $t$. Hence it must also hold that (note the restriction)
    \begin{equation}\label{eq:descent_property_2}
    \tfrac{d}{dt}\bigl[-\Lemp_{\hat p}(x(t),\hat\lambda)\bigr]
        =\langle v,\dot x(t)\rangle
        \qquad\text{for \emph{all} } v\in G\bigl(x(t)\bigr). 
    \end{equation}

    By using \cref{lem:general_lemma_chain_rule_dist_eq} with $G = \mathcal{D}_{\hat p}(\cdot,\hat\lambda)|_{\mathcal{X}}$ and \eqref{eq:descent_property_2}, we get that $\| \dot x(t) \| = \operatorname{dist}\Big(0, G\big(x(t)\big)\Big)$, and $\tfrac{d}{dt}\bigl[-\Lemp_{\hat p}(x(t),\hat\lambda)\bigr]
        = \| \dot x(t) \|^2$, hence
    \[
    \tfrac{d}{dt}\bigl[\Lemp_{\hat p}(x(t),\hat\lambda)\bigr]
        = -\| \dot x(t) \|^2 = -\operatorname{dist}^2\Big(0, G\big(x(t)\big)\Big).
    \]
    Since $\Lemp_{\hat p}(\cdot, \hat \lambda)$ is locally Lipschitz, and $x$ is absolutely continuous, the composition $\Lemp_{\hat p}(\cdot, \hat \lambda) \circ x$ is also absolutely continuous. Therefore, by integrating both sides of the equation, we get that
    \[
    \Lemp_{\hat p}(x(T), \hat \lambda)-\Lemp_{\hat p}(x(0), \hat \lambda)=-\int_0^T\operatorname{dist}^2\Big(0, G\big(x(t)\big)\Big)\,dt\;\le\;0.
    \]
    Since $x(t) \in \mathcal{X}$ holds, we have $\Phi\big(x(t)\big)=\Lemp_{\hat p}(x(t),\hat\lambda)$ for all $t \geq 0$ by \cref{lem:reg_lyapunov}. Hence, we conclude that
    \[
    \Phi\big(x(T)\big)-\Phi\big(x(0)\big)=-\int_0^T\operatorname{dist}^2\Big(0, G\big(x(t)\big)\Big)\,dt\;\le\;0.
    \]
    Now assume that $0~\notin~G\big(x(0)\big)$.
    Then there must exist $T > 0$ such that for any $t \in [0, T]$ it holds that $\dist\Big(0, G\big(x(t)\big)\Big) > 0$: in particular,
    \[
    -\int_0^T\operatorname{dist}^2\Big(0, G\big(x(t)\big)\Big) < 0.
    \]
    Suppose this is not true. Then there must exist a sequence $t_j \downarrow 0$, such that $v^j \in G\big(x(t_j)\big)$ and $v^j = 0$. Since this holds for $G$, then it also must hold for $\mathcal{D}_{\hat p}(x(t_j), \hat \lambda)$. As $x$ is continuous, we have that $x(t_j) \to x(0)$. By the outer-semicontinuity, we have that $0 \in D_{\hat p}(x(0), \hat \lambda ) = G\big(x(0)\big)$ which is a contradiction.

    Hence $t\mapsto\Phi\big(x(t)\big)$ is descending, so it must hold that $\Phi\big(x(T)\big)<\Phi\big(x(0)\big)$. Finally by monotonicity
    $\sup_{t\in[0,T]}\Phi\big(x(t)\big)=\Phi\big(x(0)\big)$.
\end{proof}


\subsection{Counter Example - No subgradient-type method escapes the flat region.}\label{sec:counter_example}

In this section, we provide a simple counterexample that shows non-convergence of SPBM and also of the state-of-the-art methods for constrained machine learning (SSw, SSL-ALM). Both the objective and the constraint functions are made up of standard machine learning building blocks.

Consider a single-unit ReLU model $\hat y := [wz+b]_+$ with parameters $(w,b)$
and a dataset of $1000$ feature--label pairs $(z_i,y_i)$ with $z_i=1$ for all $i \in \{ 0, \dots, 999 \}$, $y_i=1$ for $i \in \{ 0, \dots, 99\}$, and $y_i=0$ for $i \in \{ 100, \dots, 999\}$. Assume the preactivation $x:=w+b$. We set the objective function $f$ as a mean squared error:
\begin{align}
f(x) &= \frac{1}{1000}\sum_{i = 0}^{999} \left( \hat y_i - y_i \right)^2 \\
& = 0.1\big([x]_+-1\big)^2 + 0.9\,[x]_+^2 = \begin{cases} x^2-0.2\,x+0.1, & \text{if }x\ge0,\\[2pt] 0.1, &   \text{if } x<0.\end{cases}
\end{align}
We assume a single constraint $g$ -- we bound the mean squared error on subgroup $\mathcal{R}$ (15 positive, 5 negative samples) within $\varepsilon := 0.25$:
\[
g(x) = \frac{1}{|\mathcal{R}|}\sum_{(z_i, y_i)\in \mathcal{R}}\big([x]_+-y_i\big)^2 - \varepsilon
= \begin{cases} x^2-1.5\,x+0.5, &  \text{if }x\ge0,\\[2pt] 0.5, &  \text{if }x<0.\end{cases}
\]
The mixed labels in $\mathcal{R}$ make the constraint estimate $\bar g$ genuinely stochastic. For
$x<0$, and mini-batch size $B=1$; the single-per-sample constraint estimate is $\bar g(x) = y_i^2-\varepsilon\in\{-0.25,\,0.75\}$. By estimating the constraint on the subgroup $\mathcal{R}$, $\bar g$ follows an $\varepsilon$-shifted Bernoulli distribution with a parameter $q:=0.75$, its moments read
\[
\mathbb{E}\big[\bar g(x,\xi)\big] = q-\varepsilon = 0.5,
\qquad
\operatorname{Var}\big(\bar g(x,\xi)\big) = q(1-q) = \tfrac{3}{16}.
\]
We show later in \cref{prop:zero_grad} that the per-sample \emph{gradients} are zero (the value of the mini-batched constraint $\bar g$ does not depend on $w, b$ in any way): the method receives a noisy signal that the constraint is violated, but the gradient is completely flat. Same applies to the function estimate $\bar f$: for $x < 0$, we get a constant value independent of $x$. 

\begin{figure}[htbp]
    \centering
    \includegraphics[width=\linewidth]{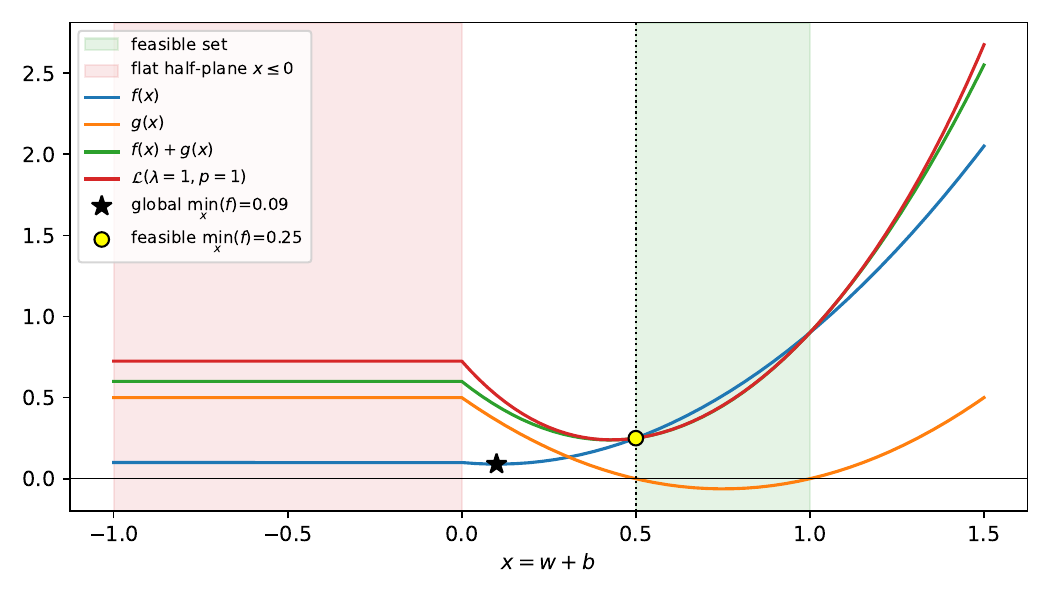}
    \hfill
    \caption{The counterexample problem. The objective function $f$ (blue) and the constraint function $g$ (orange) plotted against the \textit{pre-activation} $x = w + b$. The feasible region is shaded green, and the ``flat'' region is shaded red. A classical Lagrangian $f(x) +\lambda g(x)$ with $\lambda = 1$ is plotted in green, and the SPBM Lagrangian with $\lambda=1$, $p = 1$, and the Quadratic-Logarithmic penalty-barrier function is plotted in red.}
    \label{fig:counterexample_shape}
\end{figure}

\cref{fig:counterexample_shape} illustrates the problem. The objective, constraint, and Lagrangian functions are completely flat when $x < 0$ no matter the values of the dual parameters ($\lambda$, $p$). This leads to the primal minimization methods being stuck when the initial iterates $w, b$ are chosen such that $x = w+b < 0$.

Let us prove some properties of this counter example.

\begin{proposition}\label{kkt_counterexample}
    The problem has a unique solution in $x^* = 0.5$, with $f(x^*) = 0.25$; it satisfies KKT conditions. The feasible set is $[0.5, 1]$, hence is non-empty and compact. 
\end{proposition}
\begin{proof}
    For $x < 0$ we have $g \equiv 0.5 > 0$, so no such point is feasible. By computing the roots of $g(x)$ for $x\geq 0$, we get that the feasible set is $[0.5, 1]$, hence non-empty and compact. The constrained minimiser is the unique point $x^\star = 0.5$, with $f(x^\star) = 0.25$. Both $f$ and $g$ are continuously differentiable on a neighbourhood of $x^\star$. From the KKT stationarity condition, we find the multiplier $\lambda^* = \frac{8}{5}$, which gives dual feasibility and active constraint gives complementarity, therefore KKT holds.
\end{proof}

\begin{proposition}\label{prop:weakly_convex_setting}
    The functions $g, f$ are \textbf{not} weakly-convex.
\end{proposition}
\begin{proof}
    Let us start with $g$. Reminding the definition of weak-convexity in \cref{def:weakly_convex_function}, we can fix an arbitrary $0\leq C < \infty$. Now fix the convex combination $\alpha = 0.5$. We argue that for the fixed $C$, we can choose $x_1, x_2$ that breaks the convexity. Indeed, assume $x_1 = -x_2$ and $x_2 > 0$. That gives for $\hat g := g + \frac{C}{2}(\cdot )^2$ 
    \begin{align}\label{eq:weak_conv1}
        \hat g(\frac{1}{2}x_1 + \frac{1}{2}x_2) = \hat g(0) = \frac{1}{2}.
    \end{align}
    While also
    \begin{align}\label{eq:weak_conv2}
        \frac{1}{2}\hat g(x_1) + \frac{1}{2}\hat g(x_2) = x_2^2(\frac{1 +C}{2}) - \frac{3}{4}x_2 + \frac{1}{2}.
    \end{align}
    Computing roots, we get that for any
    \[
    x_2 \in (0, \frac{3}{2+2C})
    \]
    it holds that \cref{eq:weak_conv1} $>$ \cref{eq:weak_conv2}, hence $\hat g$ is not convex for any finite $C$, therefore not weakly-convex. 
    The same arguments apply to $f$ with $x_2 \in (0, \frac{1}{5(1+C)})$, which completes the proof.
\end{proof}

\begin{remark}
    The problem of \cref{sec:counter_example} is a single-ReLU ERM instance, arguably the simplest neural-network model there is, yet by \cref{prop:weakly_convex_setting} neither its objective nor its constraint is weakly convex. The guarantees of methods that assume weak convexity, such as SSw and SSL-ALM, therefore do not apply to it and the methods fail here. This raises the question of which assumptions are broad enough to cover neural-network ERM problems of this kind. We suggest local Lipschitzness and definability, assumed in our paper and verified for this instance in \cref{prop:counterexample_is_stosto}.
\end{remark}

\begin{proposition}\label{prop:counterexample_is_stosto}
    The problem defined through $f, g$ in \cref{sec:counter_example} is an instance of \cref{eq:stoctrsto}.
\end{proposition}
\begin{proof}
Let $B \in \mathbb{N}$ be a batch size, let $g, f$ be functions defined in \cref{sec:counter_example}. To show that the problem from \cref{sec:counter_example} is an instance of \cref{eq:stoctrsto}, we prove: 
\begin{enumerate}
    \item $f, g$ are expectations of $\bar f, \bar g$,
    \item $\bar f, \bar g$ are locally Lipschitz,
    \item $\bar f, \bar g$ are definable.
\end{enumerate}
\emph{(1)} Averaging the constraint over $\mathcal{R}$ gives $\mathbb{E}_\xi[g(x,\xi)] = q([x]_+-1)^2 + (1-q)[x]_+^2 - \varepsilon = g(x)$ with $q = 15/20$. Since $z_i \equiv 1$, the objective depends on a sample only through its label,
and both labels occur in $\mathcal{R}$. Writing $\rho_1 = 0.1$, $\rho_0 = 0.9$
for the label frequencies in the full dataset, the weights
$w(1) = \rho_1/q = 2/15$ and $w(0) = \rho_0/(1-q) = 18/5$ give
$f(x,\xi) := w(y(\xi))([x]_+ - y(\xi))^2$ with
$\mathbb{E}_\xi[f(x,\xi)] = 0.1([x]_+-1)^2 + 0.9[x]_+^2 = f(x)$. Therefore, by sampling from $\operatorname{Unif}(\mathcal{R})$ and by objective reweighting, we get that both $\bar f, \bar g$ are unbiased estimators of $f, g$. 

\emph{(2), (3)} Both $\bar g$ and $\bar f$ are built from the ReLU $[\cdot]_+$
(semialgebraic and $1$-Lipschitz), subtraction of a constant, squaring,
multiplication by the fixed weights $w(\cdot)$, and averaging over the batch.
Each of these is locally Lipschitz and definable in $\mathcal{O}$, and both
properties are preserved under composition and finite sums
\end{proof}



Lastly, we show that no method, considered in the \cref{sec:numexp}, is able to converge to the feasible region in the case of $x^0 < 0$. We start with the following lemma.
\begin{lemma}\label{prop:zero_grad}
Let $B \in \mathbb{N}$, and let $(w,b) \in \mathbb{R}^2$ satisfy $x := w + b < 0$.
Then for every realisation $\omega$, every $\lambda \in \mathbb{R}^m_+$ and every
$p \in \mathbb{R}^m_{++}$,
\[
  \partial_{(w,b)}\,\bar f = \partial_{(w,b)}\,\bar g
  = \partial_{(w,b)}\,\bar{\mathcal{L}}_p(\cdot,\lambda) = \{0\},
\]
so any subgradient oracle returns $\overline{\nabla} \mathcal{L}_p = 0$. If in
addition $s = (w,b)$, then the direction of line~4 of \cref{alg:spbm}
(identically \cref{alg:spbm_schedules}) satisfies $\bar y = 0$.
\end{lemma}

\begin{proof}
Since $z_i \equiv 1$, every model output is $[w+b]_+$, which vanishes on the open
set $U := \{(w,b) : w+b < 0\}$. Hence $\bar f$ and $\bar g$ are constant on $U$
for every realisation, and so is
$\bar{\mathcal{L}}_p(\cdot,\lambda) = \bar f + \sum_i \lambda_i p_i\,
\varphi(\bar g_i/p_i)$ for every fixed $\lambda, p$. A function constant on an
open set is continuously differentiable there with zero gradient, hence strictly differentiable and so all three Clarke
subdifferentials equal $\{0\}$; from which it follows  $\overline{\nabla}{\mathcal{L}}_p = 0$ irrespective of the realisation. Moreover if 
$s = (w,b)$ the proximal term $\mu\big((w, b) - s\big)$ in SPBM vanishes, so
$\bar y = -\overline{\nabla}{\mathcal{L}}_p - \mu\big((w,b)-s\big) = 0$.
\end{proof}

\begin{proof}[Proof of \cref{cor:no_methods_converge}]
Let $\mu \ge 0$, $\delta \in [0,1)$, and let $(w^0,b^0) \in \mathbb{R}^2$ with
$x^0 := w^0 + b^0 < 0$. 

\emph{Step 1 (SPBM).} By initialisation $s^0 = (w^0, b^0)$. We show by induction that
$(w^k, b^k) = s^k = (w^0, b^0)$ for all $k$. The base case holds. Assume
$(w^k, b^k) = s^k = (w^0, b^0)$; since $w^k + b^k = x^0 < 0$, \cref{prop:zero_grad} gives
$\bar y^k = 0 \in \mathbb{R}^2$, hence $(w^{k+1}, b^{k+1}) = (w^k, b^k) + \alpha^k\bar y^k
= (w^0, b^0)$ and $s^{k+1} = \delta s^k + (1-\delta)(w^{k+1}, b^{k+1}) = (w^0, b^0)$,
closing the induction. 

\emph{Step 2 (SSw and SSL-ALM, and general first-order schemes).} Both methods take primal steps on functions assembled
from $\bar f$ and $\bar g$ alone: SSw on the constraint estimate $\bar g$, and
SSL-ALM on the augmented Lagrangian
$\bar f + \langle\lambda,\bar g\rangle + \tfrac{1}{2\rho}\|\bar g\|^2 +\mu(\,\cdot - s)$. By \cref{prop:zero_grad} we have
$\partial_{(w,b)}\bar f = \partial_{(w,b)}\bar g = \{0\}$ on $\{w + b < 0\}$, so
every such direction vanishes there, and the induction from SPBM applies to both cases. 

More generally, the same holds for any scheme
\[
  (w^{k+1}, b^{k+1}) = (w^k, b^k) + h\big((w^k,b^k) \big)
\]
whose direction satisfies $h = 0$ whenever
$\partial_{(w,b)}\bar f = \partial_{(w,b)}\bar g = \{0\}$ and $s^k = (w^k,b^k)$.

\emph{Step 3 (Distance).} Since $(w^k, b^k) = (w^0, b^0)$, the preactivation satisfies
$x^k = w^k + b^k = x^0$ for every $k$. By \cref{kkt_counterexample} the feasible set is
$[0.5,1]$, so
$\operatorname{dist}(x^k,\{g\le 0\}) = 0.5 - x^0 = \operatorname{dist}(x^0,\{g\le 0\})$
for all $k$, which can be made arbitrarily large by the choice of $x^0$. As the
subdifferentials above are singletons, this holds for every realisation $\omega$.
\end{proof}

    
    
\begin{figure}[t]
    \centering
    \begin{subfigure}{0.49\textwidth}
        \includegraphics[width=\linewidth]{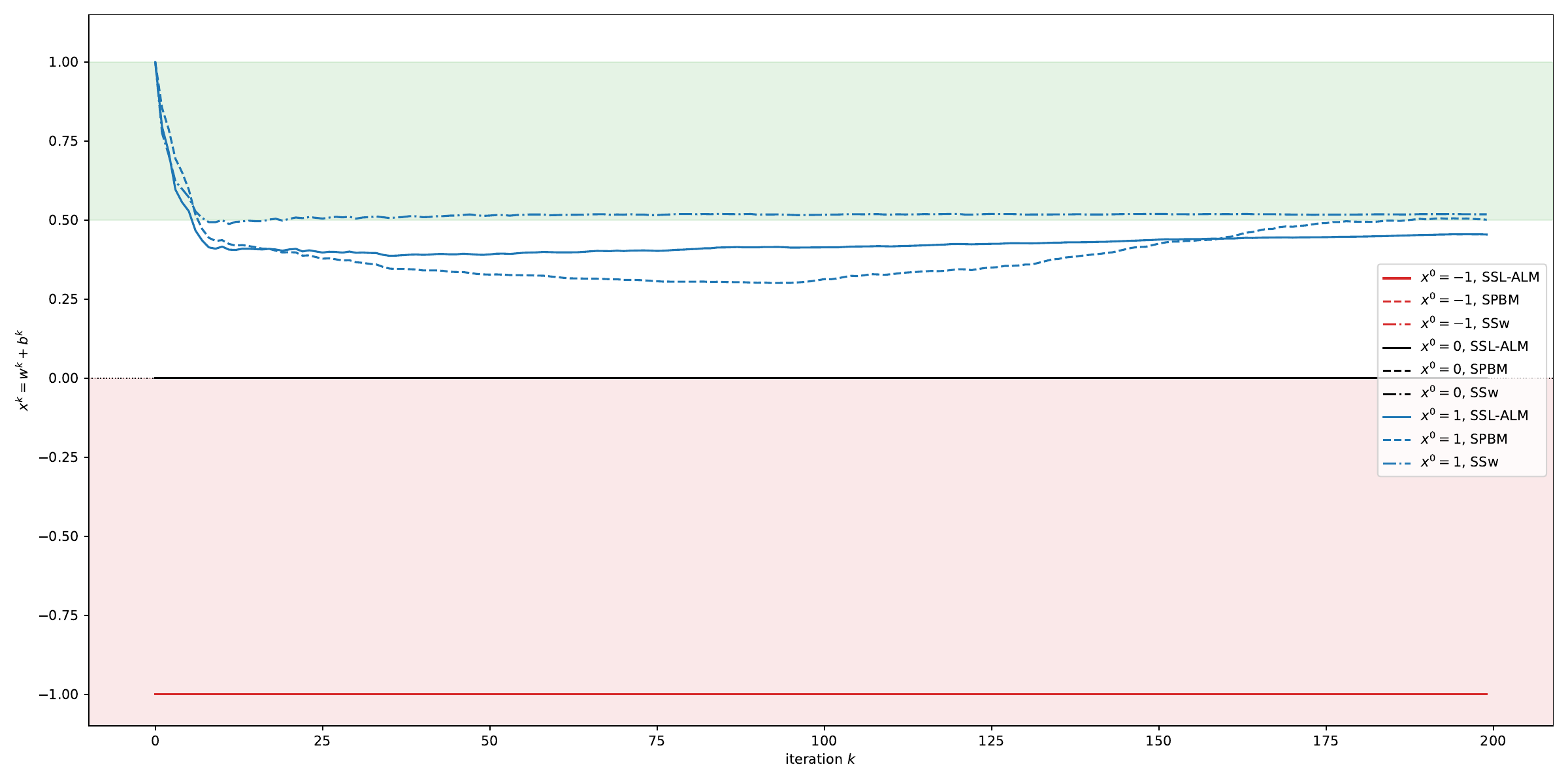}
        \label{fig:counterexample_b1_decay}
    \end{subfigure}
    \hfill
    \begin{subfigure}{0.49\textwidth}
        \includegraphics[width=\linewidth]{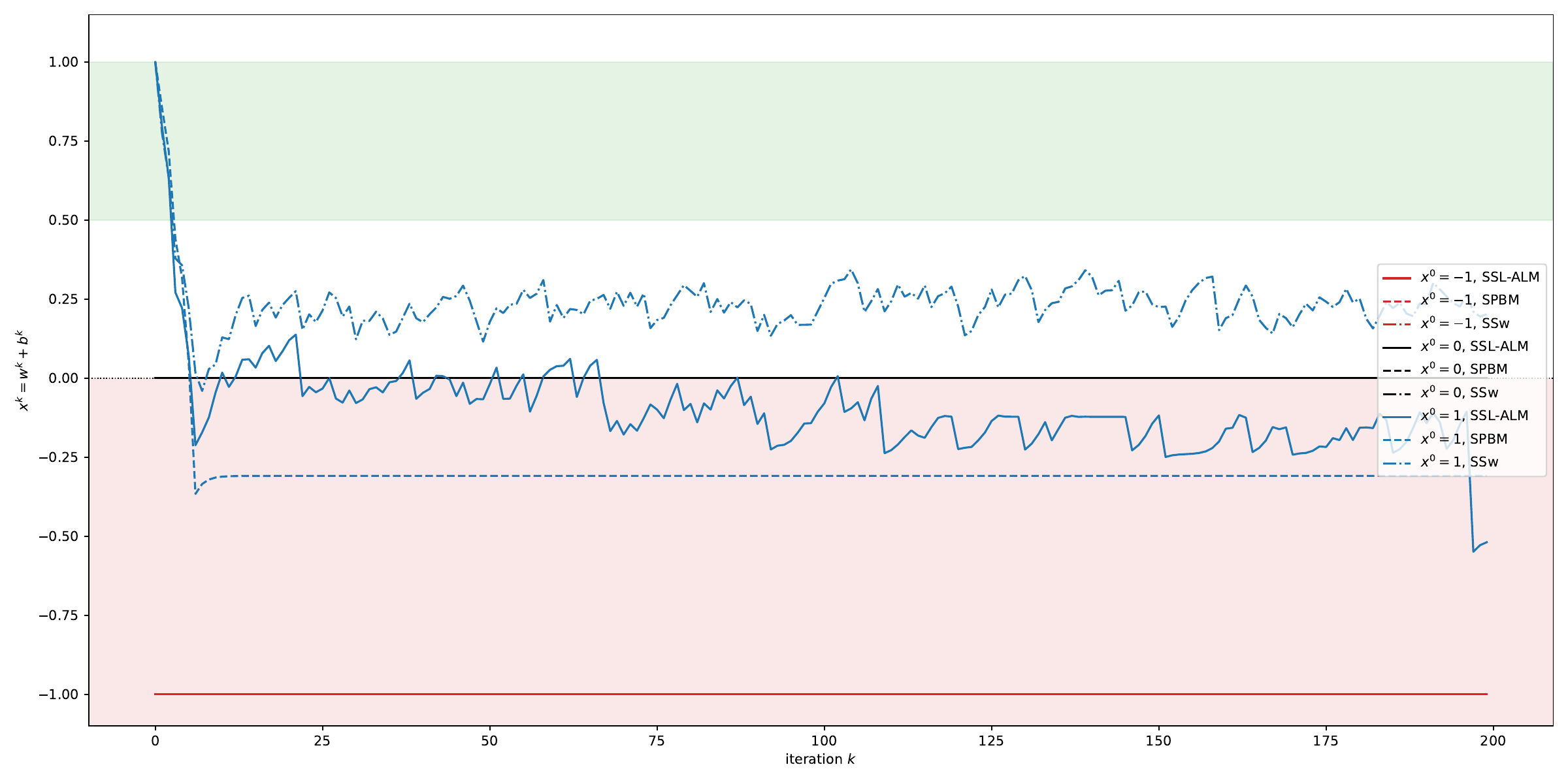}
        \label{fig:counterexample_b1}
    \end{subfigure}
    \begin{subfigure}{0.49\textwidth}
        \includegraphics[width=\linewidth]{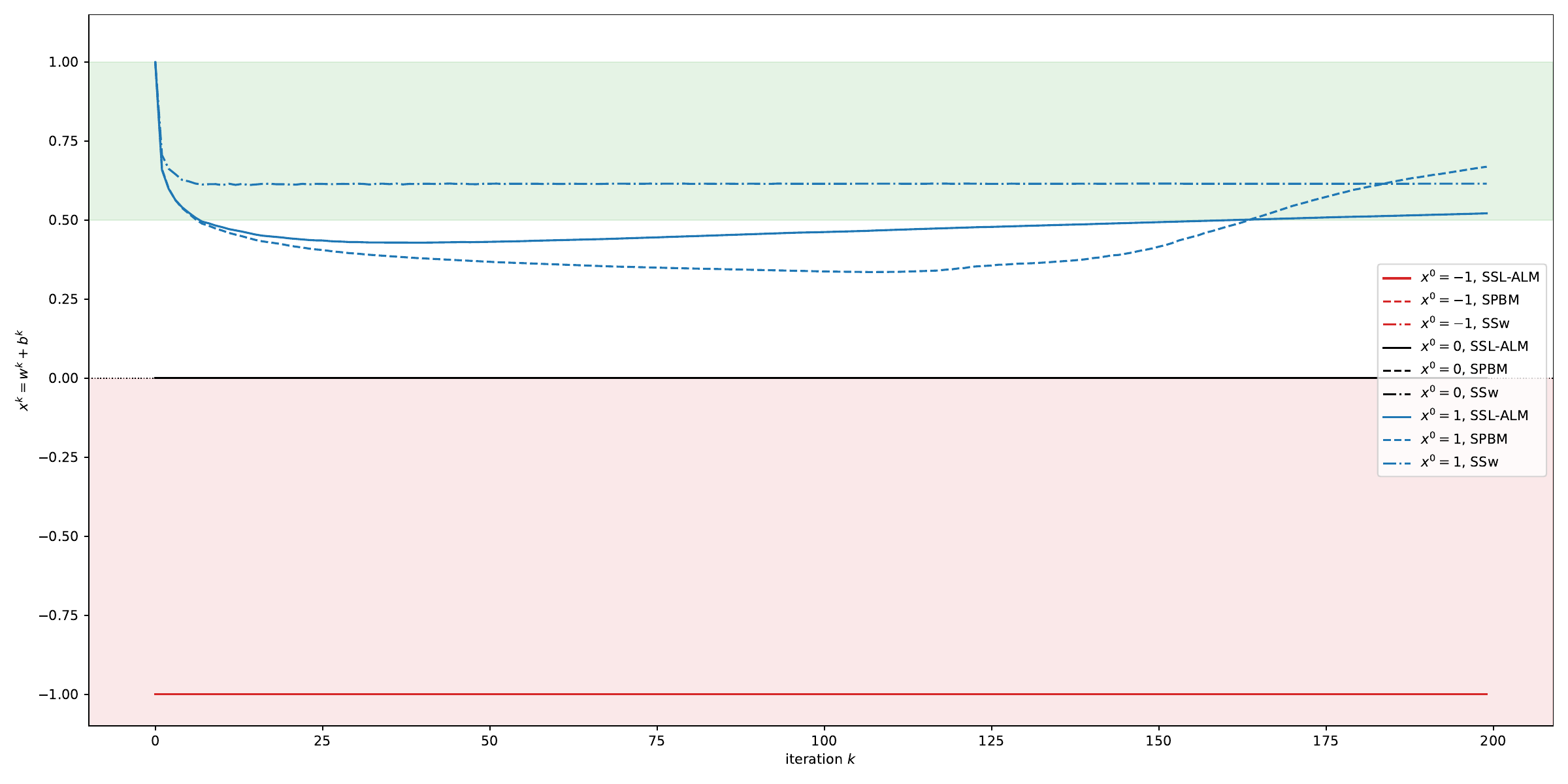}
        \label{fig:counterexample_b20_decay}
    \end{subfigure}
    \hfill
    \begin{subfigure}{0.49\textwidth}
        \includegraphics[width=\linewidth]{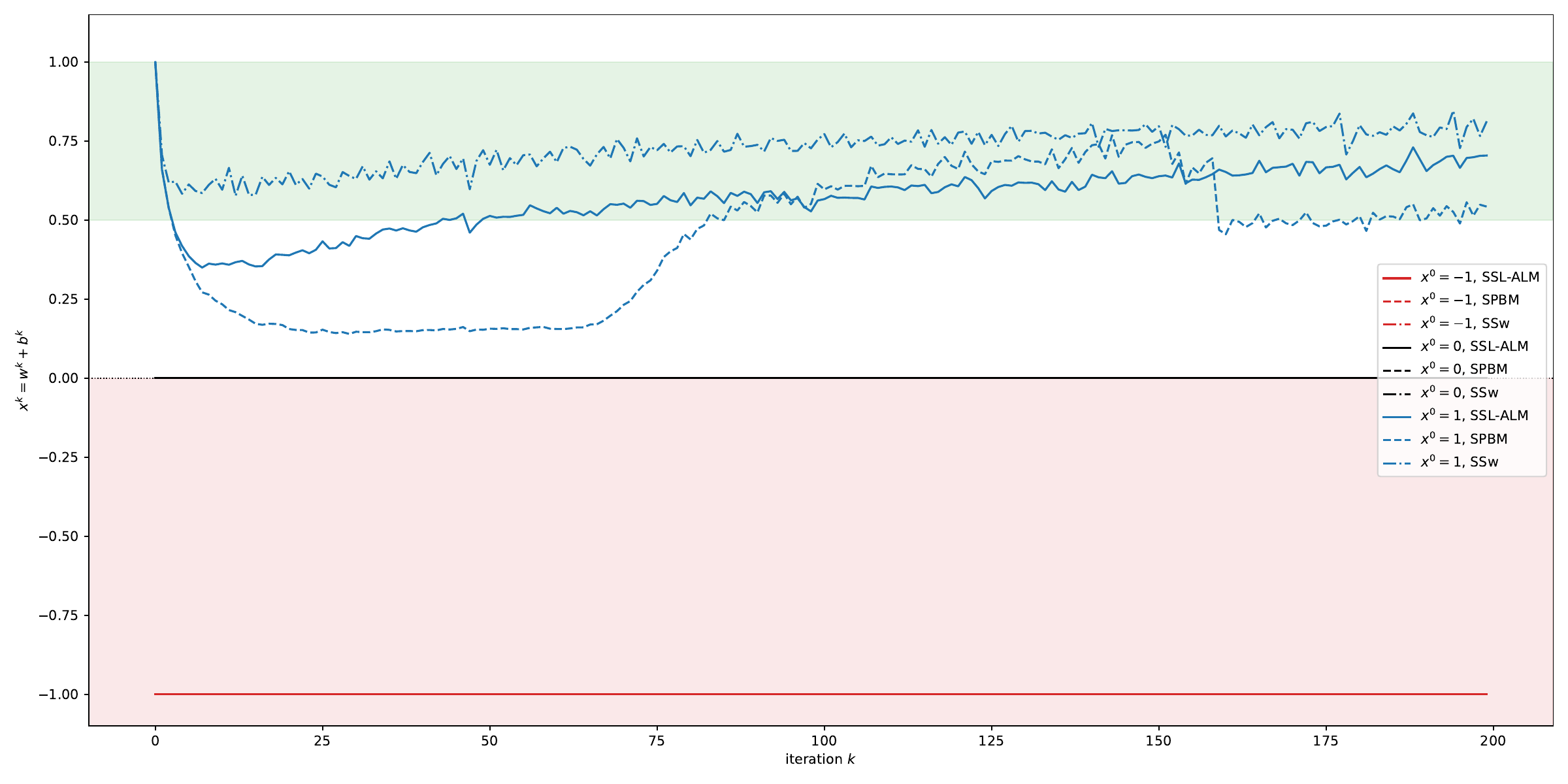}
        \label{fig:counterexample_b20}
    \end{subfigure}
    \caption{The mean trajectories of the iterates starting from $x^0=-1$ (red), $x^0=0$ (black), and $x^0=1$ (blue), produced by SSw, SSL-ALM and SPBM over 10 sampling seeds in the mini-batch setting. The plots show that for $x^0 < 0$, neither of SPBM, SSL-ALM, SSw reach the feasible region~$[0.5, 1]$. We fade the feasible region with a green color, the flat half plane with a red color. 
    \textbf{Upper left}: Batch size $B=1$ and diminishing stepsize $\alpha^0=0.1$, $\alpha_{k+1}=\frac{\alpha_k}{k}$. \textbf{Upper right}: Batch size $B=1$ and constant stepsize $\alpha_k=0.1$. \textbf{Lower left}: Batch size $B=20$ and diminishing stepsize $\alpha^0=0.1$, $\alpha^{k+1}=\frac{\alpha^k}{k}$. \textbf{Lower right}: Batch size $B=20$ and constant stepsize $\alpha_k=0.1$. }
    \label{fig:counterexample_results}
\end{figure}

\begin{remark}
    Note that both the dual variables $\lambda^k$ and the penalty parameters $p^k$ of SPBM \textbf{do} change, but their change does \textbf{not} play any role in the primal iterates, and therefore neither does it change the convergence result.
\end{remark}
What is proven in \cref{cor:no_methods_converge} is also supported by the numerics, as seen in \cref{fig:counterexample_results}.





\end{document}